\documentclass[10pt,twocolumn,letterpaper]{article}

\usepackage{cvpr}
\usepackage{times}
\usepackage{epsfig}
\usepackage{graphicx}
\usepackage{amsmath}
\usepackage{amssymb}

\usepackage{subfigure}
\usepackage{algorithm}
\usepackage{algpseudocode}
\usepackage{enumerate}
\usepackage{multirow}
\usepackage{booktabs}
\usepackage{tabularx}
\usepackage{sidecap}
\usepackage{lipsum}
\usepackage{array}
\usepackage{caption}
\usepackage[normalem]{ulem}

\newcommand\blfootnote[1]{%
  \begingroup
  \renewcommand\thefootnote{}\footnote{#1}%
  \addtocounter{footnote}{-1}%
  \endgroup
}

\usepackage[pagebackref=true,breaklinks=true,letterpaper=true,colorlinks,bookmarks=false,citecolor=blue]{hyperref} 

\cvprfinalcopy 

\def\cvprPaperID{2168} 

\ifcvprfinal\fi

\begin{document}
\title{Collaborative Distillation for Ultra-Resolution Universal Style Transfer}

\author{
Huan Wang$^{1,2*}$,~~Yijun Li$^3$,~~Yuehai Wang$^1$,~~Haoji Hu$^{1\dagger}$,~~Ming-Hsuan Yang$^{4,5}$ \\
$^1$Zhejiang University~~~$^2$Notheastern University~~~$^3$Adobe Research~~~$^4$UC Merced~~~$^5$Google Research  \\
{\tt\small wang.huan@husky.neu.edu~~yijli@adobe.com~~\{wyuehai,haoji\_hu\}@zju.edu.cn~~mhyang@ucmerced.edu}
}


\twocolumn[{
\renewcommand\twocolumn[1][]{#1}
\maketitle
\begin{center}
    \centering
    \renewcommand{\arraystretch}{0.05} 
    \begin{tabular}{c@{\hspace{0.005\linewidth}}c}
      	\includegraphics[width = 0.99\linewidth]{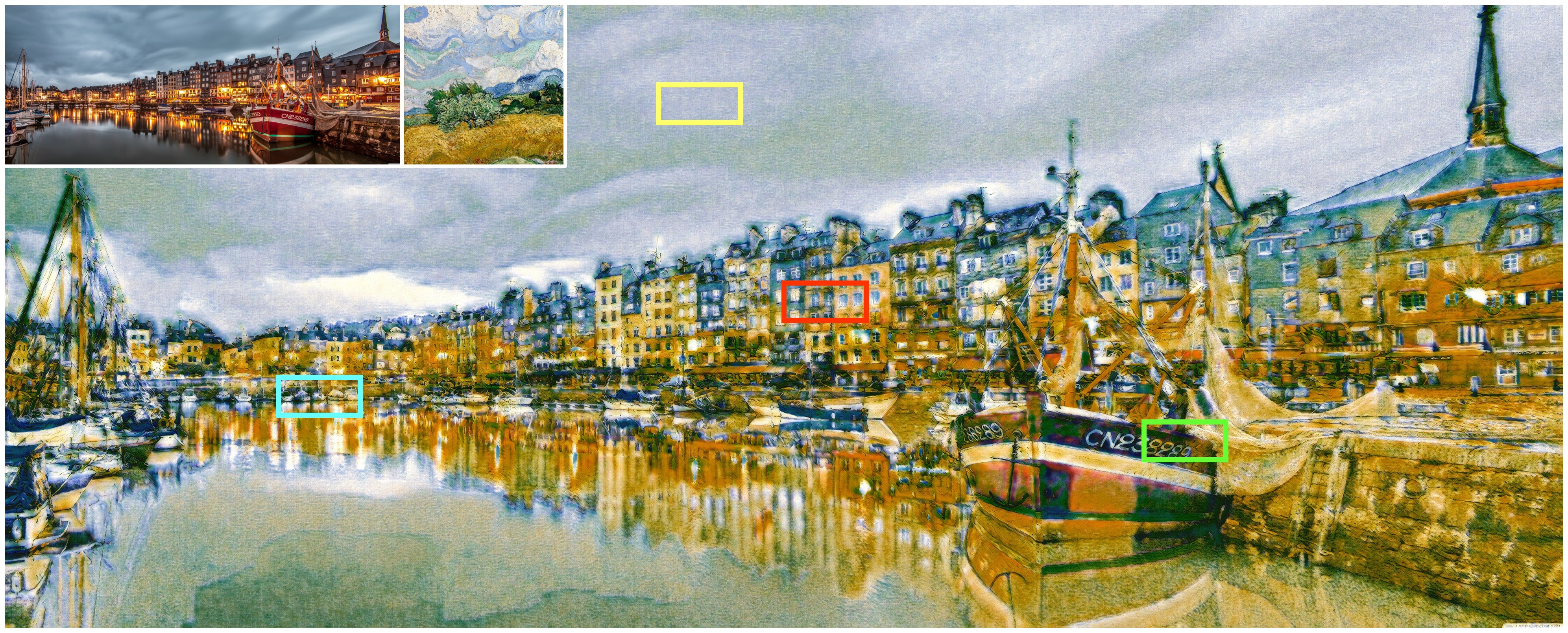}
    \end{tabular}
    \begin{tabular}{c@{\hspace{0.003\linewidth}}c@{\hspace{0.003\linewidth}}c@{\hspace{0.003\linewidth}}c@{\hspace{0.003\linewidth}}c}
      	\includegraphics[width = 1.683in, height=0.774in]{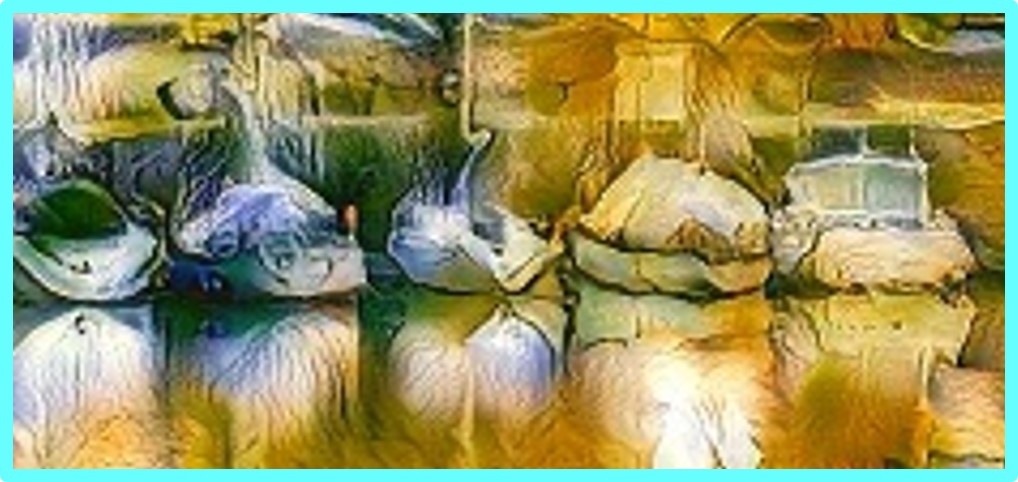} &
		\includegraphics[width = 1.683in, height=0.774in]{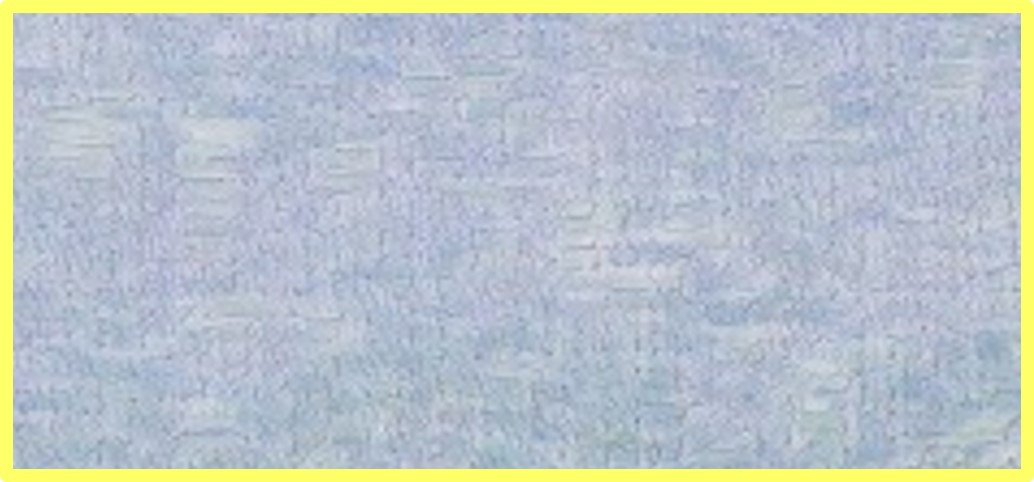} &
		\includegraphics[width = 1.683in, height=0.774in]{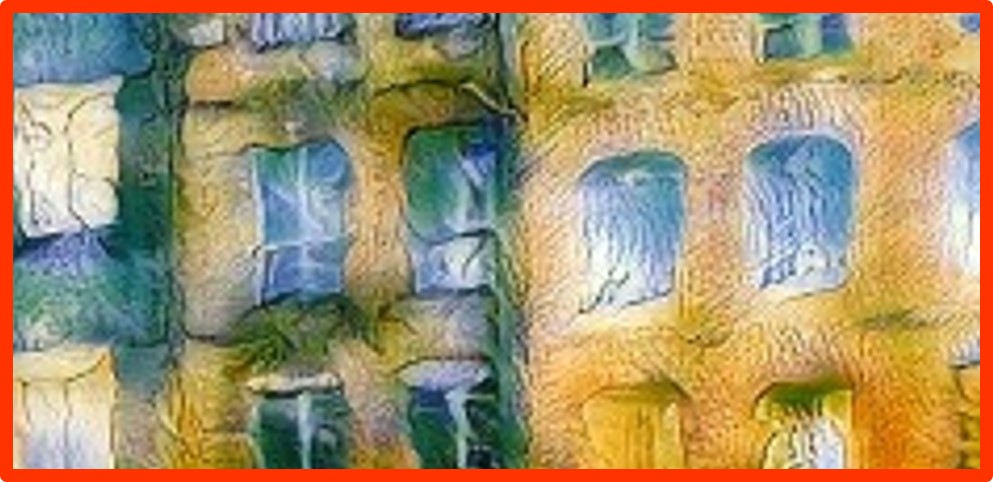} &
		\includegraphics[width = 1.683in, height=0.774in]{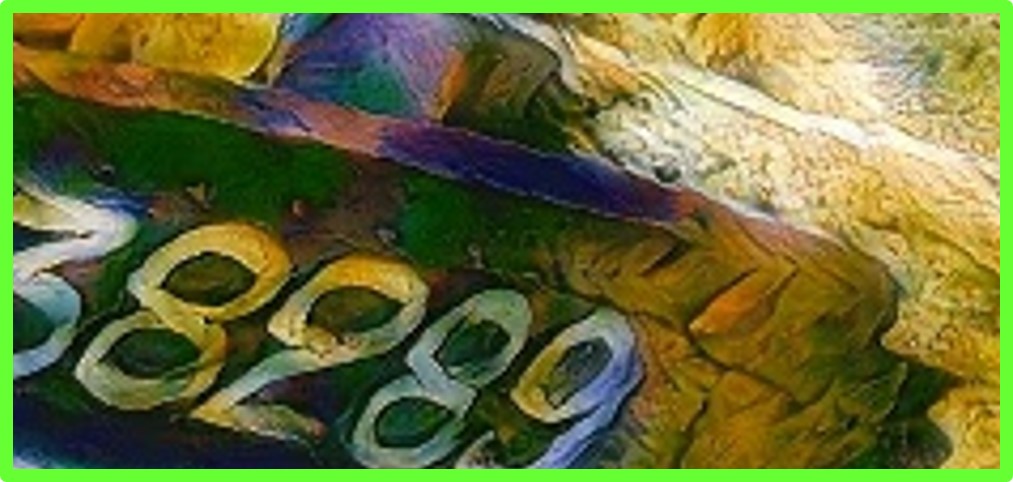} &
    \end{tabular}
  \vspace{0.05em}
  \captionof{figure}{An ultra-resolution stylized sample ($10240\times4096$ pixels), rendered in about $31$ seconds on a single Tesla P100 (12GB) GPU. On the upper left are the content and style. Four close-ups ($539\times248$) are shown under the stylized image.}
  \label{fig:UHD_image1}
\end{center}
}]

\begin{abstract}
Universal style transfer methods typically leverage rich representations from deep Convolutional Neural Network (CNN) models (e.g., VGG-19) pre-trained on large collections of images.\blfootnote{* This work is mainly done when Huan Wang was with the Department of ISEE, Zhejiang University, China. $\dagger$ Corresponding author.} 
Despite the effectiveness, its application is heavily constrained by the large model size to handle ultra-resolution images given limited memory.
In this work, we present a new knowledge distillation method (named Collaborative Distillation) for encoder-decoder based neural style transfer to reduce the convolutional filters.
The main idea is underpinned by a finding that the encoder-decoder pairs construct an exclusive collaborative relationship, which is regarded as a new kind of knowledge for style transfer models.
Moreover, to overcome the feature size mismatch when applying collaborative distillation, a linear embedding loss is introduced to drive the student network to learn a linear embedding of the teacher's features.
Extensive experiments show the effectiveness of our method when applied to different universal style transfer approaches (WCT and AdaIN), even if the model size is reduced by 15.5 times. Especially, on WCT with the compressed models, we achieve ultra-resolution (over 40 megapixels) universal style transfer on a 12GB GPU for the first time. Further experiments on optimization-based stylization scheme show the generality of our algorithm on different stylization paradigms.
Our code and trained models are available at {\color{black}{\href{https://github.com/mingsun-tse/collaborative-distillation}{https://github.com/mingsun-tse/collaborative-distillation}}}.

\end{abstract}

\section{Introduction}
Universal neural style transfer (NST) focuses on composing a content image with new styles from any reference image.
This often requires a model with considerable capacity to extract effective representations for capturing the statistics of arbitrary styles.
%
Recent universal style transfer methods based on neural networks~\cite{GatysTransfer-CVPR2016,Chen-2016-swap,Huang-2017-arbitrary,li2017universal,li2018closed,li2018learning} consistently show that employing the representations extracted by a pre-trained deep neural network like VGG-19~\cite{VGG-2014} achieves both visually pleasing transferred results and generalization ability on arbitrary style images.
%
However, given limited memory on hardware, the large model size of VGG-19 greatly constrains the input image resolution.
Up to now, current universal style transfer approaches only report results around one megapixels (\eg, $1024\times1024$ pixels) on one single GPU with 12GB memory.
Although it is likely to achieve higher resolution style transfer through multiple GPUs, the fundamental problem of the massive model size of VGG-19 remains, hindering NST from practical applications, especially on mobile devices.
%
%
%
%

\begin{figure}[t]
   \centering
   \renewcommand{\arraystretch}{0.8} 
	\begin{tabular}{@{\hspace{0\linewidth}}c@{\hspace{0.01\linewidth}}c@{\hspace{0.01\linewidth}}c@{\hspace{0.01\linewidth}}c@{\hspace{0.01\linewidth}}c}
	    \includegraphics[width=0.19\linewidth, height=0.185\linewidth]{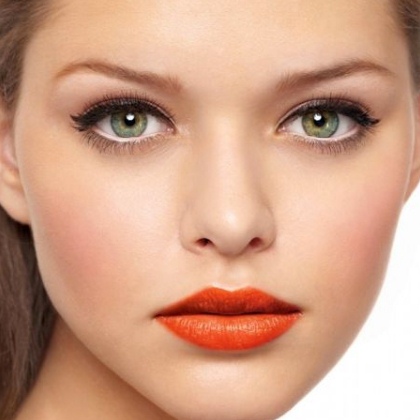} &
		\includegraphics[width=0.19\linewidth, height=0.185\linewidth]{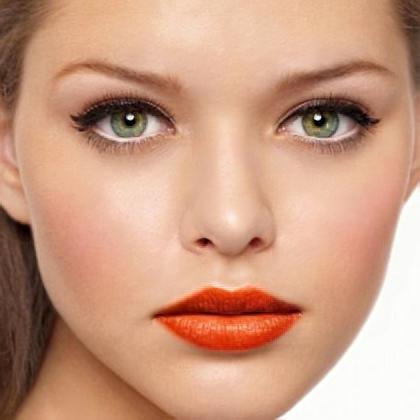} &
		\includegraphics[width=0.19\linewidth, height=0.185\linewidth]{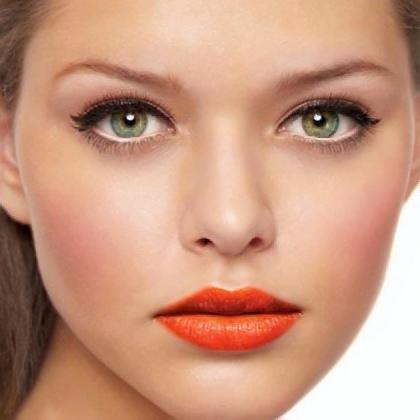} &
		\includegraphics[width=0.19\linewidth, height=0.185\linewidth]{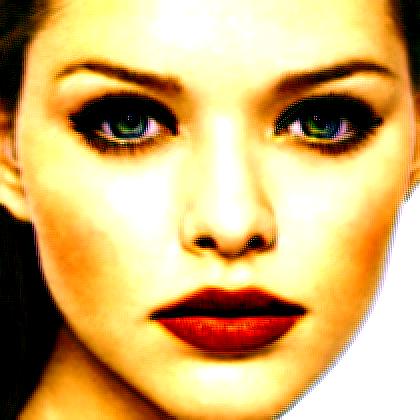} &
		\includegraphics[width=0.19\linewidth, height=0.185\linewidth]{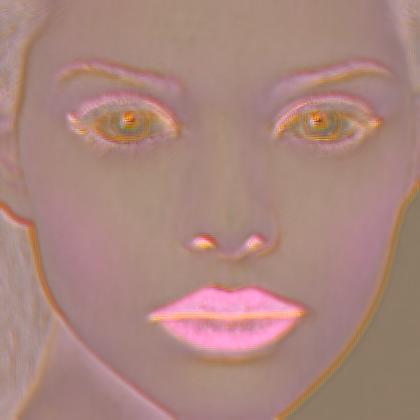} \\
	    \includegraphics[width=0.19\linewidth, height=0.185\linewidth]{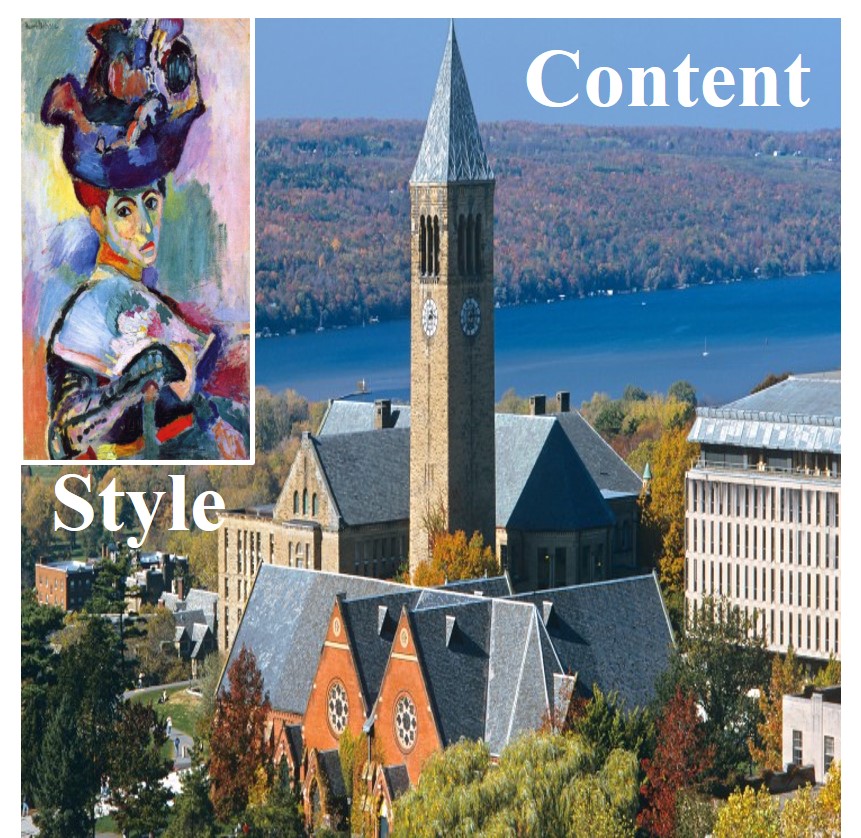} &
		\includegraphics[width=0.19\linewidth, height=0.185\linewidth]{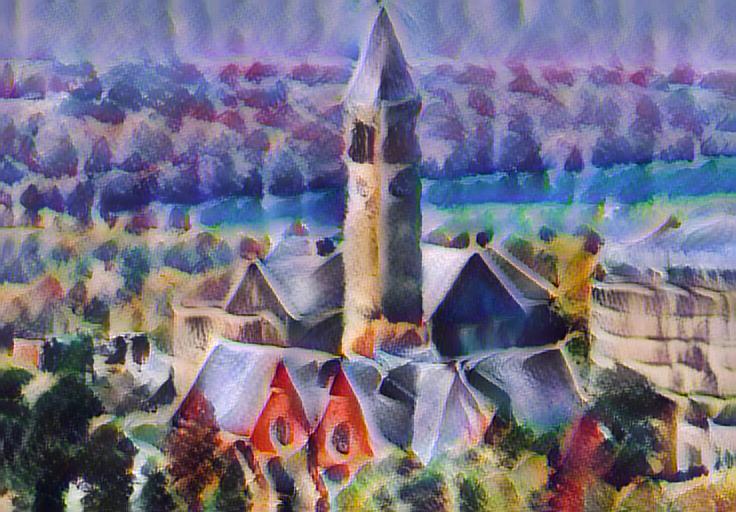} & 
		\includegraphics[width=0.19\linewidth, height=0.185\linewidth]{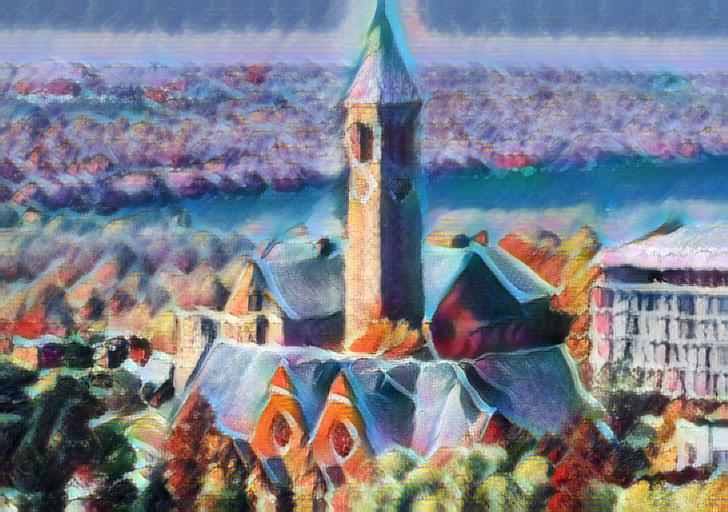} & 
		\includegraphics[width=0.19\linewidth, height=0.185\linewidth]{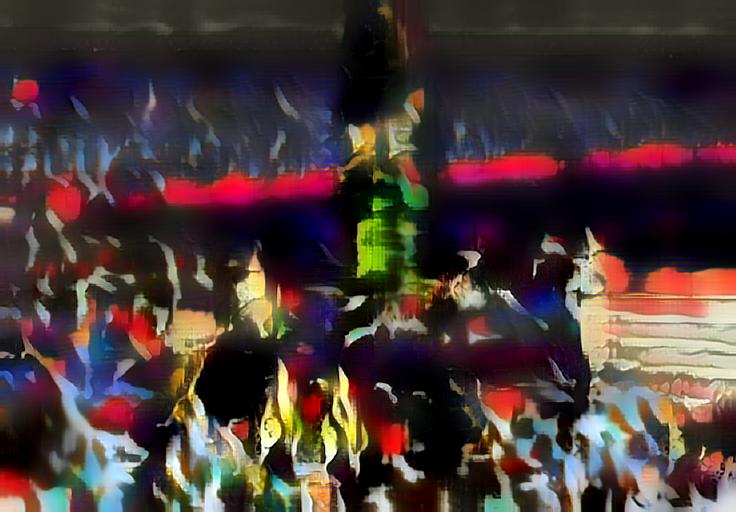} &
		\includegraphics[width=0.19\linewidth, height=0.185\linewidth]{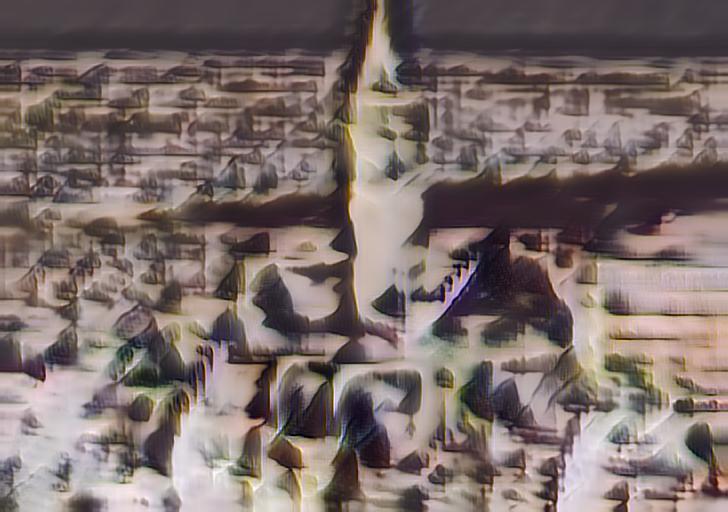} \\
		Input &  E1+D1 & E2+D2 &  E1+D2 & E2+D1 \\
	\end{tabular}
	\vspace{-0.5em}
	\caption{Examples of the exclusive collaboration phenomenon on two different encoder-decoder collaborative relationships: image reconstruction of WCT~\cite{li2017universal} (Row $1$) and style transfer of AdaIN~\cite{Huang-2017-arbitrary} (Row $2$). Column $1$ is the input, the other four columns show the outputs using different encoder-decoder combinations. If the two encoder-decoder pairs (E1-D1, E2-D2) are trained independently, the encoder can only work with its \emph{matching} decoder.}
	\label{fig:collabration}
	\vspace{-1.5em}
\end{figure}

Meanwhile, recent years have witnessed rapid development in the area of model compression~\cite{han2015deep,hinton2015distilling,li2017pruning,he2017channel}, which aims at reducing the parameters of a large CNN model without considerable performance loss.
Despite the progress, most model compression methods only focus on high-level tasks, \eg, classification~\cite{han2015learning,wen2016learning,wang2017structured,rastegari2016xnor} and detection~\cite{ZhaZouHeSun16,he2017channel}. 
Compressing models for low-level vision tasks is still less explored. 
Knowledge distillation (KD)~\cite{bucilua2006model,ba2014deep,hinton2015distilling} is a promising model compression method by transferring the knowledge of large networks (called teacher) to small networks (called student), where the knowledge can be softened probability (which can reflect the inherent class similarity structure known as \emph{dark knowledge}) or sample relations (which can reflect the similarity structure among different samples)~\cite{liu2019knowledge,park2019relational,yu2019learning,peng2019correlation}. This knowledge works as extra information on top of the one-hot labels and hence can boost the student's performance. However, this extra information is mainly \emph{label-dependent}, thus hardly applicable to low-level tasks. What is the dark knowledge in low-level vision tasks (\eg, neural style transfer) remains an open question. Meanwhile, encoder-decoder based models are extensively employed in neural style transfer, where the decoder is typically trained via the knowledge of encoder. Notably, they together construct an \emph{exclusive collaborative relationship} in the stylization process, shown in Fig.~\ref{fig:collabration}. Since the decoder $D$ is trained to exclusively work with the encoder $E$, if another encoder $E'$ can also work with $D$, it means $E'$ can functionally play the role of $E$. Based on this idea, we propose a new knowledge to distill the deep models in neural style transfer: the collaborative relationship between the encoder and decoder.

Given a redundant large encoder (\eg, VGG-19), we propose a two-step compression scheme: First, train a collaborator network for the encoder, namely, the decoder in our context; second, replace the large encoder with a small encoder, then train the small encoder with the collaborator \emph{fixed}. Since the small encoder typically has fewer channels, its output feature has a smaller dimension than that of the large encoder. Therefore, the small network cannot directly work with the collaborator. To resolve this, we propose to restrict the student to learn a linear embedding of the teacher's output, so that the teacher's output can be reconstructed through a simple linear combination of the student's output before being fed into the collaborator.

Notably, we do not restrict the specific collaboration form in our approach. In this paper, we will show it can be applied to two different state-of-the-art universal style transfer schemes: WCT~\cite{li2017universal} (where the collaboration is image reconstruction) and AdaIN~\cite{Huang-2017-arbitrary} (where the collaboration is style transfer). 
The main contributions of this work are:
\begin{itemize}
  \setlength\itemsep{0em}
  \item We propose a new knowledge distillation method for universal neural style transfer. The exclusive collaborative relationship between the encoder and its decoder is identified as a new kind of knowledge, which can be applied to different collaborative relationships.
  \item To resolve the feature dimension mismatch problem between the student and teacher networks in our algorithm, we propose to restrict the student to learn linear embedding of the teacher's output, which also acts as a regularizer to fuse more supervision into the middle layers of the student so as to boost its learning.
  \item Extensive experiments show the merits of our method with different stylization frameworks (WCT~\cite{li2017universal}, AdaIN~\cite{Huang-2017-arbitrary}, and Gatys~\cite{GatysTransfer-CVPR2016}), $15.5\times$ parameter reduction with comparable to even better visual effect. Especially, on WCT, the compressed models enable us to conduct ultra-resolution ($40$ megapixels) universal style transfer for the first time on a single 12GB GPU.
\end{itemize}



\section{Related Work}
\vspace{0.5em}
\noindent \textbf{Style transfer.}~Prior to the deep learning era, image style transfer is mainly tackled by non-parametric sampling~\cite{Efros1-ICCV1999}, non-photorealistic rendering~\cite{gooch2001non,strothotte2002non} or image analogy~\cite{Hertz-2001-analogy}. 
However, these methods are designed for some specific styles and rely on low-level statistics. 
Recently, Gatys \etal~\cite{GatysTransfer-CVPR2016} propose the neural style transfer, which employs deep features from the pre-trained VGG-19 model~\cite{VGG-2014} and achieves the stylization by matching the second-order statistics between the generated image and given style image. 
Numerous methods have been developed to improve the visual quality~\cite{MrfTransfer-CVPR2016,Wang-2016-highres,sanakoyeu2018style,zhang2019multimodal}, speed~\cite{li2016precomputed,Texturenet-ICML2016,Perceptual-ECCV2016,wei2020video,li2018learning}, user controls~\cite{liao2017visual,HistogramLoss-2017,Gatys2016-control}, style diversities~\cite{GoogleMultiTexture-2016,MSRA-2017-stylebank,Me-2017-diversified,Huang-2017-arbitrary,li2017universal}. 
However, one common limitation of all those neural network based approaches is that they cannot handle the ultra-resolution content and style images given limited memory.
Some approaches~\cite{Texturenet-ICML2016,Perceptual-ECCV2016,sanakoyeu2018style} achieve high-resolution stylization results (up to $10$ megapixels, \eg, 3000$\times$3000 pixels) by learning a small feedforward network for a specific style example or category, but they do not generalize to other unseen styles.
In contrast, our goal is to realize the ultra-resolution image style transfer for \emph{universal} styles with \emph{one} model only.

\vspace{0.400em}
\noindent \textbf{Model compression.}~Model compression and acceleration have also attracted much attention recently, which aim to obtain a smaller and faster model without a considerable compromise in performance.
Existing methods broadly fall into five categories, \ie, low-rank decomposition~\cite{denton2014exploiting,jaderberg2014speeding,lebedev2014speeding,ZhaZouHeSun16}, pruning~\cite{lecun1990optimal,han2015learning,han2015deep,li2017pruning,wen2016learning,he2017channel,wang2017structured,wang2019structured}, quantization~\cite{courbariaux2016binarized,rastegari2016xnor,zhou2016dorefa,hubara2017quantized}, knowledge distillation~\cite{bucilua2006model,ba2014deep,hinton2015distilling,zagoruyko2016paying} and compact architecture redesign or search~\cite{iandola2016squeezenet,howard2017mobilenets,sandler2018mobilenetv2,zhang2017shufflenet,ma2018shufflenet,tan2019efficientnet,elsken2019neural}. 
However, these methods are mainly explored in high-level vision tasks, typically the classification and detection. 
Few approaches have paid attention to low-level vision tasks such as style transfer, where many methods are also limited by the massive model size of CNNs. 
Unlike the CNN compression for high-level vision where it only needs to maintain the global semantic information of features to retain accuracies, the extra challenge of model compression for low-level vision may be how to maintain the local structures, \eg, local textures and color diversity in style transfer.

In this work, we develop a deeply-supervised knowledge distillation method to learn a much smaller model from pre-trained redundant VGG-19~\cite{VGG-2014}. 
The compressed model enjoys more than $15\times$ parameter and computation reduction. 
More importantly, the decrease of model size enables universal style transfer on ultra-resolution images.
To our best knowledge, only one recent work~\cite{junginger2018unpaired} employs GAN~\cite{goodfellow-2014-GAN} to learn unpaired style transfer network on ultra-resolution images. 
However, they achieve this by working on image \emph{subsamples} and then merging them back to a whole image. In contrast, our method fundamentally reduces the model complexity, which can directly process the \emph{whole} image.

\begin{figure*}[t]
    \centering
	\includegraphics[width=\linewidth]{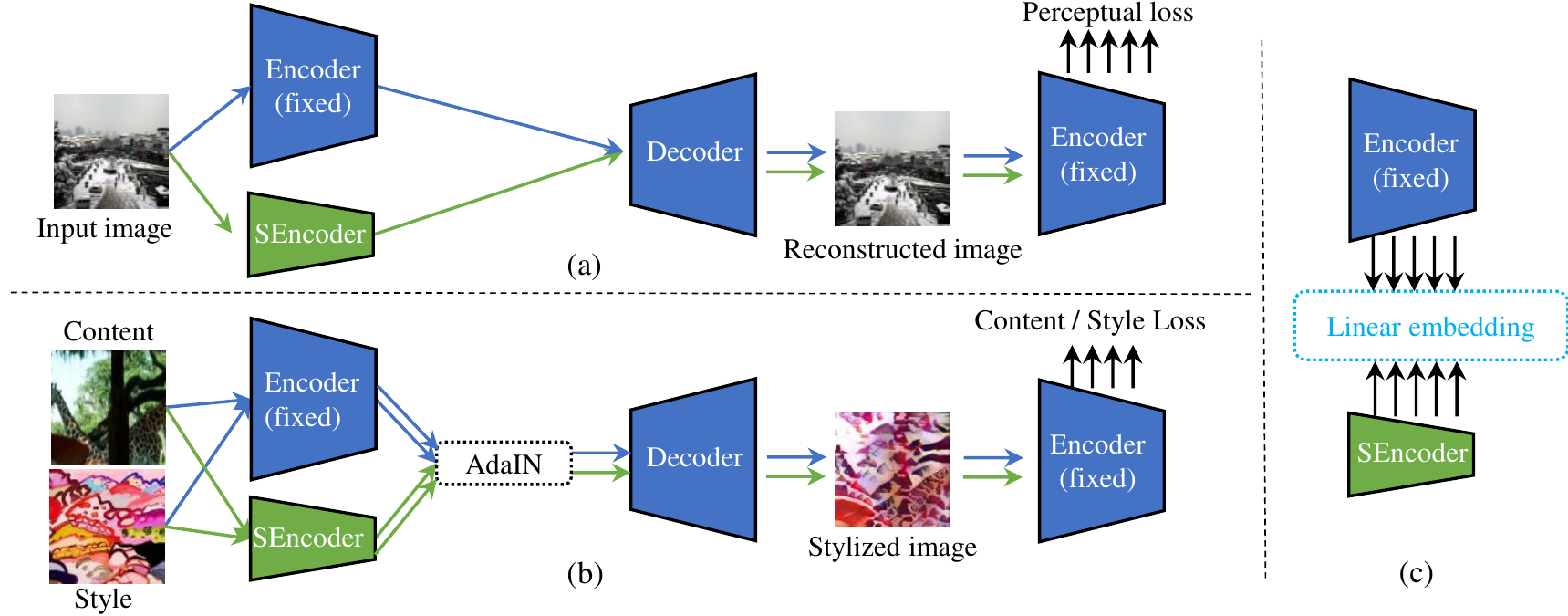} \\
	\vspace{-0.5em}
   	\caption{Illustration of the proposed Collaborative Distillation framework (best viewed in color). (a) and (b) depict two kinds of the encoder-decoder collaborative relationship for universal neural style transfer: image reconstruction for WCT~\cite{li2017universal} and style transfer for AdaIN~\cite{Huang-2017-arbitrary}, respectively. {\color{blue}{Blue}} arrows show the forward path when training the collaborator network (namely, the decoder). {\color{green}{Green}} arrows show the forward path when the small encoder (``SEncoder") is trained to functionally replace the original encoder (``Encoder"). (c) shows the proposed linear embedding scheme to resolve the feature size mismatch problem and infuse more supervision into the middle layers of the small encoder.}
	\label{fig:system}
\end{figure*}

\section{Proposed Method}
\subsection{Collaborative distillation}
Style-agnostic stylization methods typically adopt an encoder-decoder scheme to learn deep representations for style rendering and then invert them back into the stylized images.
Since the style information is not directly encoded in the model, the encoder part needs to be expressive enough to extract informative representations for universal styles.
Existing methods~\cite{GatysTransfer-CVPR2016,Chen-2016-swap,Huang-2017-arbitrary,li2017universal} commonly choose VGG-19~\cite{VGG-2014} as the encoder considering its massive capacity and hierarchical architecture.
As for the decoder, depending on different stylization schemes, it can have different collaborative relationships with the encoder. Two state-of-the-art arbitrary style transfer approaches, WCT~\cite{li2017universal} and AdaIN~\cite{Huang-2017-arbitrary}, are discussed here. (i) For WCT, the stylization process is to apply Whitening and Coloring Transform~\cite{WCT-2016} to the content features using the second-order statistics of style feature. Then the transformed content feature is inverted to an image by the decoder. Hence, the decoder training is not directly involved with stylization. The collaborative relationship in WCT is essentially image reconstruction. (ii) For AdaIN, unlike WCT, its decoder training is directly involved in the stylization. Two images (content and style) are fed into the encoder, then in the feature space, the content feature is rendered by the statistics (mean and variance) of the style feature. Finally, the decoder inverts the rendered content feature back to stylized images. The stylized images are supposed to be close to the content (or style) in terms of content (or style) distance. Therefore, the collaborative relationship for AdaIN is style transfer.

Despite paradigm difference for the above two schemes, they are both encoder-decoder based, and the decoder is trained through the knowledge of the encoder. This means, during the training of the decoder, the knowledge of the encoder is leaked into the decoder. Presumably and confirmed empirically, the decoder $D$ can only work with its \emph{matching} encoder $E$ like a nut with its bolt. For another encoder $E'$, even if it has the same architecture as $E$, $D$ and $E'$ cannot work together (see Fig.~\ref{fig:collabration}). This exclusivity shows the decoder has some inherent information specific to its encoder. If we can find a way to make the network $E'$ compatible with $D$ too, it means $E'$ can functionally replace the original encoder $E$. If $E'$ is meanwhile much smaller than $E$, then we achieve the model compression goal.
Based on this idea, we propose a new distillation method specific to NST, named \emph{Collaborative Distillation}, consisting of two steps. 

For the first step, based on the task at hand, we train a collaborator network for the large encoder. As shown in Fig.~\ref{fig:system}(a), for WCT~\cite{li2017universal}, the decoder is trained to invert the feature to be as faithful to the input image as possible (\ie, image reconstruction), where both the pixel reconstruction loss and the perceptual loss~\cite{Perceptual-ECCV2016} are employed,
\begin{equation}
  \mathcal{L}^{(k)}_{r} = \| \mathcal{I}_{r} - \mathcal{I}_{o} \|^2_2 + \lambda_p  \sum_{i=1}^k  \|\mathcal{F}^{(i)}_r - \mathcal{F}^{(i)}_o\|^2_2,
  \label{eqn:pixel_perceptual_loss}
\vspace{-0.5em}
\end{equation}
where $k\in \{1,2,3,4,5\}$ denotes the $k$th stage of VGG-19; $\mathcal{F}^{(i)}$ denotes the feature maps of the \verb+ReLU_i_1+ layer; $\lambda_p$ is the weight to balance the perceptual loss and pixel reconstruction loss; $\mathcal{I}_{\text{o}}$ and $\mathcal{I}_{r}$ denote the original image and reconstructed image, respectively. For AdaIN~\cite{Huang-2017-arbitrary}, the decoder is involved in style transfer directly. Thus, its decoder loss is made up of both the content loss and the style loss,
\begin{equation}
  \mathcal{L}_{\text{st}} = \| \mathcal{F}^{(4)}_{\text{st}} - \mathcal{F}^{(4)}_c \|^2_2 +  \lambda_s \sum^4_{i=1} \| \mathcal{G}^{(i)}_{\text{st}} - \mathcal{G}^{(i)}_{s}\|^2_2,
  \label{eqn:style_transfer_loss}
\vspace{-0.5em}
\end{equation}
where $\mathcal{G}$ is the Gram matrix to describe style~\cite{GatysTexture-NIPS2015,GatysTransfer-CVPR2016}, $\lambda_s$ is the weight to balance the style loss and the content loss; subscript ``st'', ``$c$'', and ``$s$'' represent the stylized image, content image, and style image, respectively.

After we have the collaborator, the second step of our algorithm is to replace the original encoder $E$ with a small encoder $E'$. For simplicity, in this work we use $E'$ with the same architecture of $E$ but fewer filters in each layer. We expect that the small encoder $E'$ can functionally equivalent to the original encoder $E$, as shown in (a) and (b) of Fig.~\ref{fig:system}. Similar to the first step, the collaborator network loss, denoted as $\mathcal{L}_{\text{collab}}$, can take different forms depending on the specific collaboration tasks. In the context of this work, $\mathcal{L}_{\text{collab}} = \mathcal{L}_{r}$ for WCT and $\mathcal{L}_{\text{collab}} = \mathcal{L}_{\text{st}}$ for AdaIN.

\subsection{Linear embedding}
In the proposed collaborative distillation method, the small encoder is connected with the decoder network. In their interface comes a \emph{feature size mismatch} problem. Specifically, if the original encoder outputs a feature of size $C \times H \times W$, and thus the input of the decoder is also supposed to be of size $C \times H \times W$. However, because the small encoder has fewer filters, it will output a feature of size $C' \times H \times W$ $(C' < C)$, which cannot be accepted by the decoder. To resolve this, we first look at how the channel numbers play a role in the stylization process. As pioneered by Gatys~\cite{GatysTexture-NIPS2015,GatysTransfer-CVPR2016}, the style of an image is described by the gram matrix of the deep features extracted from VGG-19,
\begin{equation}
	\mathcal{G} = \mathcal{F} \cdot \mathcal{F}^T,
\vspace{-0.5em}
\end{equation}
where $\mathcal{F}$ is the deep feature of size $C \times HW$ extracted from certain convolutional layer of VGG-19; $\mathcal{G}$ denotes the Gram matrix of size $C\times C$; $T$ stands for matrix transpose. Since we aim to compress these features, \ie, they are regarded as redundant, it can be formulated as that $\mathcal{F}$ is a linear combination of some feature basis vectors in a lower dimension,
\begin{equation}
	\mathcal{F} = Q \cdot \mathcal{F'},
\vspace{-0.5em}
\end{equation}
where $Q$ is a transform matrix of size $C \times C'$, $\mathcal{F'}$ is the feature basis matrix of size $C' \times HW$, which can be viewed as the linear embedding of the original deep feature $\mathcal{F}$. Then it is easy to see that the Gram matrix $\mathcal{G'} = \mathcal{F'} \cdot \mathcal{F'}^T$ for the new feature $\mathcal{F'}$ has the same number of eigenvalues as the original Gram matrix $\mathcal{G}$. In other words, the style description power is maintained if we adopt $\mathcal{F'}$ in place of the original redundant $\mathcal{F}$. In the context of our method, $\mathcal{F}$ is the output of the original encoder, $\mathcal{F'}$ is the output of the small encoder. The transformation matrix $Q$ is learned through a fully-connected layer \emph{without non-linear activation function} to realize the linearity assumption. Hence, the linear embedding loss can be formulated as
\begin{equation}
	\mathcal{L}_{\text{embed}} = ||\mathcal{F} - Q \cdot \mathcal{F'}||_2^2.
	\label{eqn:linear_embedding_loss}
\vspace{-0.5em}
\end{equation}

One more step forward, the proposed solution above is not limited to the final output layer. For the middle layers of the small encoder, it can also find an application. Concretely, we apply the linear embedding to the other four middle layers (\verb+ReLU_k_1+, $k \in \{1, 2, 3, 4\}$) between the original encoder and the small encoder, as shown in Fig.~\ref{fig:system}(c). We have two motivations for this. First, in the proposed method, when the small encoder is trained with the SGD algorithm, its only gradient source would be the decoder, passed through the fully-connected layer $Q$. However, $Q$ does not have many parameters, so it will actually form an information bottleneck, slowing down the learning of the student. With these branches plugged into the middle layers of the network, they will infuse more gradients into the student and thus boost its learning, especially for deep networks that are prone to gradient vanishing. Second, in neural style transfer, the style of an image is typically described by the features of \emph{many} middle layers~\cite{GatysTransfer-CVPR2016,Huang-2017-arbitrary,li2017universal}. Therefore, adding more supervision to these layers is necessary to ensure that they do not lose much the style description power for subsequent use in style transfer.

To this end, the total loss to train the small encoder in our proposed algorithm can be summarized as
\begin{equation}
\mathcal{L}_{\text{total}} = \beta \sum^k_{i=1} \mathcal{L}_{\text{embed}} + \mathcal{L}_{\text{collab}},
\label{eqn:total_loss}
\vspace{-0.5em}
\end{equation}
where $\beta$ is the weight factor to balance the two loss terms.

\section{Experimental Results}
In this section, we will first demonstrate the effectiveness of our compressed VGG-19 compared with the original VGG-19 within the universal style transfer frameworks WCT~\cite{li2017universal}. Then we show the proposed collaborative distillation is not limited to one kind of collaborative relationship on AdaIN~\cite{Huang-2017-arbitrary}. Finally, to show the generality of the proposed approach, we also evaluate on the optimization-based style transfer using Gatys~\cite{GatysTransfer-CVPR2016}, where the collaborative relationship is the same as WCT, \ie, image reconstruction.
We first conduct comparisons in the maximal image resolution which the original VGG-19 can possibly handle (3000$\times$3000), then show some stylized samples in larger resolutions (\ie, ultra-resolutions) with the small models.
All experiments are conducted on one Tesla P100 12GB GPU, namely, given the same limited memory.
%
%
%

\vspace{0.400em}
\noindent\textbf{Evaluated compression method.}~Since there are few model compression methods specifically designed for low-level image synthesis tasks, we compare our method with a typical compression algorithm in the high-level image classification task, \ie, filter pruning (FP)~\cite{li2017pruning}. 
Specifically, we first apply FP to VGG-19 in classification to obtain a compressed model with the same architecture as ours.  
It is then fine-tuned on ImageNet~\cite{deng2009imagenet} to regain performance.
Finally, its decoder is obtained by optimizing the loss in (\ref{eqn:pixel_perceptual_loss}).

\begin{figure*}[t]
	\centering
	\small
	\renewcommand{\arraystretch}{0.3} 
	\begin{tabular}{c@{\hspace{0.005\linewidth}}c@{\hspace{0.005\linewidth}}c@{\hspace{0.005\linewidth}}c@{\hspace{0.005\linewidth}}c@{\hspace{0.005\linewidth}}c}
		\includegraphics[width = 0.19\linewidth, height=0.17\linewidth]{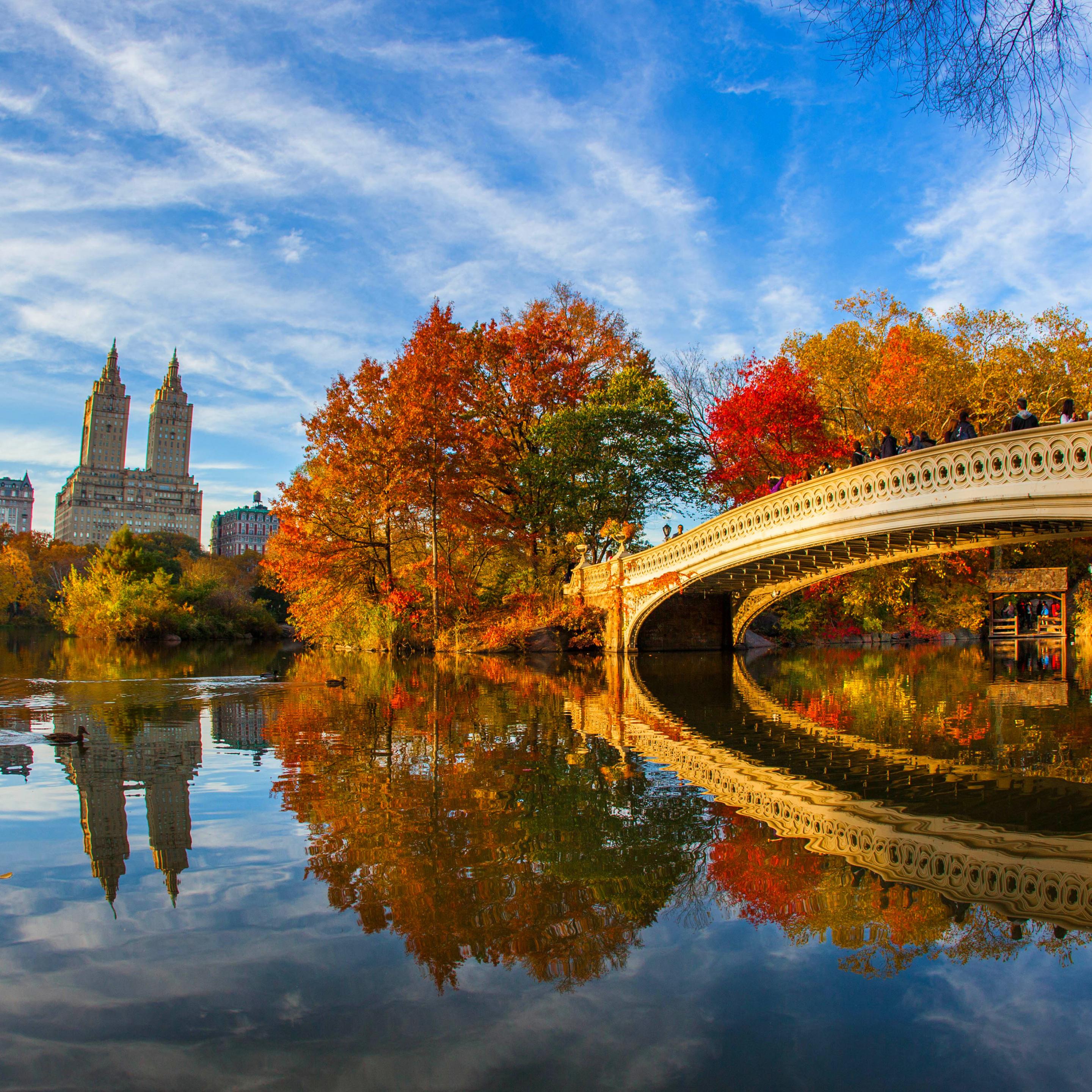} &
		\includegraphics[width = 0.19\linewidth, height=0.17\linewidth]{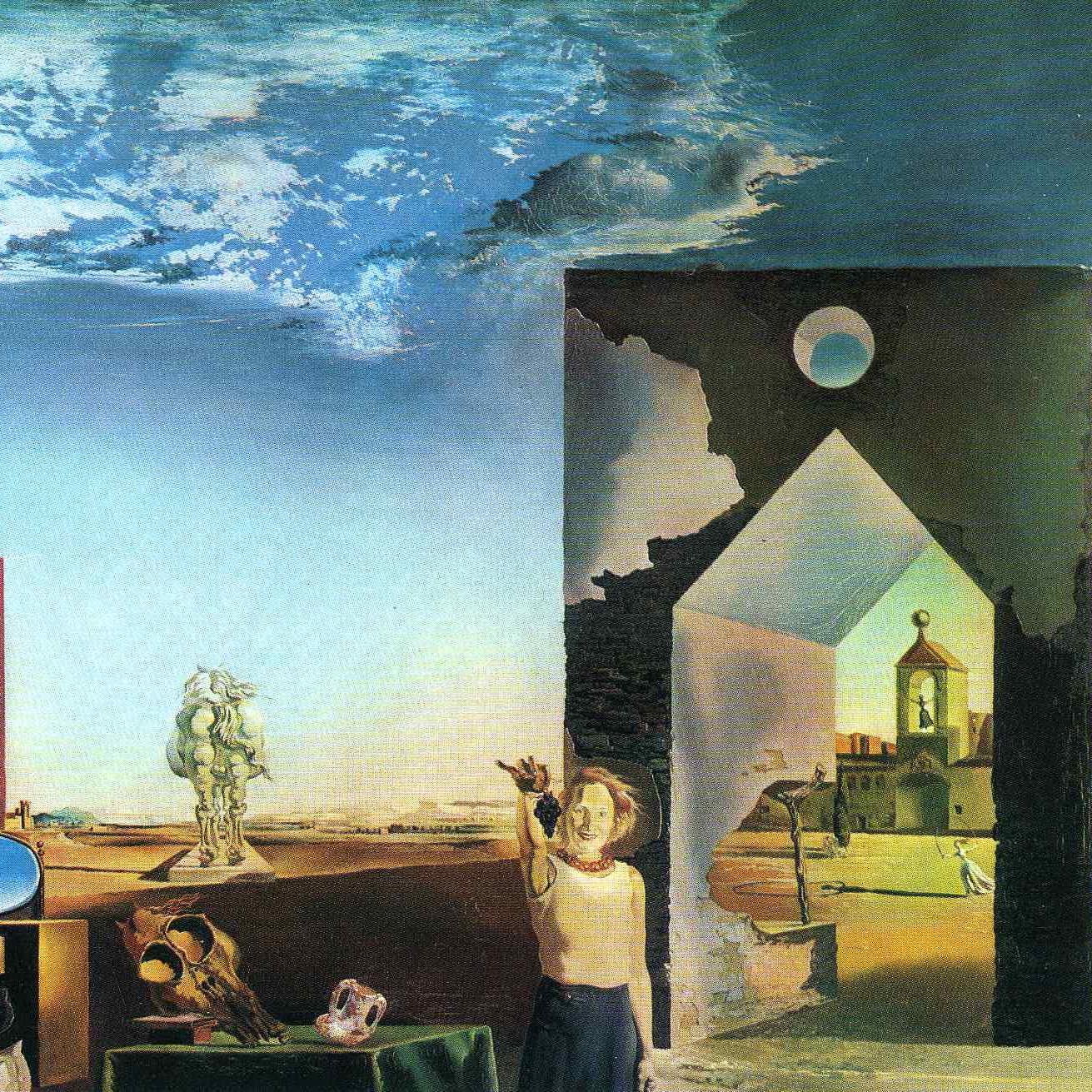} &
		\includegraphics[width = 0.19\linewidth, height=0.17\linewidth]{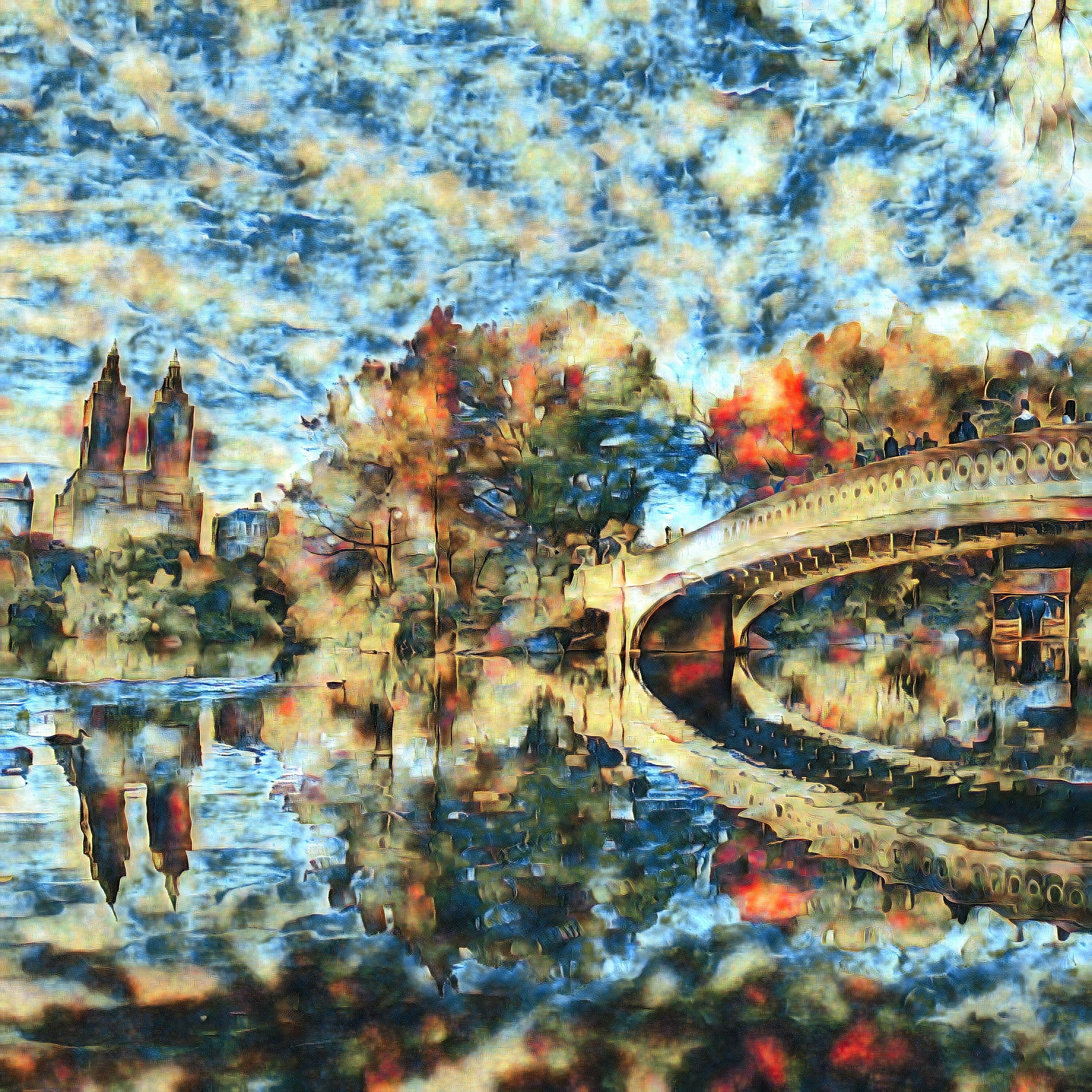} &
		\includegraphics[width = 0.19\linewidth, height=0.17\linewidth]{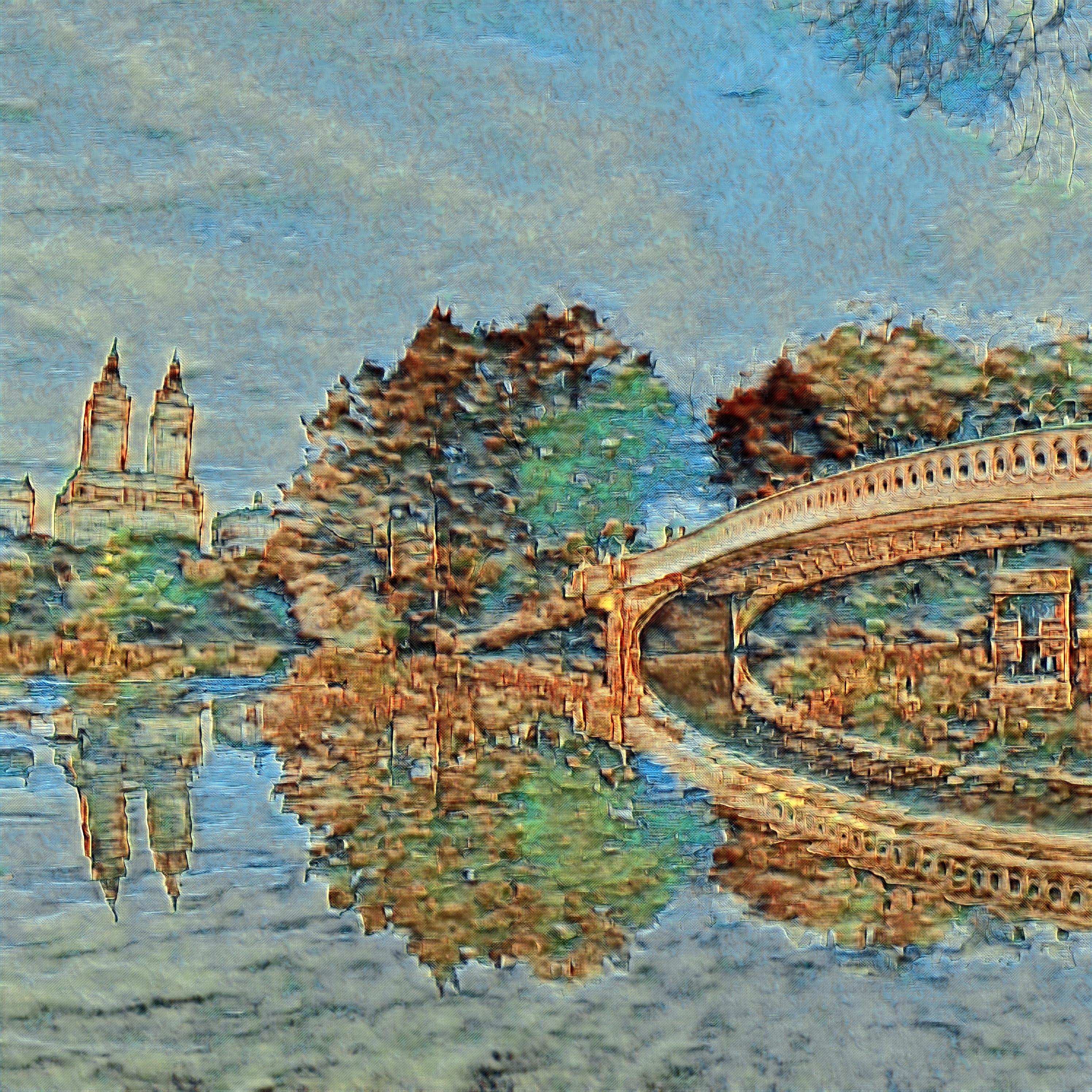} &
		\includegraphics[width = 0.19\linewidth, height=0.17\linewidth]{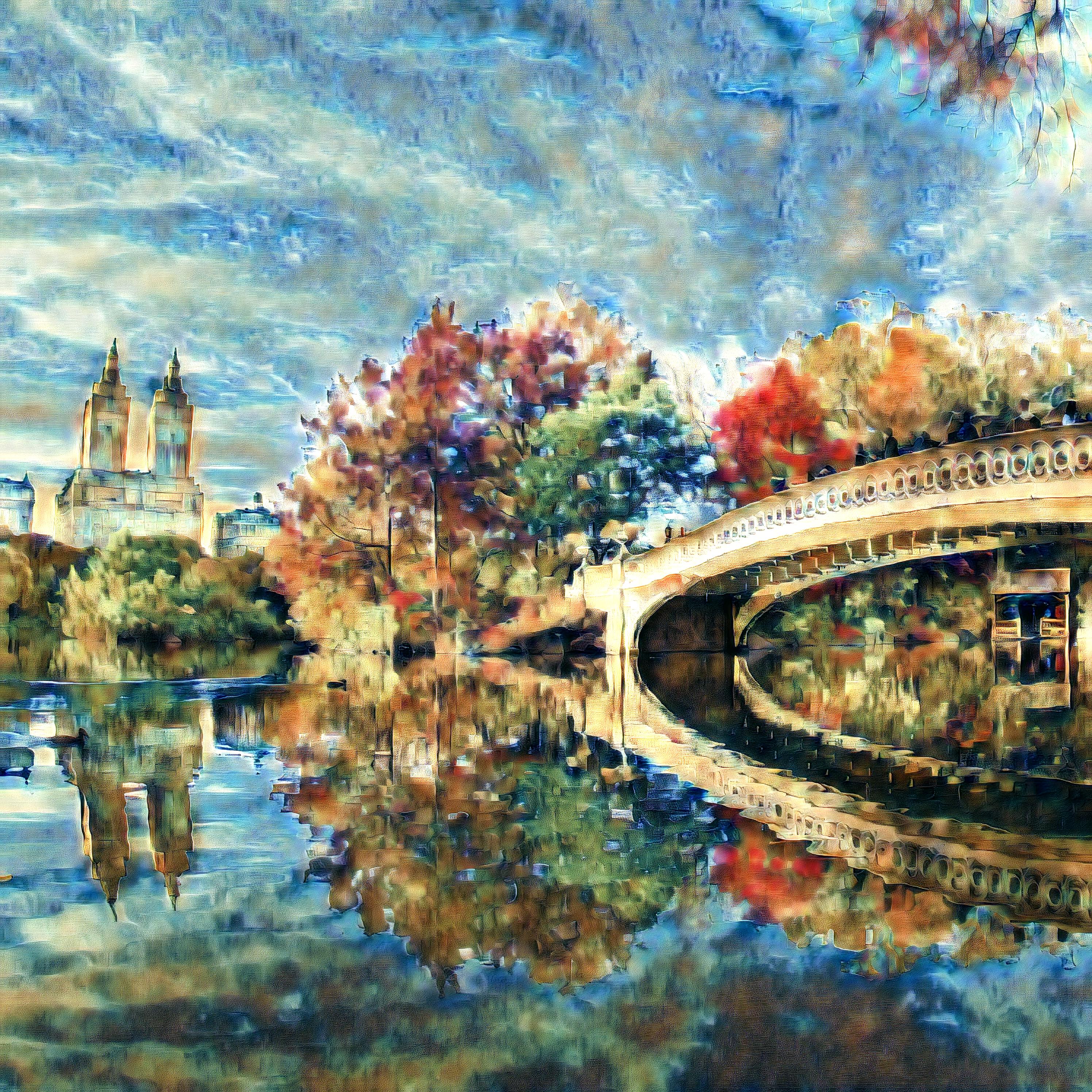} & \\
		\includegraphics[width = 0.19\linewidth, height=0.17\linewidth]{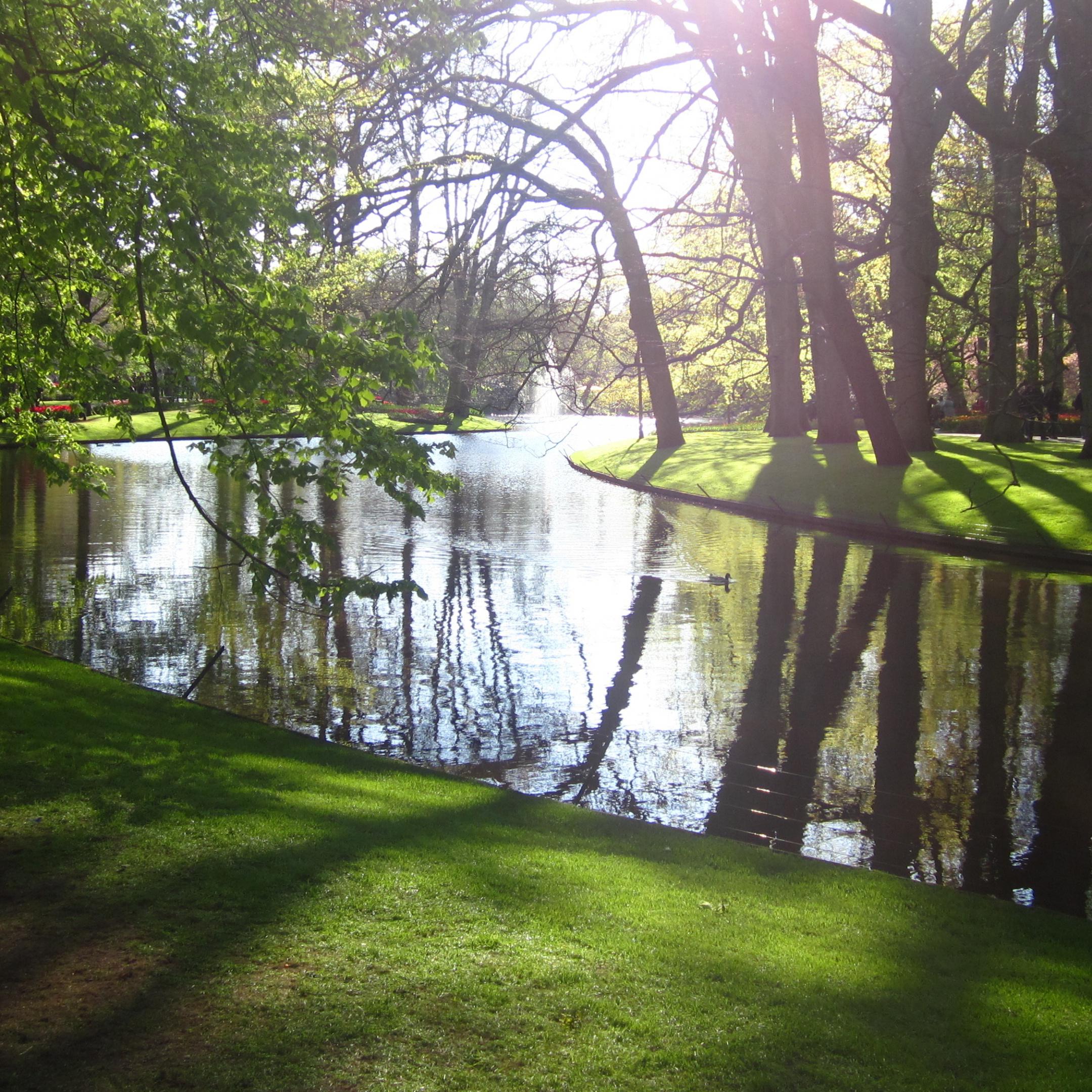} &
		\includegraphics[width = 0.19\linewidth, height=0.17\linewidth]{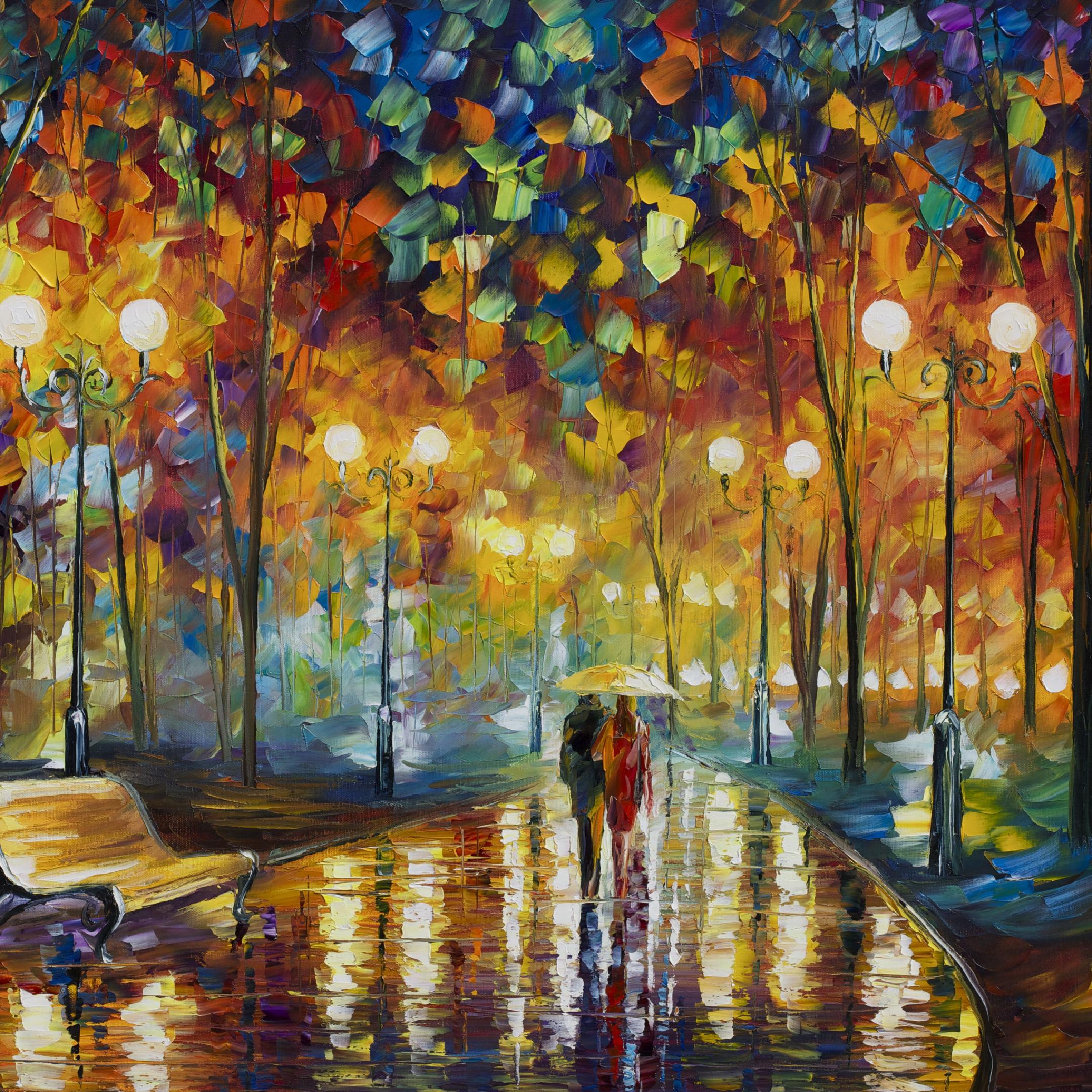} &
		\includegraphics[width = 0.19\linewidth, height=0.17\linewidth]{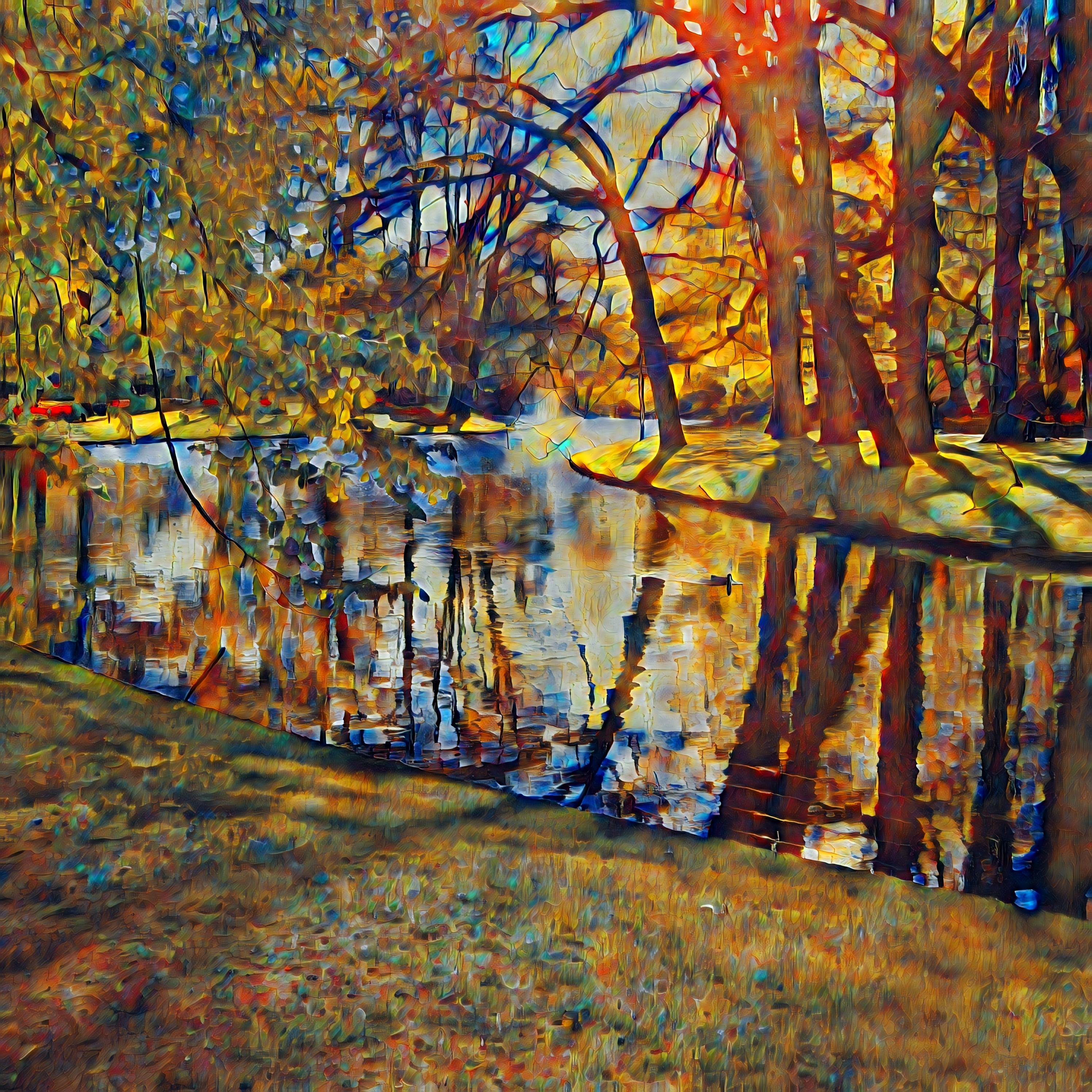} &
		\includegraphics[width = 0.19\linewidth, height=0.17\linewidth]{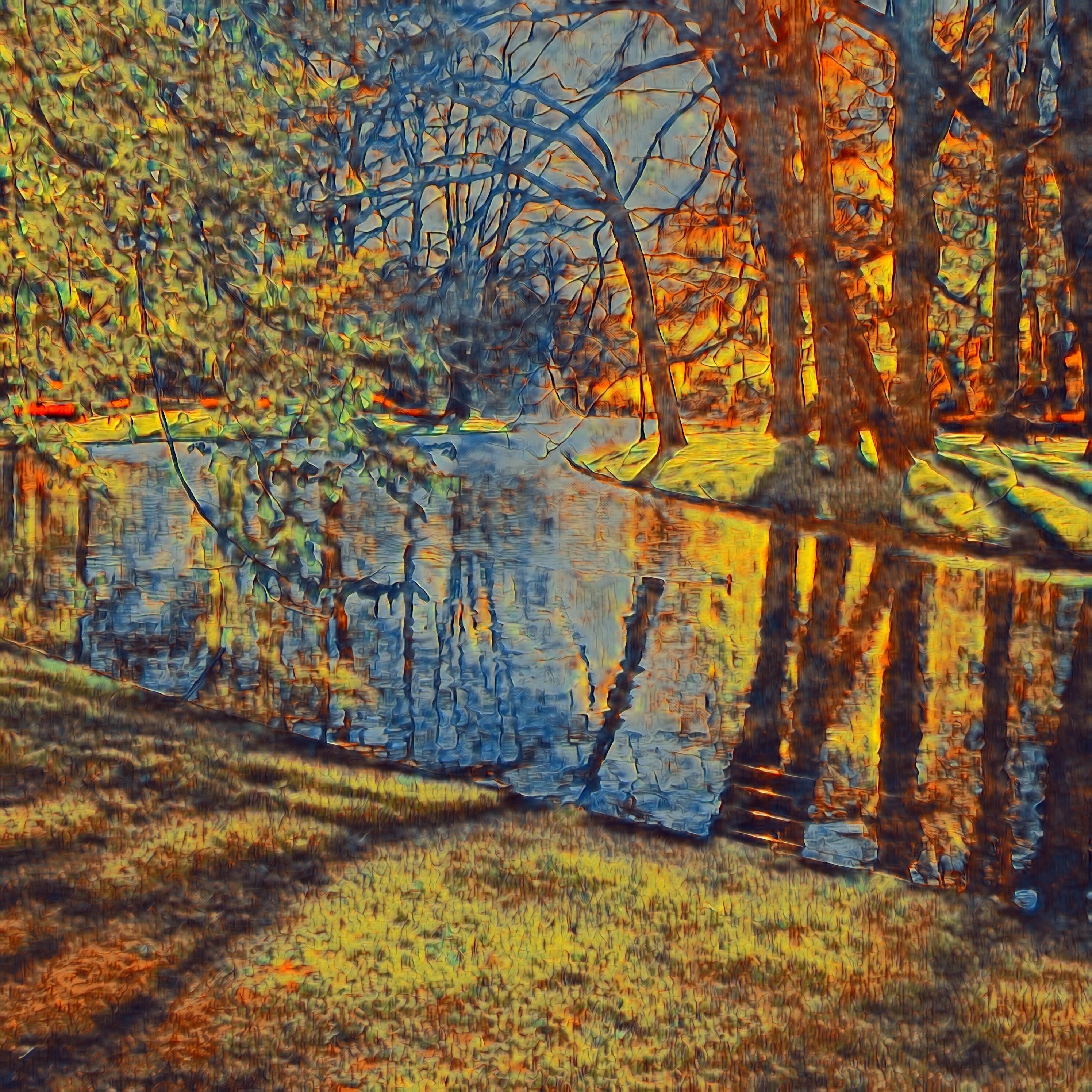} &
		\includegraphics[width = 0.19\linewidth, height=0.17\linewidth]{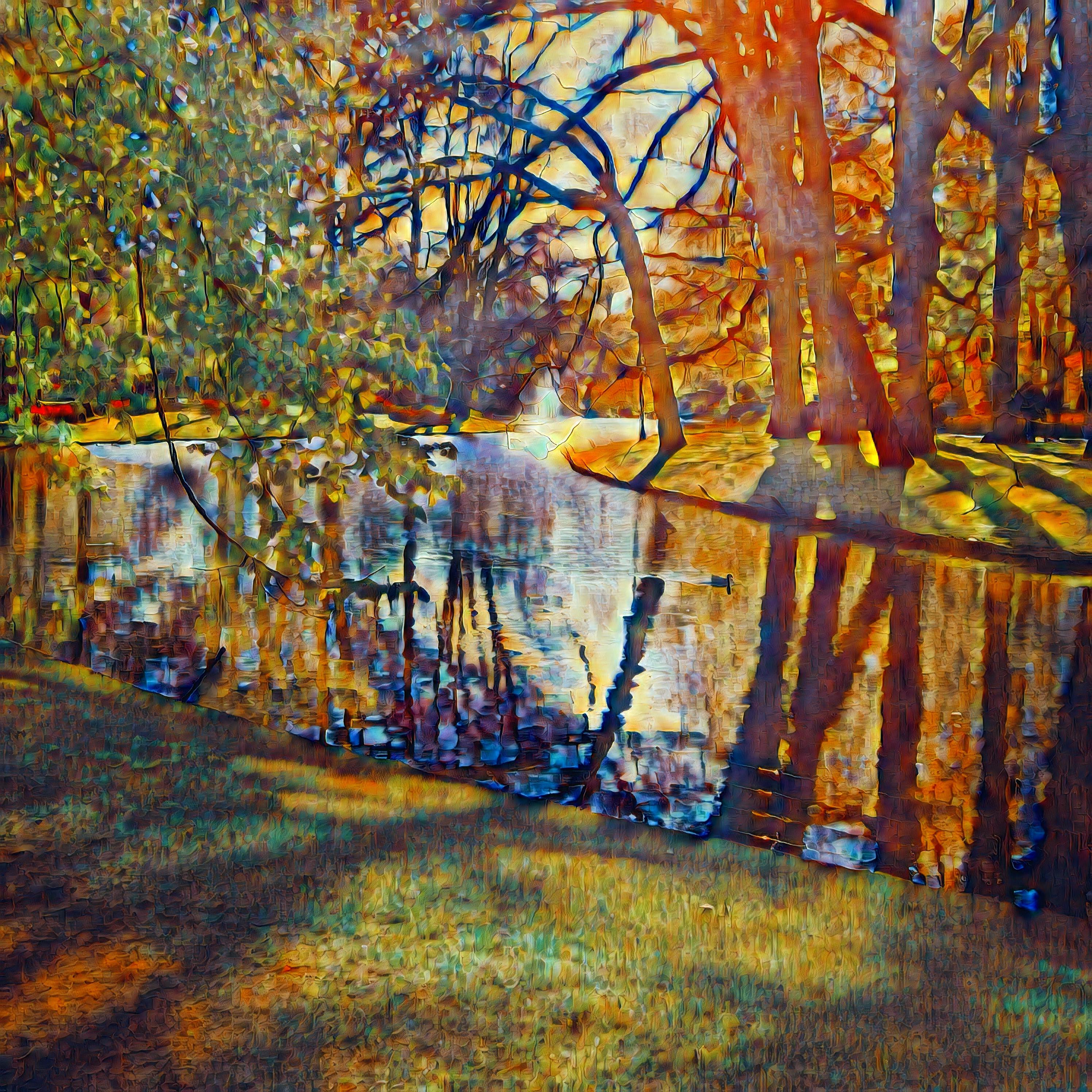} & \\
		\includegraphics[width = 0.19\linewidth, height=0.17\linewidth]{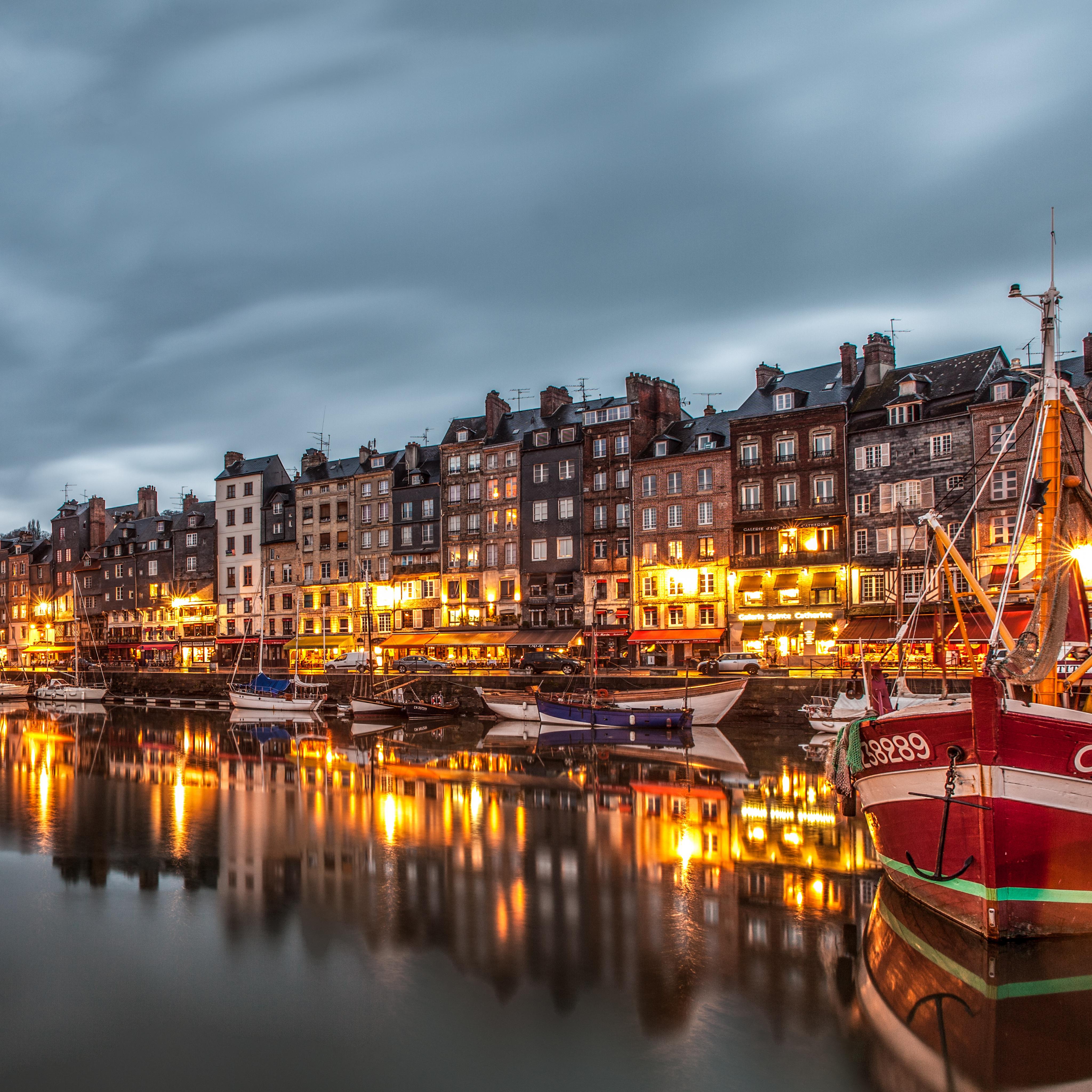} &
		\includegraphics[width = 0.19\linewidth, height=0.17\linewidth]{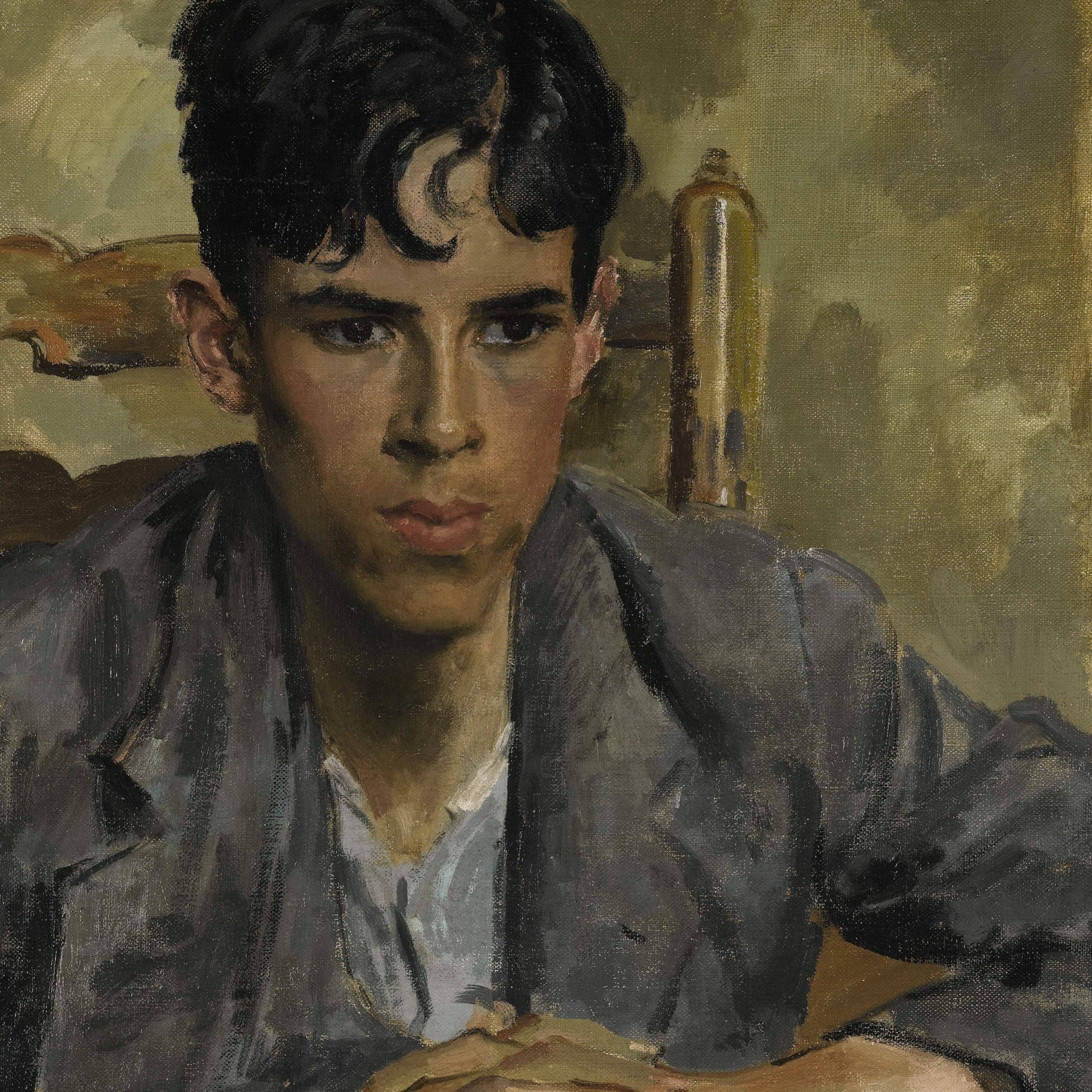} &
		\includegraphics[width = 0.19\linewidth, height=0.17\linewidth]{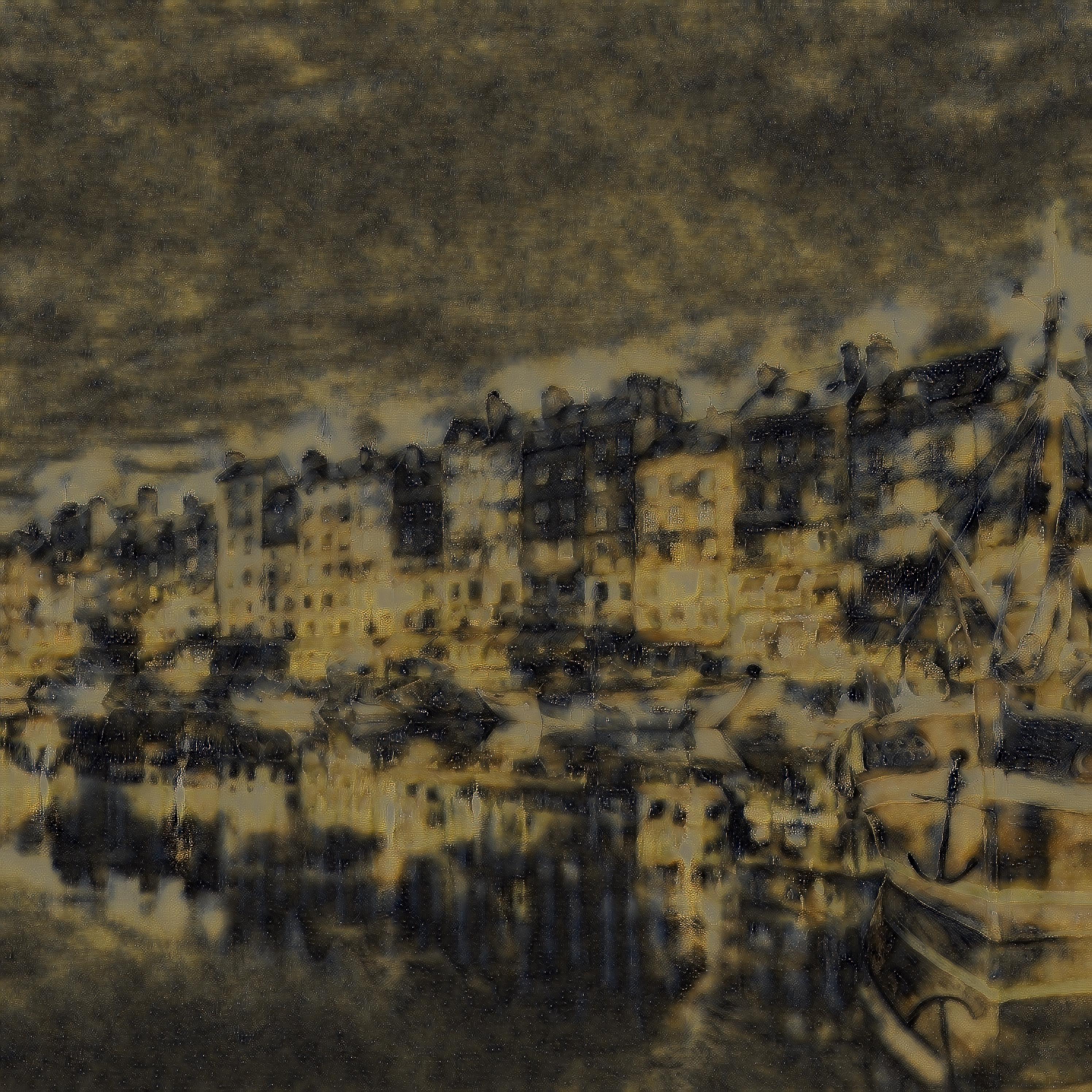} &
		\includegraphics[width = 0.19\linewidth, height=0.17\linewidth]{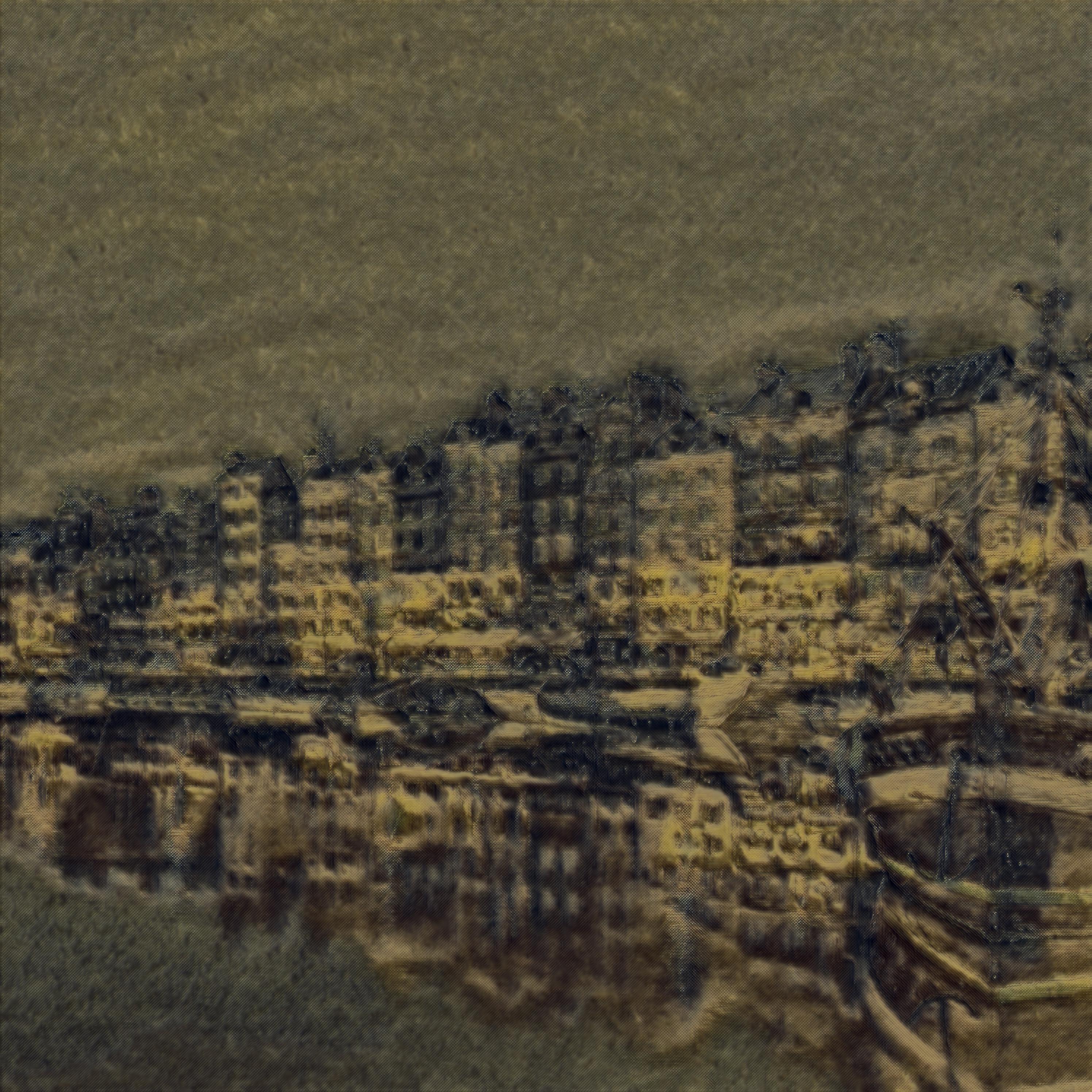} &
		\includegraphics[width = 0.19\linewidth, height=0.17\linewidth]{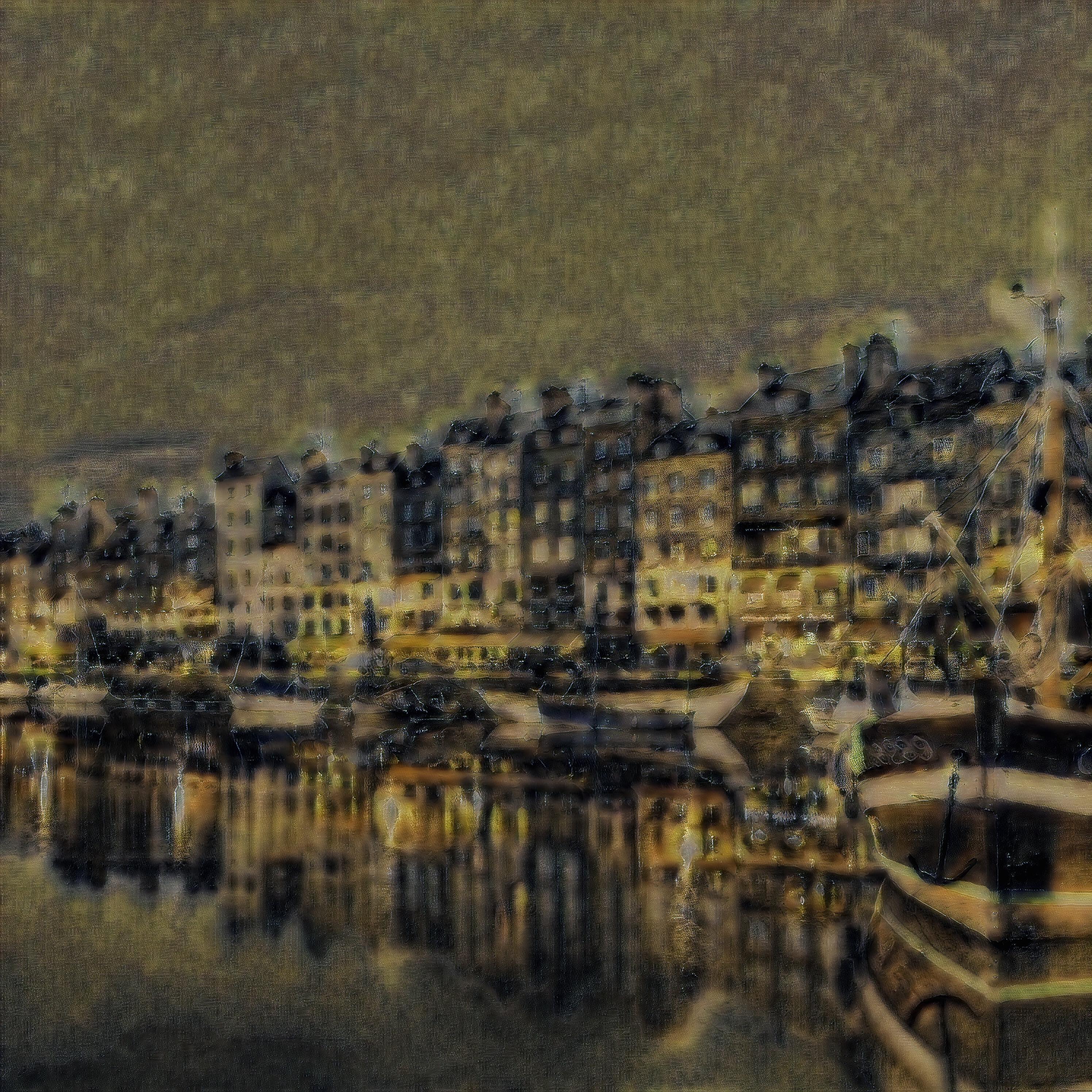} & \\
		\includegraphics[width = 0.19\linewidth, height=0.17\linewidth]{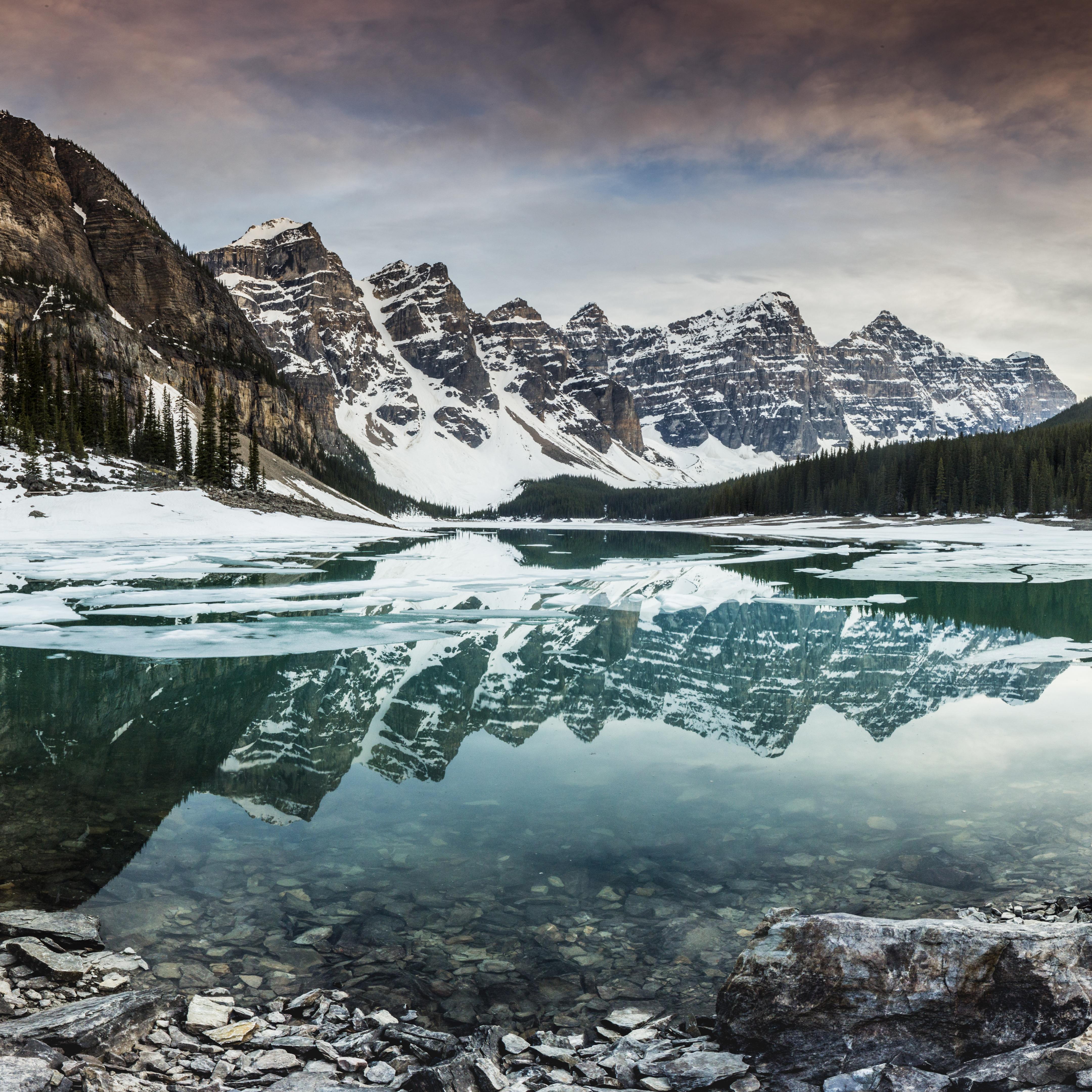} &
		\includegraphics[width = 0.19\linewidth, height=0.17\linewidth]{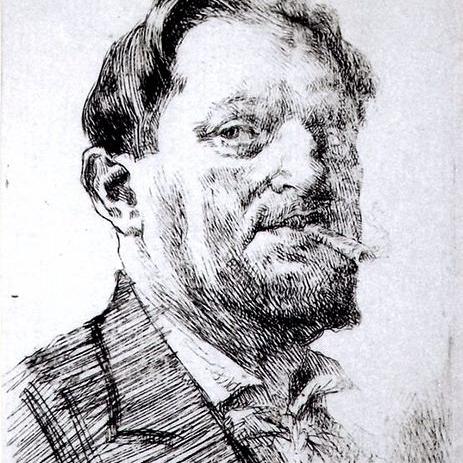} &
		\includegraphics[width = 0.19\linewidth, height=0.17\linewidth]{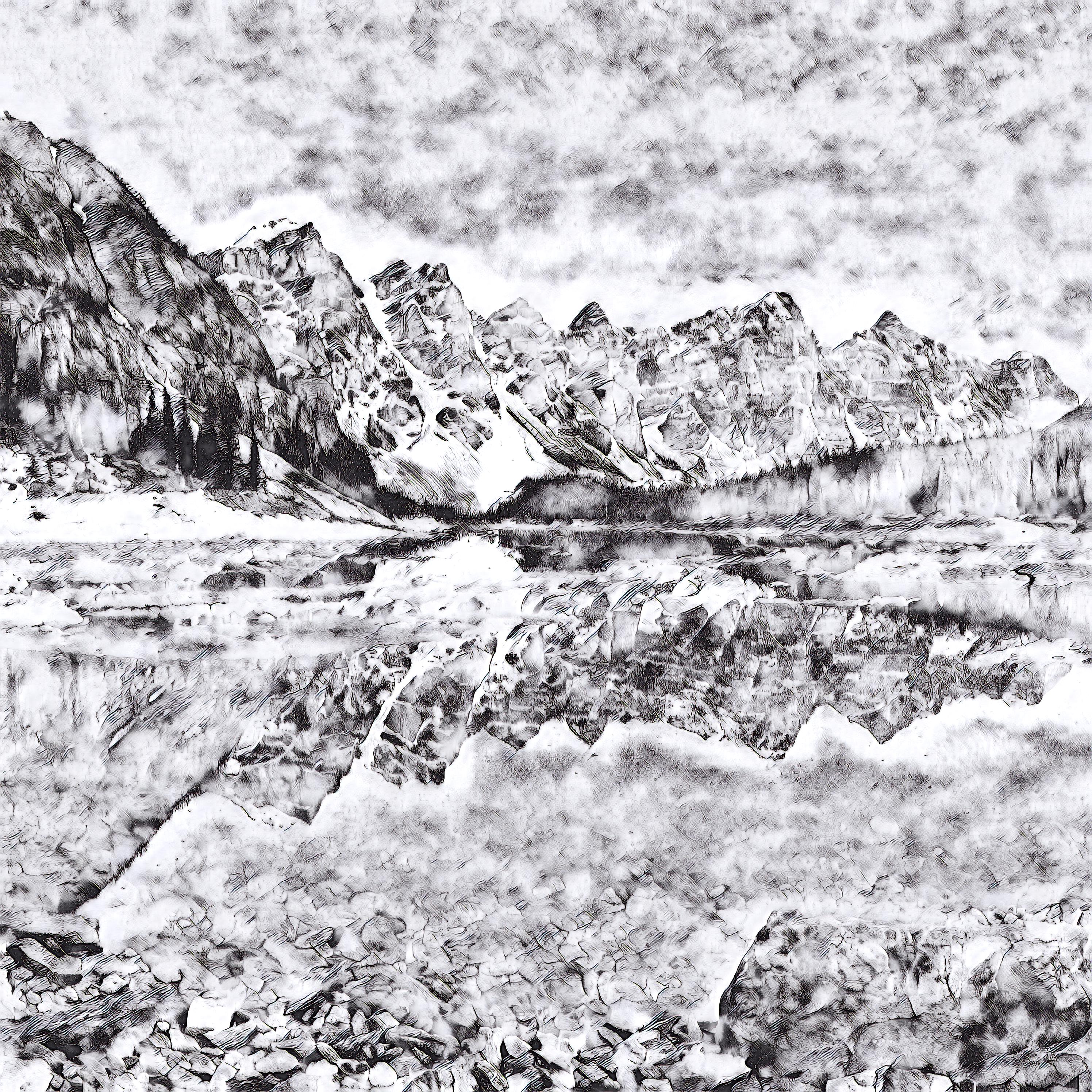} &
		\includegraphics[width = 0.19\linewidth, height=0.17\linewidth]{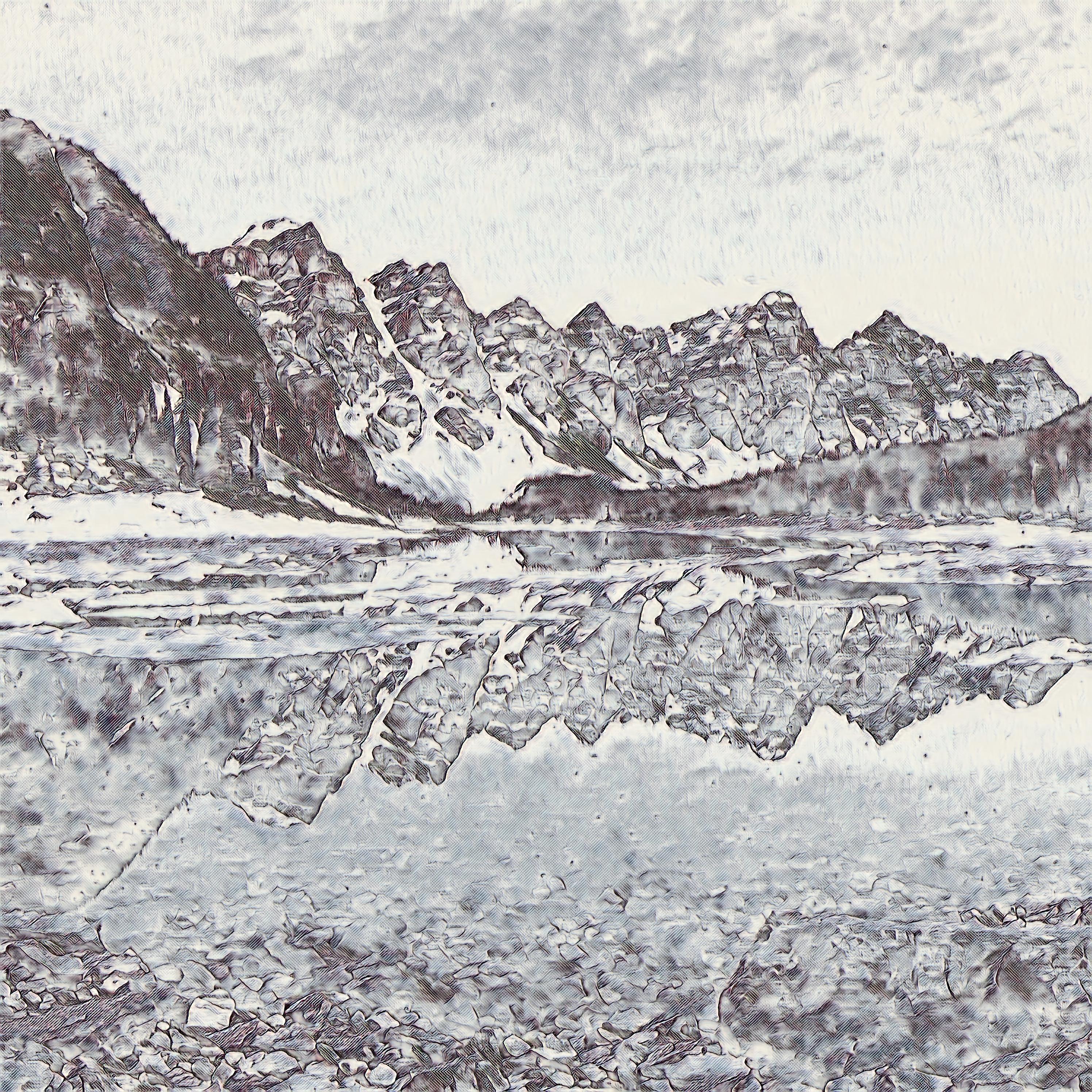} &
		\includegraphics[width = 0.19\linewidth, height=0.17\linewidth]{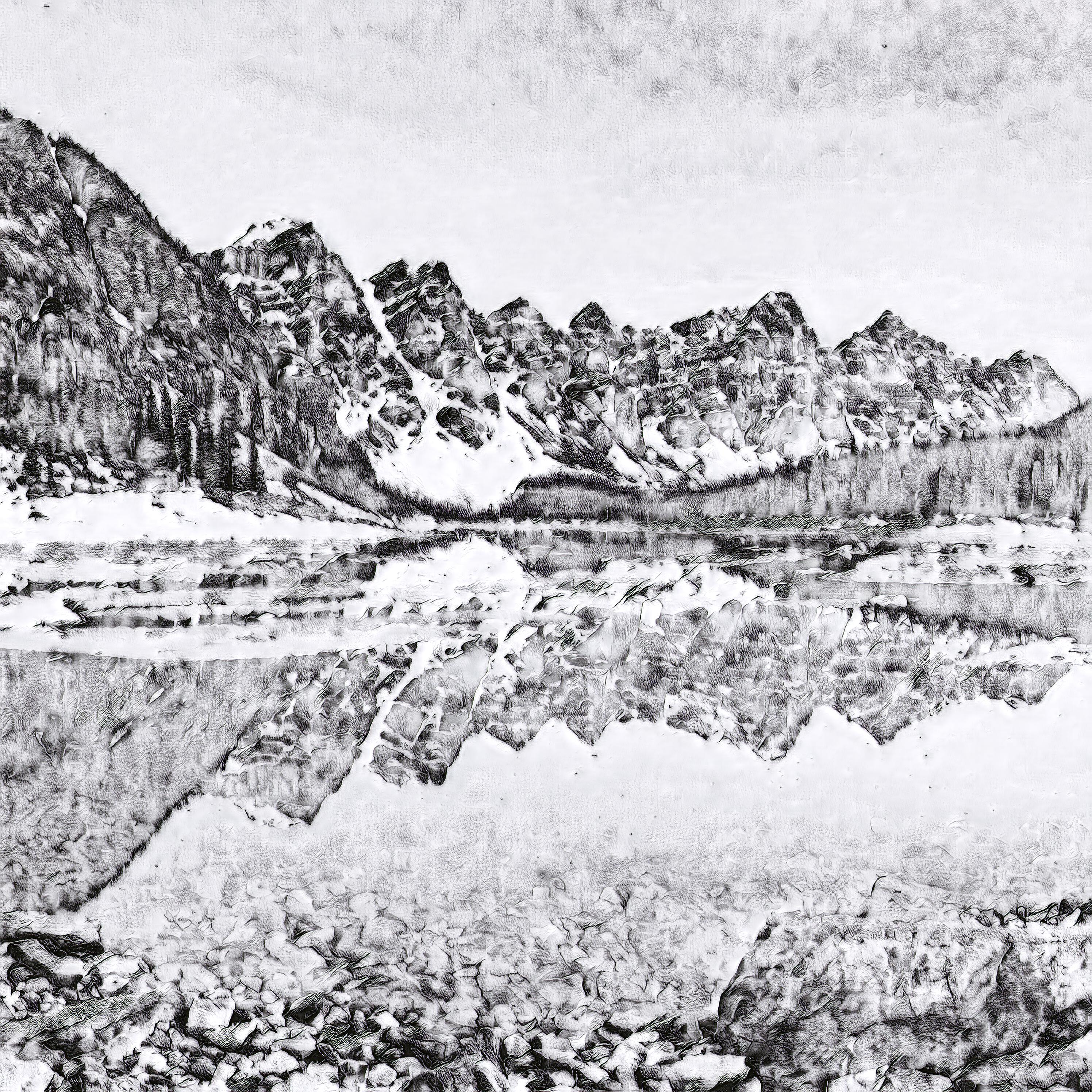} & \\
		{(a) Content} & {(b) Style} & {(c) Original} & {(d) FP} & {(e) Ours} \\
	\end{tabular}
  	\vspace{-0.5em}
  	\caption{Comparison of stylized images in 3000$\times$3000 produced by three different models, \ie, the original VGG-19, FP-slimmed VGG-19 and our compressed VGG-19 (best viewed in color and zoomed in).}
	\label{fig:wct_stylized_sample}
	\vspace{-2em}
\end{figure*}

\subsection{Style transfer on WCT}
\label{sec:exp_wct}
\vspace{-0.5em}
Since we need the original decoder as collaborator , we first train a decoder with the mirrored architecture of VGG-19 on the MS-COCO dataset~\cite{COCO-lin2014-microsoft} for image reconstruction.   
During the training, the encoder is \emph{fixed}, with $\lambda_p = 1$. 
We randomly crop $256\times256$ patches from $300\times300$ images as input.
Adam~\cite{kingma2014adam} is used as optimization solver with fixed learning rate $10^{-4}$, batch size $16$.
In WCT~\cite{li2017universal}, a cascaded coarse-to-fine stylization procedure is employed for best results, so the decoders of the $5$-stage VGG-19 (up to \verb+ReLU_k_1+, $k \in \{1,2,3,4,5\}$) are all trained.
Then an encoder-decoder network is constructed with loss~(\ref{eqn:total_loss}), in which $\beta$ is set to $10$.
The compressed encoder can be randomly initialized, but we empirically find using the largest filters (based on $L_1$-norms) from the original VGG-19 as initialization will help the compressed model converge faster, so we use this initialization scheme in all our experiments. 
After $20$-epoch training, we obtain the compressed encoders.  
Their mirrored decoders are trained by the same rule as the first step using the loss~(\ref{eqn:pixel_perceptual_loss}).

\begin{figure}[t]
	\centering
	\renewcommand{\arraystretch}{0.6} 
	\begin{tabular}{c@{\hspace{0.01\linewidth}}c@{\hspace{0.01\linewidth}}c@{\hspace{0.01\linewidth}}c}
		\includegraphics[width = .3\linewidth, height=0.7in]{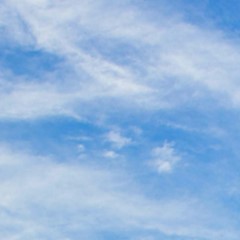} &
		\includegraphics[width = .3\linewidth, height=0.7in]{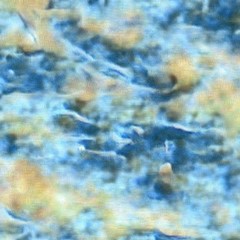} &
		\includegraphics[width = .3\linewidth, height=0.7in]{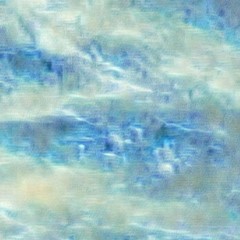} & \\
		\includegraphics[width = .3\linewidth, height=0.7in]{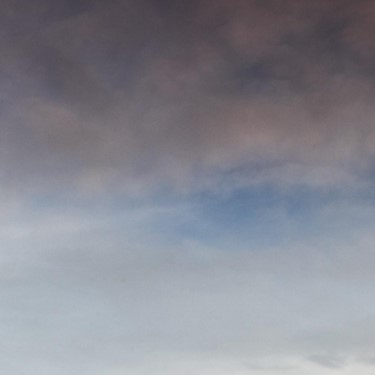} &
		\includegraphics[width = .3\linewidth, height=0.7in]{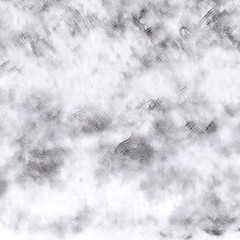} &
		\includegraphics[width = .3\linewidth, height=0.7in]{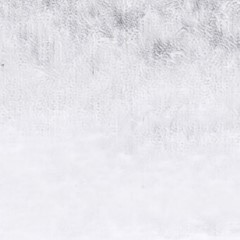} & \\
		\includegraphics[width = .3\linewidth, height=0.7in]{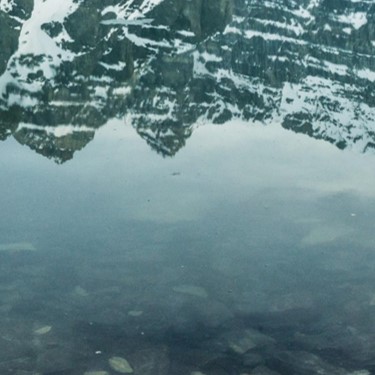} &
		\includegraphics[width = .3\linewidth, height=0.7in]{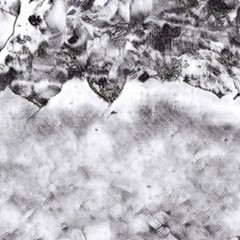} &
		\includegraphics[width = .3\linewidth, height=0.7in]{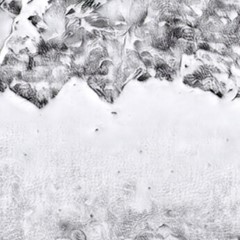} & \\
		{(a) Content }& {(b) Original}& {(c) Ours} \\
	\end{tabular}
	\vspace{-0.5em}
	\caption{Stylization detail comparison of the nuanced textures (\eg, the sky and water area) in content images, using the original model and our compressed model.}
	\label{fig:nuance}
	\vspace{-2em}
\end{figure}

Fig.~\ref{fig:wct_stylized_sample} shows the comparison of the stylized results.
Generally, our model achieves comparable or even more visually pleasing stylized images with much fewer parameters.  
Stylized images produced by original VGG-19 model and our compressed model are fairly better-looking (more colorful and sharper) than those by the FP-slimmed model.
Compared with the original VGG-19 model, our compressed model tends to produce results with \emph{fewer} messy textures, while the original model often highlights too many textures in a stylized image. 
For example, sky and water usually look smooth in an image, but actually, there are no absolutely smooth parts in natural images, so there are still nuances in the sky and water area.  
In Fig.~\ref{fig:nuance}, the original VGG-19 model tends to highlight these nuances to so obvious an extent that the whole image looks messy, while our model only emphasizes the most semantically salient parts.
This phenomenon can be explained since a model with fewer parameters has limited capacity, which is less prone to overfitting.  
It is natural that an over-parameterized model will spend its extra capacity fitting the noises in data, \eg, the nuanced textures of the sky and water in this context.
Meanwhile, even though a compressed model tends to be less overfitting, the model pruned by FP~\cite{li2017pruning} has a problem of losing description power, which is embodied in two aspects in Fig.~\ref{fig:wct_stylized_sample}.
First, as we see, FP-slimmed model tends to produce the stylized images with \emph{less} color diversity.
Second, when looked closer, the stylized images by FP-slimmed model have serious checkerboard artifacts. 
%
%
%

\vspace{0.400em}
\noindent \textbf{User study}. Style transfer has an open problem of lacking broadly-accepted comparison criteria~\cite{jing2017nstreview}, mainly because stylization is quite subjective.  
For a more objective comparison, previous works~\cite{li2017universal} conducted user studies to investigate user preference over different stylized results.  
Here we adopt this idea by investigating which is the most visually pleasing among the stylized images produced by the three models.  
We generate $20$-pair stylized images using the three models.  
Among them, $10$ pairs are randomly selected for each subject to choose which one is the most visually pleasing.  
In this part, we received $600$ valid votes.
The user study results are shown in Tab.~\ref{tab:user_study}, where the stylized images by our compressed model are top-rated on average, in line with the qualitative comparison in Fig.~\ref{fig:wct_stylized_sample}.

\vspace{0.400em}
\noindent \textbf{Style distance}. To further quantitatively evaluate the three models, we explore the style similarity between the stylized image and the style image.
Intuitively, a more successfully stylized image should be closer to the style image in the style space, so we define \emph{style distance} as
\begin{equation}
    \mathcal{D}_\text{style}^{(k)} = \|\mathcal{G}(\mathcal{F}^{(k)}(\mathcal{I}_{\text{stylized}})) - \mathcal{G}(\mathcal{F}^{(k)}(\mathcal{I}_{\text{style}}))\|_2,
\label{eqn:style_distance}
\end{equation}
where $\mathcal{G}$ is the Gram matrix.
Particularly, we feed the $20$-pair stylized images in the user study to the original VGG-19 to extract features of the $5$-stage layers (\verb+ReLU_k_1+, $k \in \{1,2,3,4,5\}$), then the style distances are calculated based on these features.
Tab.~\ref{tab:style_distance} shows that the stylized images generated by the original and our compressed models are significantly closer to the style images than those by the FP-slimmed model. 
Our model is fairly comparable with the original one, which agrees with the user preference (Tab.~\ref{tab:user_study}).

\begin{table}
    \centering
    \caption{User study of preference over different deep models within three universal artistic style transfer frameworks.}
    \vspace{-0.5em}
    \begin{tabular}{lccc}
    \toprule
       Stylization scheme                         & Original              & FP-slimmed        & Ours \\
    \midrule
        WCT~\cite{li2017universal}                & $33.0\%$              & $24.3\%$          & $\mathbf{42.7\%}$ \\
        AdaIN~\cite{Huang-2017-arbitrary}         & $\mathbf{51.1\%}$     & $6.5\%$          & $42.4\%$ \\
        Gatys~\cite{GatysTransfer-CVPR2016}       & $\mathbf{46.5\%}$     & $22.3\%$          & $31.2\%$ \\
    \bottomrule
    \end{tabular}
    \label{tab:user_study}
\end{table}

\begin{table}
    \centering
    \caption{Style distance comparison among three models. Values are the style distances defined as~Eq.~(\ref{eqn:style_distance}) based on the $5$-stage deep features (smaller is better).}
    \vspace{-0.5em}
    \begin{tabular}{l  p{0.75cm}<{\centering}  p{0.75cm}<{\centering}  p{0.75cm}<{\centering}  p{0.75cm}<{\centering}  p{0.75cm}<{\centering}}
    \toprule
       Model & Conv1 & Conv2 & Conv3 & Conv4 & Conv5 \\
    \midrule
        Original           & $\mathbf{25.5}$ & $44.6$          & $\mathbf{236.3}$        & $465.5$          & $476.5$ \\
        FP-slimmed                 & $43.8$          & $64.3$          & $416.4$                 & $597.8$          & $482.1$ \\
        Ours               & $28.1$          & $\mathbf{43.8}$ & $269.7$                 & $\mathbf{446.4}$ & $\mathbf{399.0}$ \\
    \bottomrule
    \end{tabular}
    \label{tab:style_distance}
    \vspace{-1em}
\end{table}

\begin{table*}[t]
    \centering
    \caption{Summary of the original models and our compressed models. Storage is measured in PyTorch model. GFLOPs and Inference time are measured when the content and style are both RGB images of $3000\times3000$ pixels. Max input resolution is measured on GPU using PyTorch implementation, with content and style are \emph{square} RGB images of the same size.} 
    \vspace{-0.5em}
    \begin{tabular}{lccccc}
    \toprule
       Model & \#~Params ($10^6$) & Storage (MB) & \#~GFLOPs & Inference time (GPU/CPU, s) & Max resolution \\
    \midrule
        Original       & $17.1$              &$66.6$              & $6961.7$             & $31.2$/$937.7$                    & $3000\times3000$ \\
        Ours           & $1.1(15.5\times)$  &$5.2(12.8\times)$  & $451.6(15.4\times)$ & $8.6(3.6\times)$/$366.0(2.6\times)$  & $6000\times6000$ \\
    \bottomrule
    \end{tabular}
    \label{tab:speedup_etc}
    \vspace{-1em}
\end{table*}

Aside from the high-resolution stylized images shown above, we show an ultra-resolution stylized image in Fig.~\ref{fig:UHD_image1}.
The shapes and textures are still very clear even zoomed in on a large scale.  
To our best knowledge, this is the \emph{first} time that we can obtain ultra-resolution universal style transfer results using a single 12GB GPU.
%
%
We also report the statistics about the model size and speedup in Tab.~\ref{tab:speedup_etc}.
Our compressed model is $15.5\times$ smaller and $3.6\times$ faster than the original VGG-19 on GPU.
Note that the total size for all the 5-stage models is only $10$MB, easy to fit in mobile devices.

\begin{figure}[t]
   	\centering
	\begin{tabular}{c@{\hspace{0.008\linewidth}}c@{\hspace{0.008\linewidth}}c@{\hspace{0.008\linewidth}}c@{\hspace{0.008\linewidth}}c}
		\includegraphics[width = 0.18\linewidth, height=0.17\linewidth]{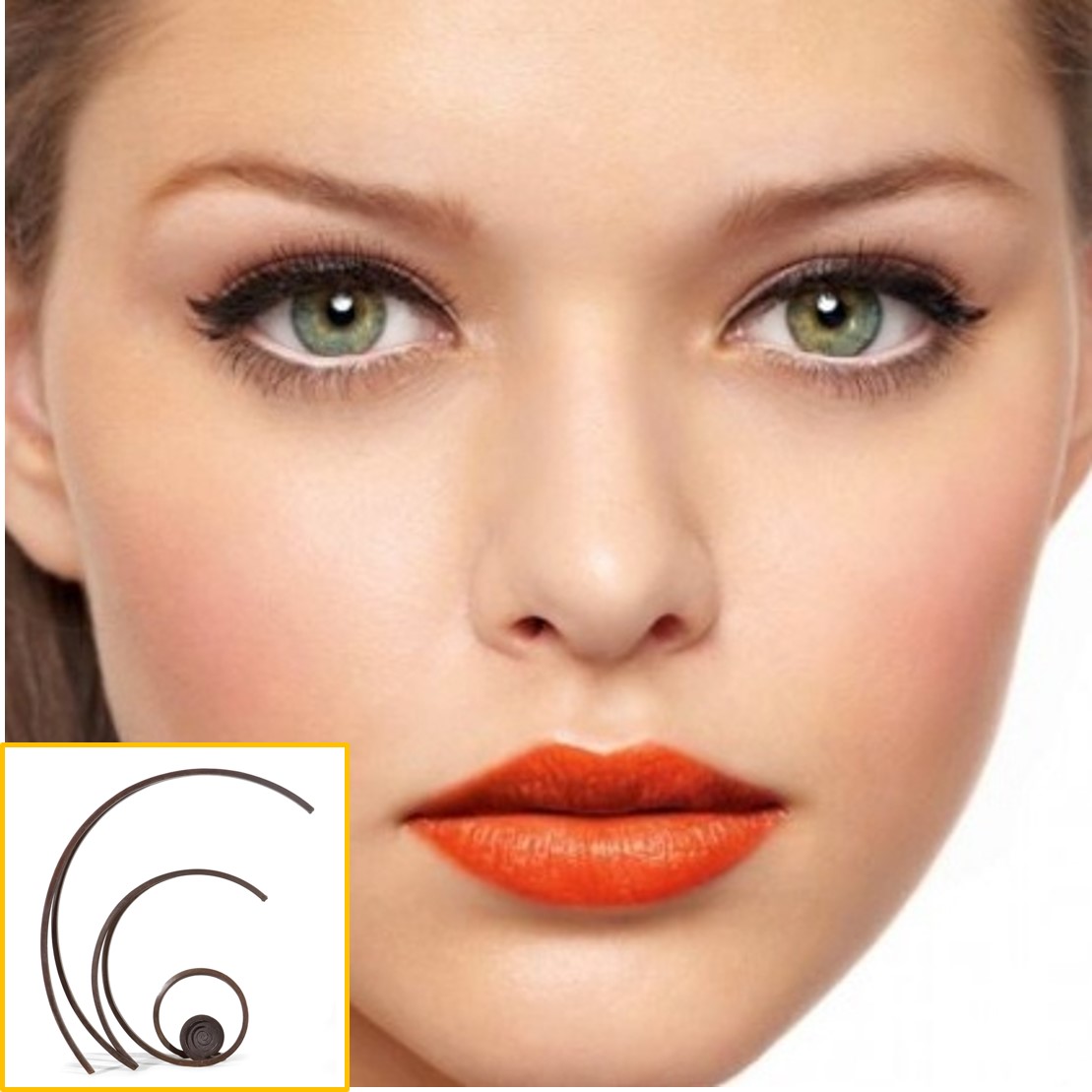} &
		\includegraphics[width = 0.18\linewidth, height=0.17\linewidth]{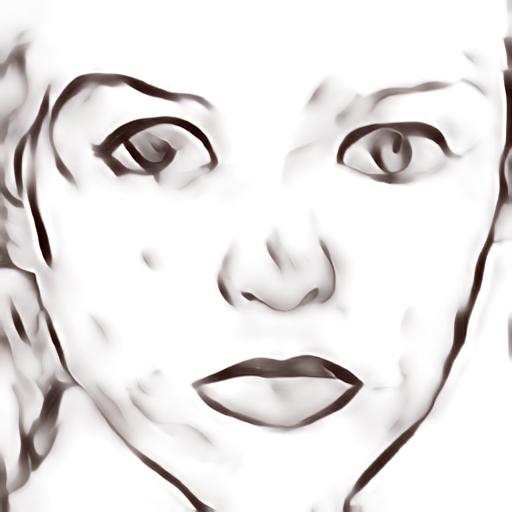} &
		\includegraphics[width = 0.18\linewidth, height=0.17\linewidth]{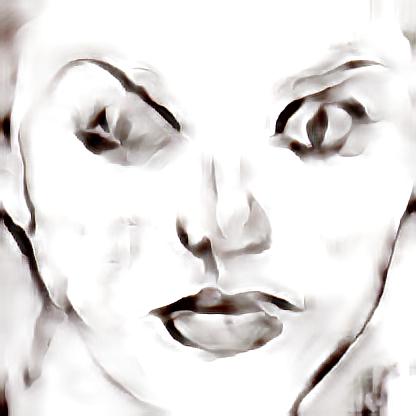} &
		\includegraphics[width = 0.18\linewidth, height=0.17\linewidth]{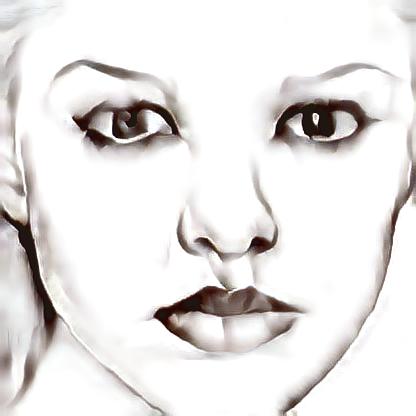} &
		\includegraphics[width = 0.18\linewidth, height=0.17\linewidth]{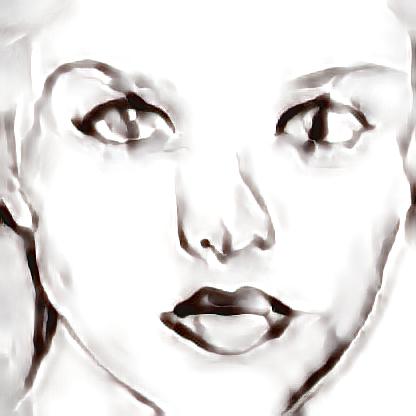} \\
		\includegraphics[width = 0.18\linewidth, height=0.17\linewidth]{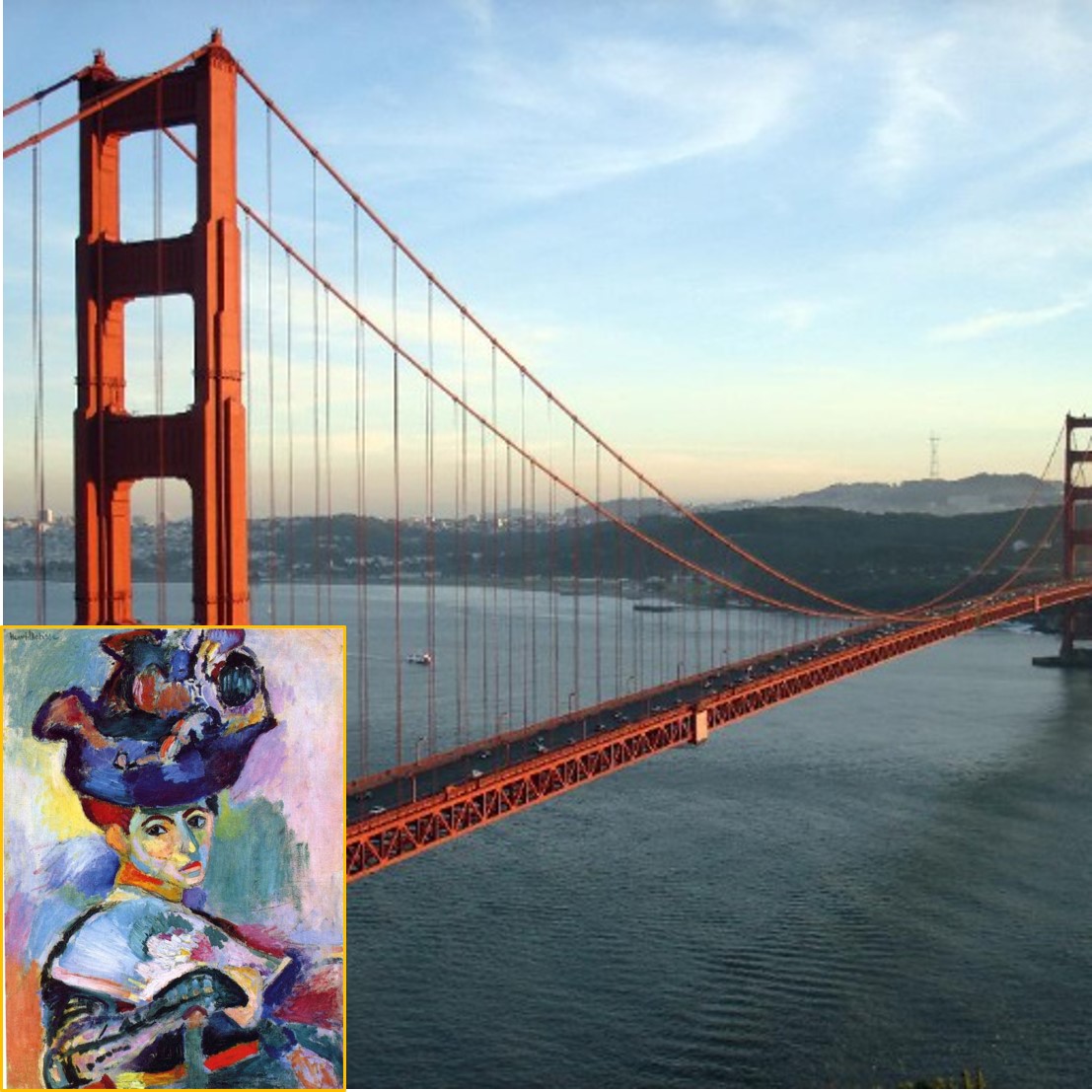} &
		\includegraphics[width = 0.18\linewidth, height=0.17\linewidth]{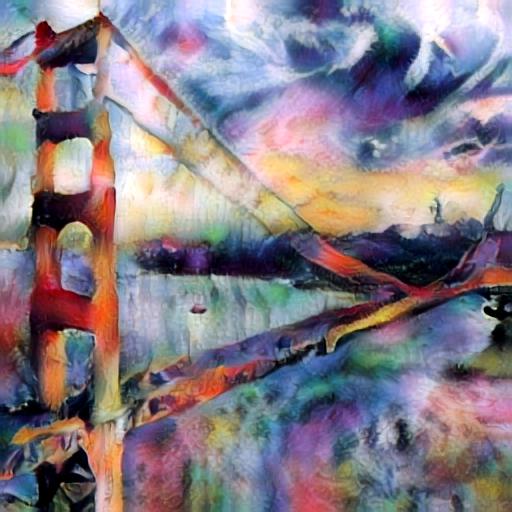} &
		\includegraphics[width = 0.18\linewidth, height=0.17\linewidth]{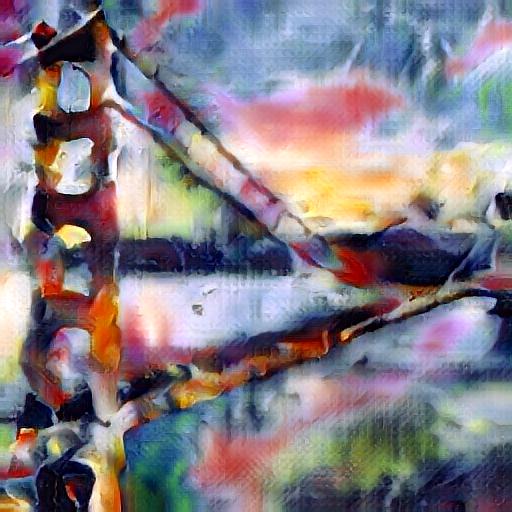} &
		\includegraphics[width = 0.18\linewidth, height=0.17\linewidth]{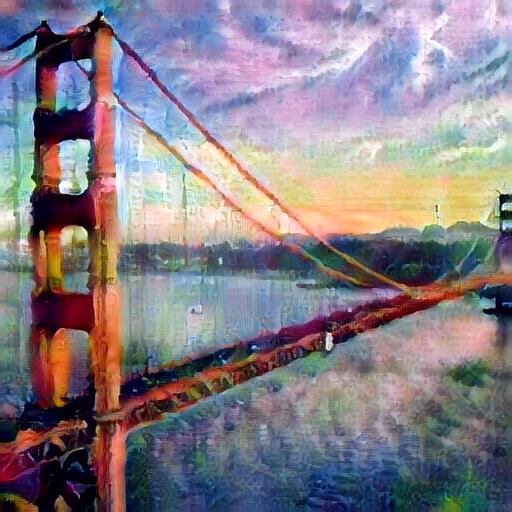} &
		\includegraphics[width = 0.18\linewidth, height=0.17\linewidth]{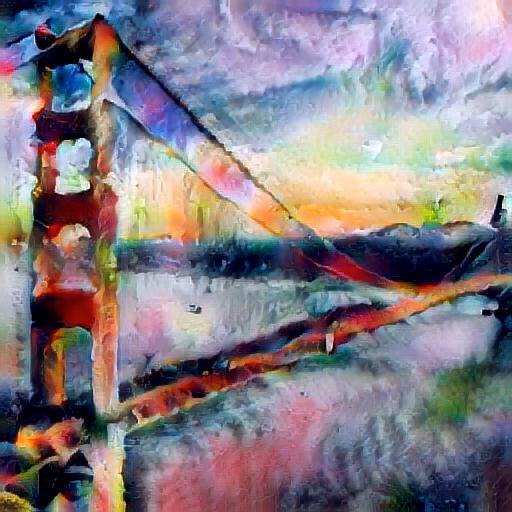} \\
		{(a) C/S} & {(b) Original} & {(c) $\mathcal{L}_{\text{embed}}$} & {(d) $\mathcal{L}_{\text{collab}}$} & {(e) Both}
    \end{tabular}
    \vspace{-0.8em}
	\caption{Ablation study of the two proposed losses on WCT. (a) Content and style. (b) Use the original models. (c) Use $\mathcal{L}_{\text{embed}}$. (d) Use $\mathcal{L}_{\text{collab}}$. (e) Both losses are used.}
	\label{fig:ablation}
	\vspace{-0.7em}
\end{figure}

\vspace{0.400em}
\noindent \textbf{Ablation study}. Here we explore the effect of the two proposed sub-schemes, collaborative distillation ($\mathcal{L}_{\text{collab}}$) and linear embedding ($\mathcal{L}_{\text{embed}}$). The ablation results in Fig.~\ref{fig:ablation} show that, both the linear embedding and collaboration losses can transfer considerable knowledge from the teacher to the student. Generally, both schemes can independently produce fairly comparable results as the original models even if $15.5\times$ smaller. Between the two losses, $\mathcal{L}_{\text{embed}}$ only transfers the information of feature domain rather than image domain, so its results have more distortion and also some checkerboard effects (please zoom in and see the details of Fig.~\ref{fig:ablation}(c)). Meanwhile, $\mathcal{L}_{\text{collab}}$ mainly focuses on the image reconstruction, so it produces sharper results with less distortion, which confirms our intuition that the decoder has considerable knowledge which we can leverage to train a small encoder. When both losses employed, we obtain the sweet spot of the two: The results do not have checkerboards, and also properly maintain the artistic distortion.


\subsection{Style transfer on AdaIN}
\begin{figure}[t]
	\centering
	\renewcommand{\arraystretch}{0.25} 
	\begin{tabular}{c@{\hspace{0.008\linewidth}}c@{\hspace{0.008\linewidth}}c@{\hspace{0.008\linewidth}}c@{\hspace{0.008\linewidth}}c}
		\includegraphics[width = 0.18\linewidth, height=0.17\linewidth]{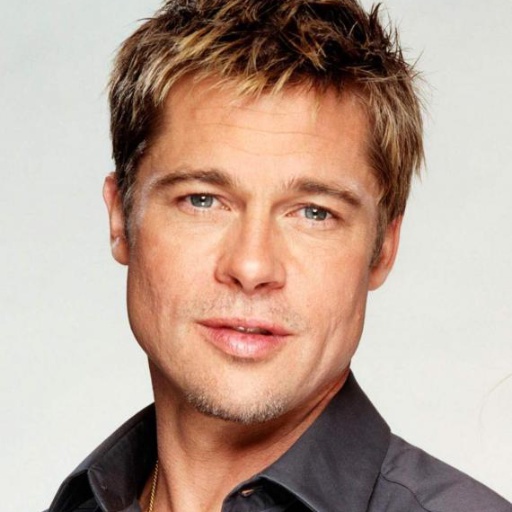} &
		\includegraphics[width = 0.18\linewidth, height=0.17\linewidth]{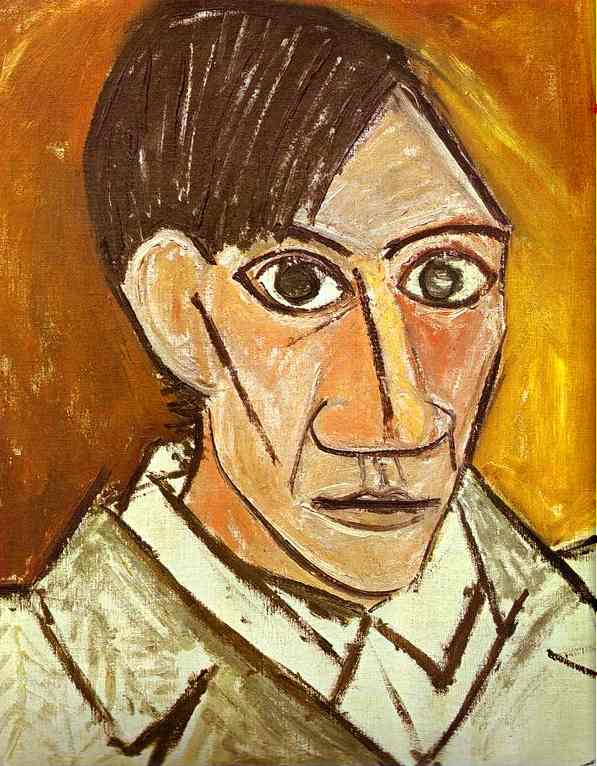} &
		\includegraphics[width = 0.18\linewidth, height=0.17\linewidth]{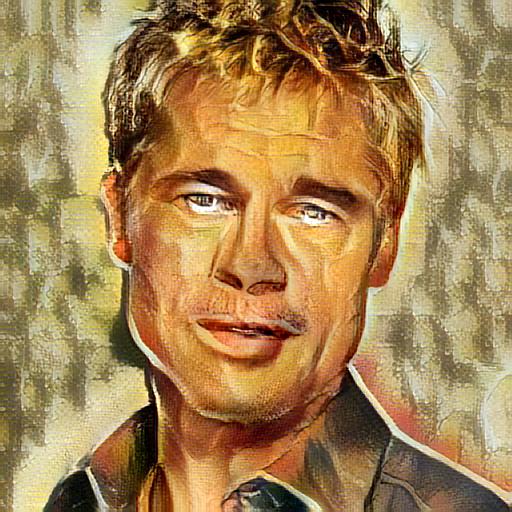} &
		\includegraphics[width = 0.18\linewidth, height=0.17\linewidth]{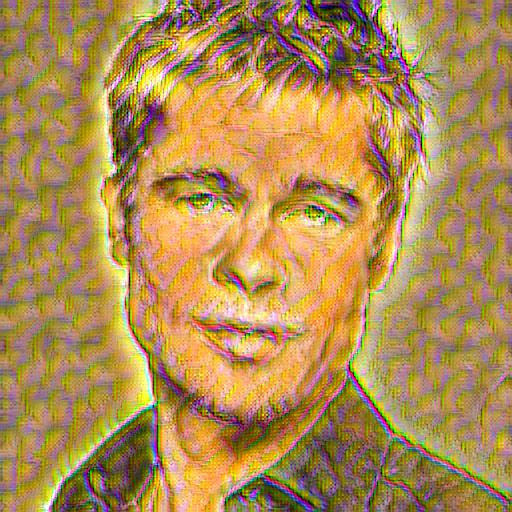} &
		\includegraphics[width = 0.18\linewidth, height=0.17\linewidth]{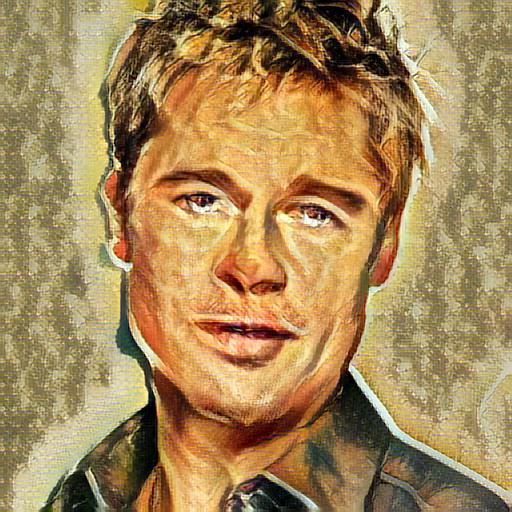} \\
		\includegraphics[width = 0.18\linewidth, height=0.17\linewidth]{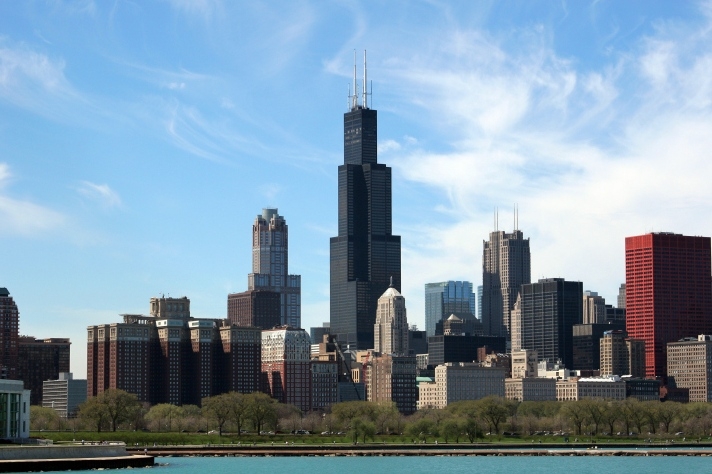} &
		\includegraphics[width = 0.18\linewidth, height=0.17\linewidth]{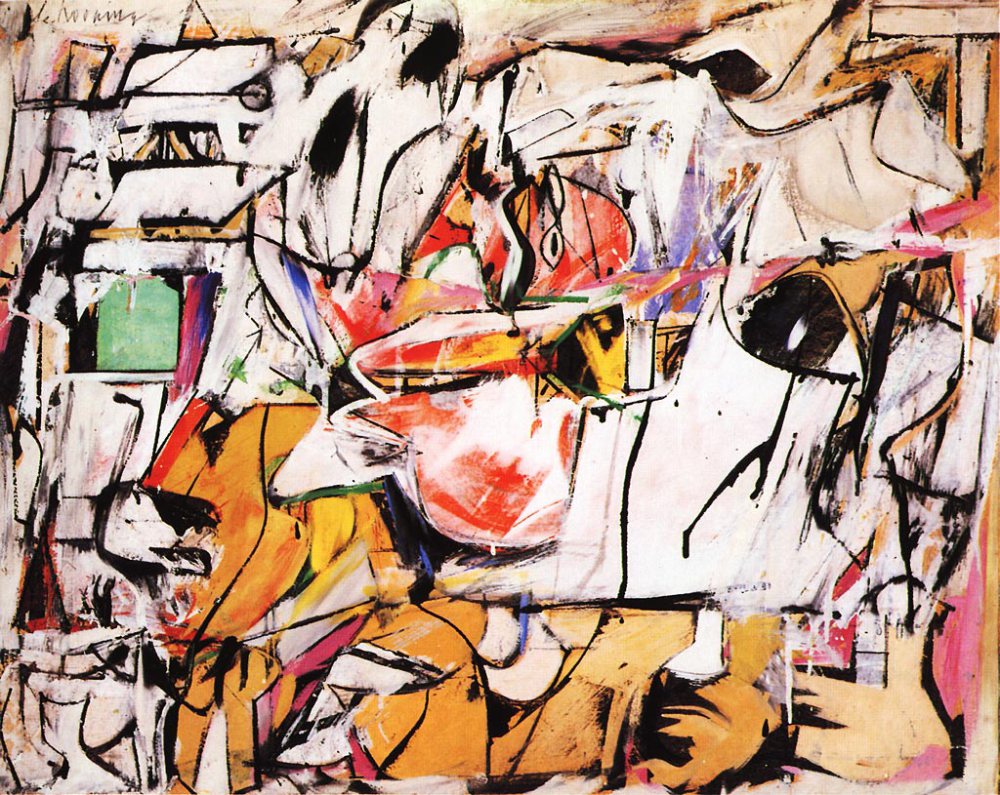} &
		\includegraphics[width = 0.18\linewidth, height=0.17\linewidth]{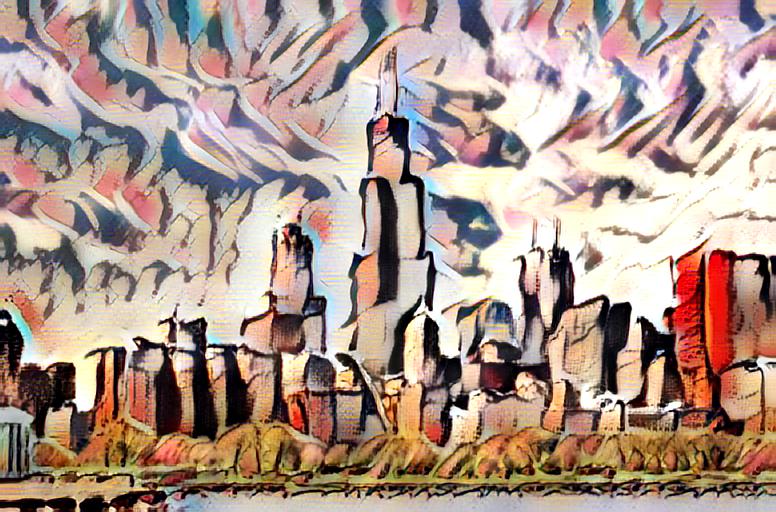} &
		\includegraphics[width = 0.18\linewidth, height=0.17\linewidth]{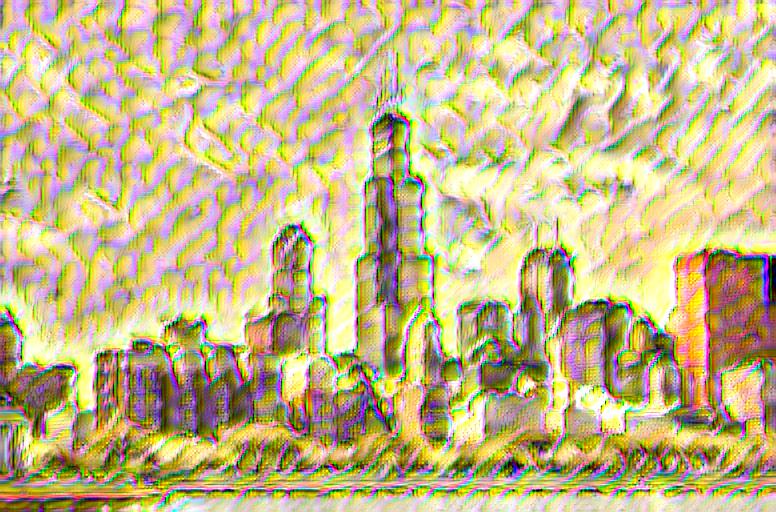} &
		\includegraphics[width = 0.18\linewidth, height=0.17\linewidth]{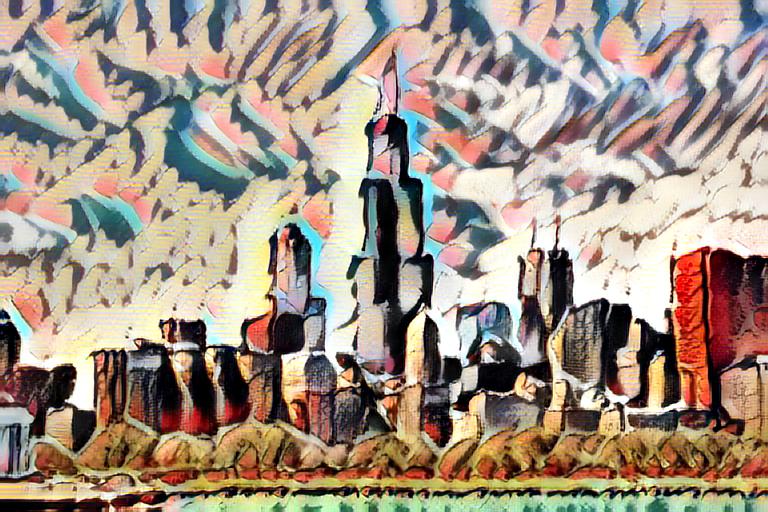} \\
		\includegraphics[width = 0.18\linewidth, height=0.17\linewidth]{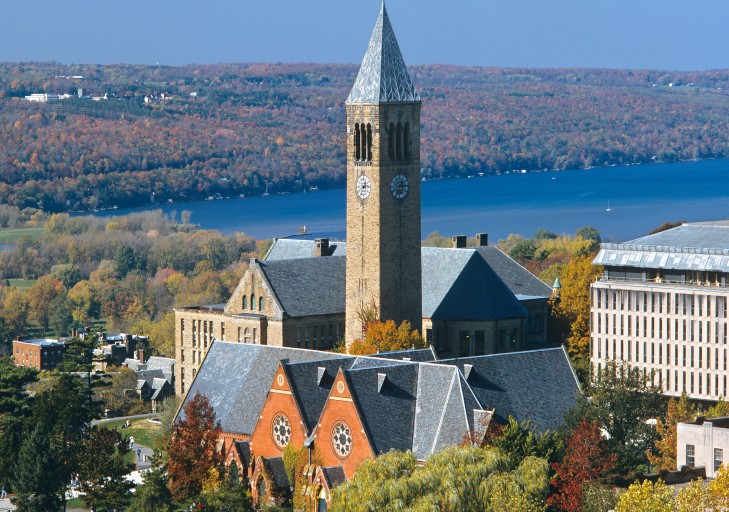} &
		\includegraphics[width = 0.18\linewidth, height=0.17\linewidth]{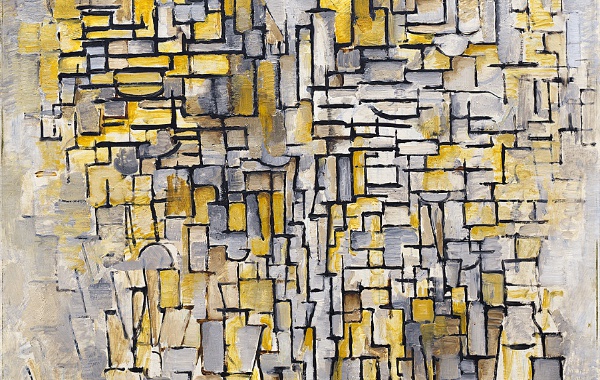} &
		\includegraphics[width = 0.18\linewidth, height=0.17\linewidth]{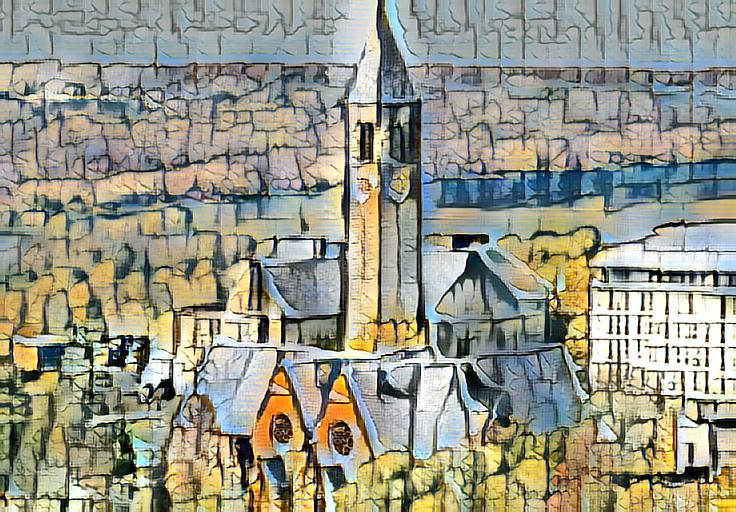} &
		\includegraphics[width = 0.18\linewidth, height=0.17\linewidth]{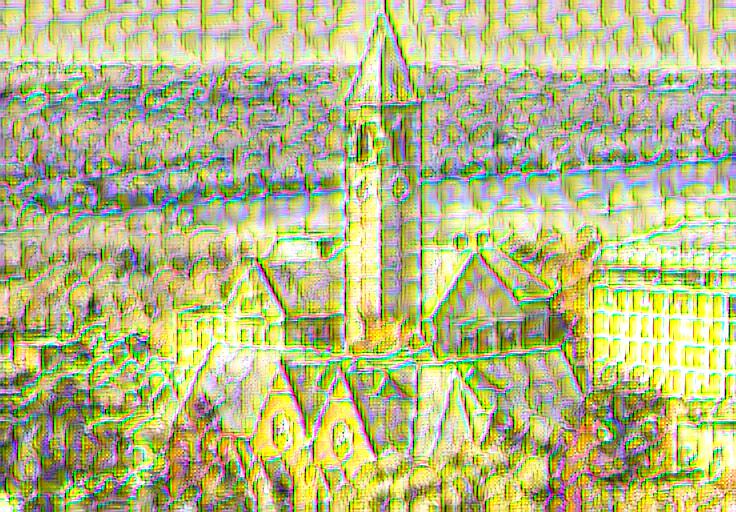} &
		\includegraphics[width = 0.18\linewidth, height=0.17\linewidth]{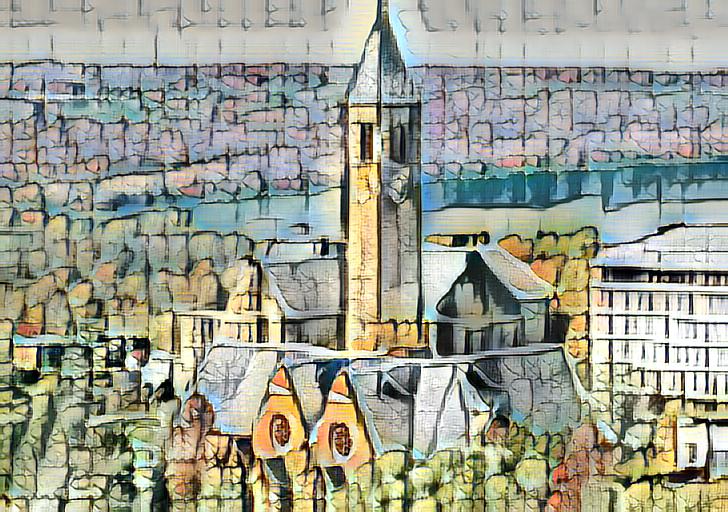} \\
		{Content} & {Style} & {Original} & {FP} & {Ours} \\
	\end{tabular}
	\vspace{-0.5em}
	\caption{Comparison of the three models with AdaIN.}
	\label{fig:adain}
	\vspace{-1em}
\end{figure}

We further evaluate the proposed method on AdaIN, where the encoder-decoder collaboration task is style transfer. The training process and network architectures are set the same way as Sec.~\ref{sec:exp_wct}, except that the $\mathcal{L}_{\text{st}}$ (Eq.~\ref{eqn:style_transfer_loss}) is now utilized as the collaborator loss. $\lambda_s$ is set to $10$ and $\beta$ is set to $10$. The results are shown in Fig.~\ref{fig:adain}, where our compressed encoder produces visually comparable results with the original one, while FP-slimmed model degrades the visual effect significantly. This visual evaluation is further justified by the user study in Tab.~\ref{tab:user_study}, where ours and the original model receive similar votes, significantly more than those of the model compressed by the FP algorithm~\cite{li2017pruning}.

\subsection{Optimization-based style transfer}
\begin{figure}[t]
	\centering
	\renewcommand{\arraystretch}{0.25} 
	\begin{tabular}{@{\hspace{0\linewidth}}c@{\hspace{0.008\linewidth}}c@{\hspace{0.008\linewidth}}c@{\hspace{0.008\linewidth}}c@{\hspace{0.008\linewidth}}c}
		\includegraphics[width = 0.18\linewidth, height=0.17\linewidth]{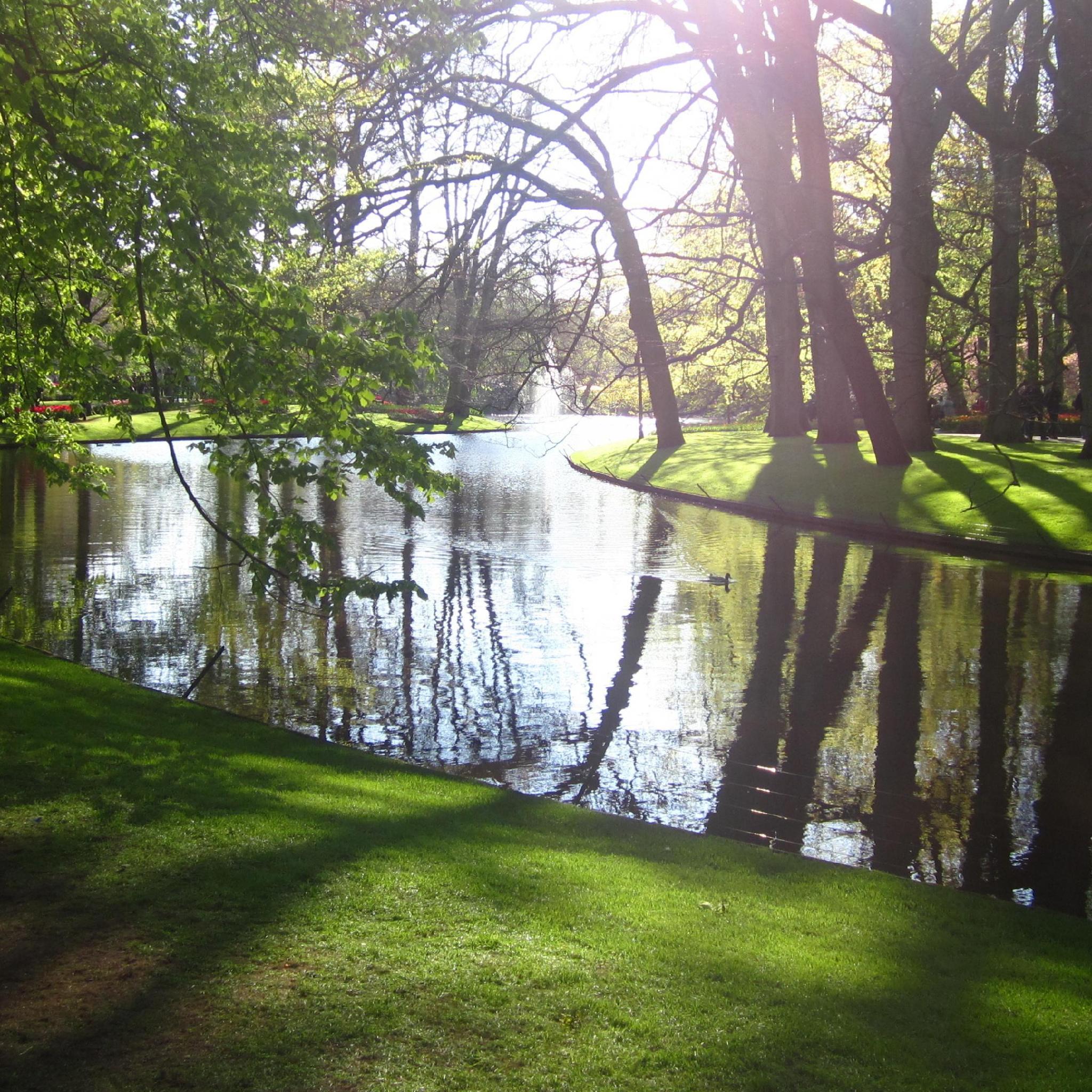} &
		\includegraphics[width = 0.18\linewidth, height=0.17\linewidth]{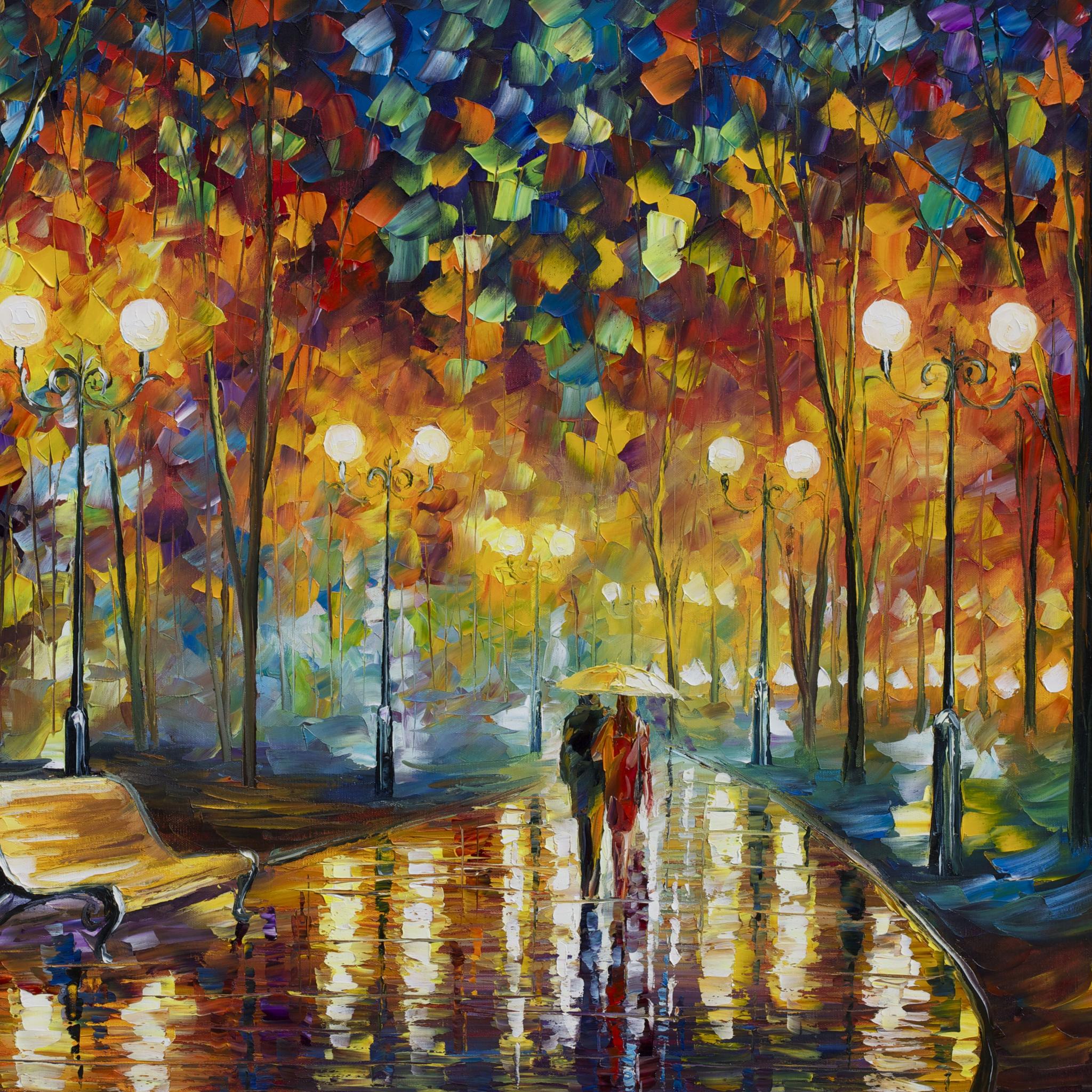} &
		\includegraphics[width = 0.18\linewidth, height=0.17\linewidth]{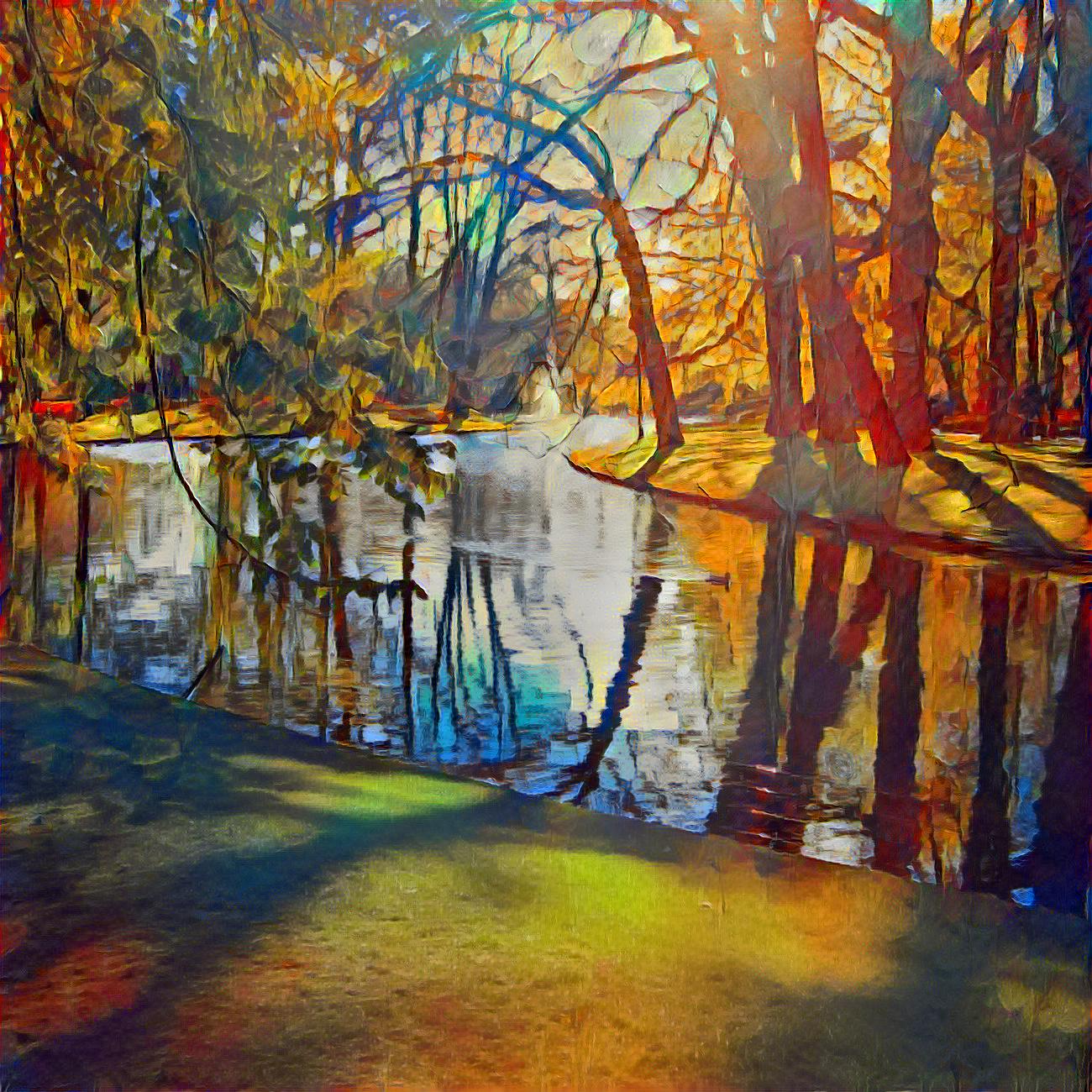} &
		\includegraphics[width = 0.18\linewidth, height=0.17\linewidth]{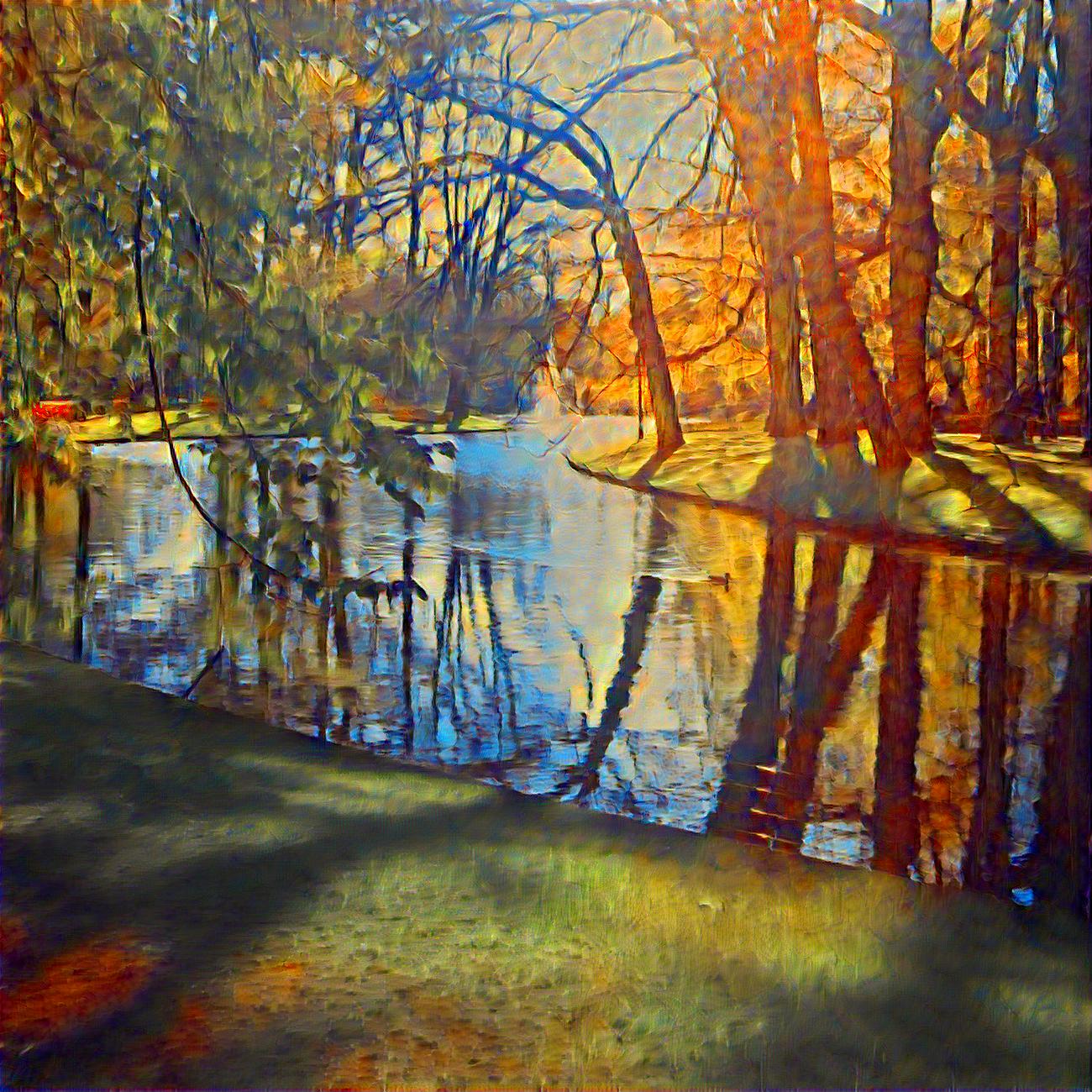} &
		\includegraphics[width = 0.18\linewidth, height=0.17\linewidth]{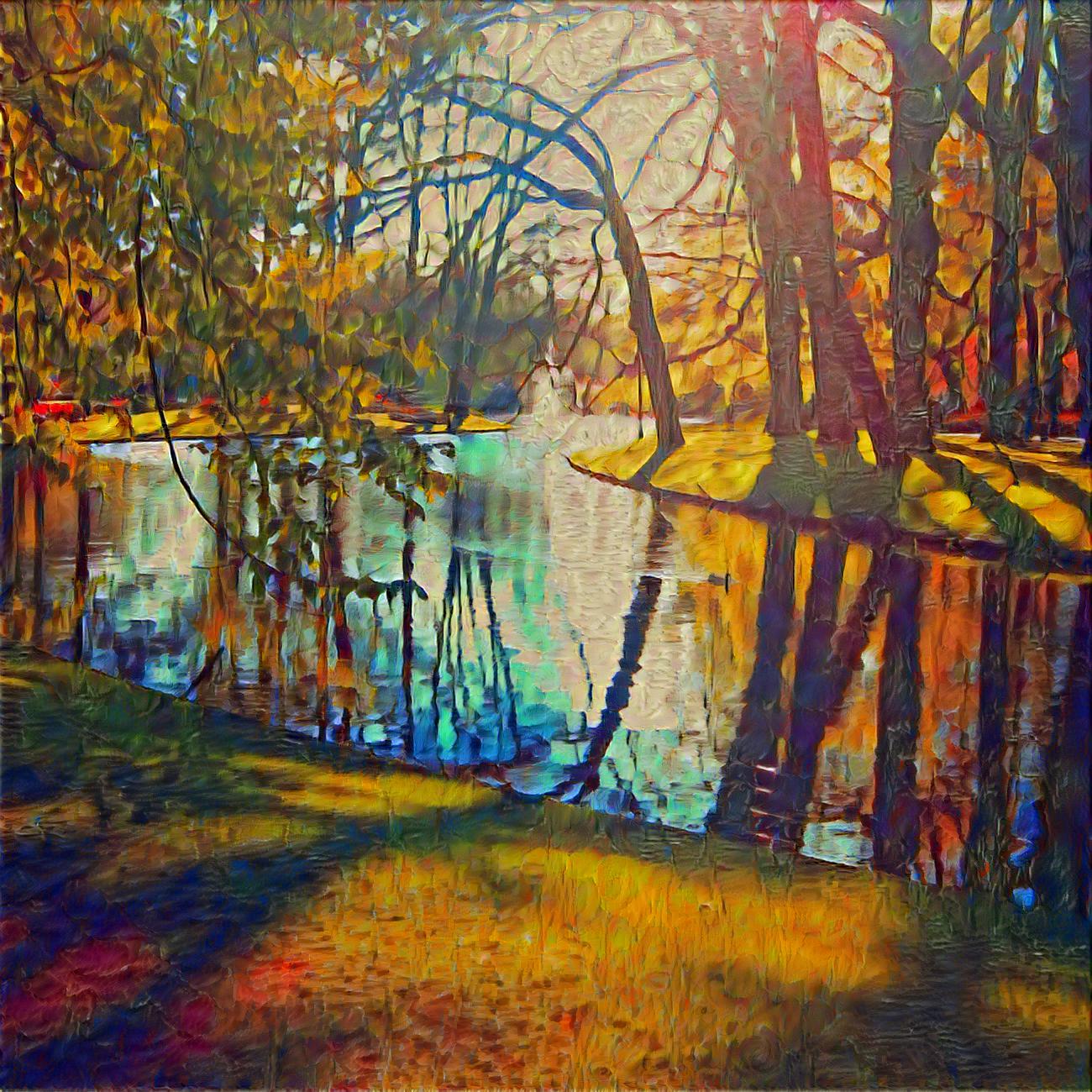} \\
		\includegraphics[width = 0.18\linewidth, height=0.17\linewidth]{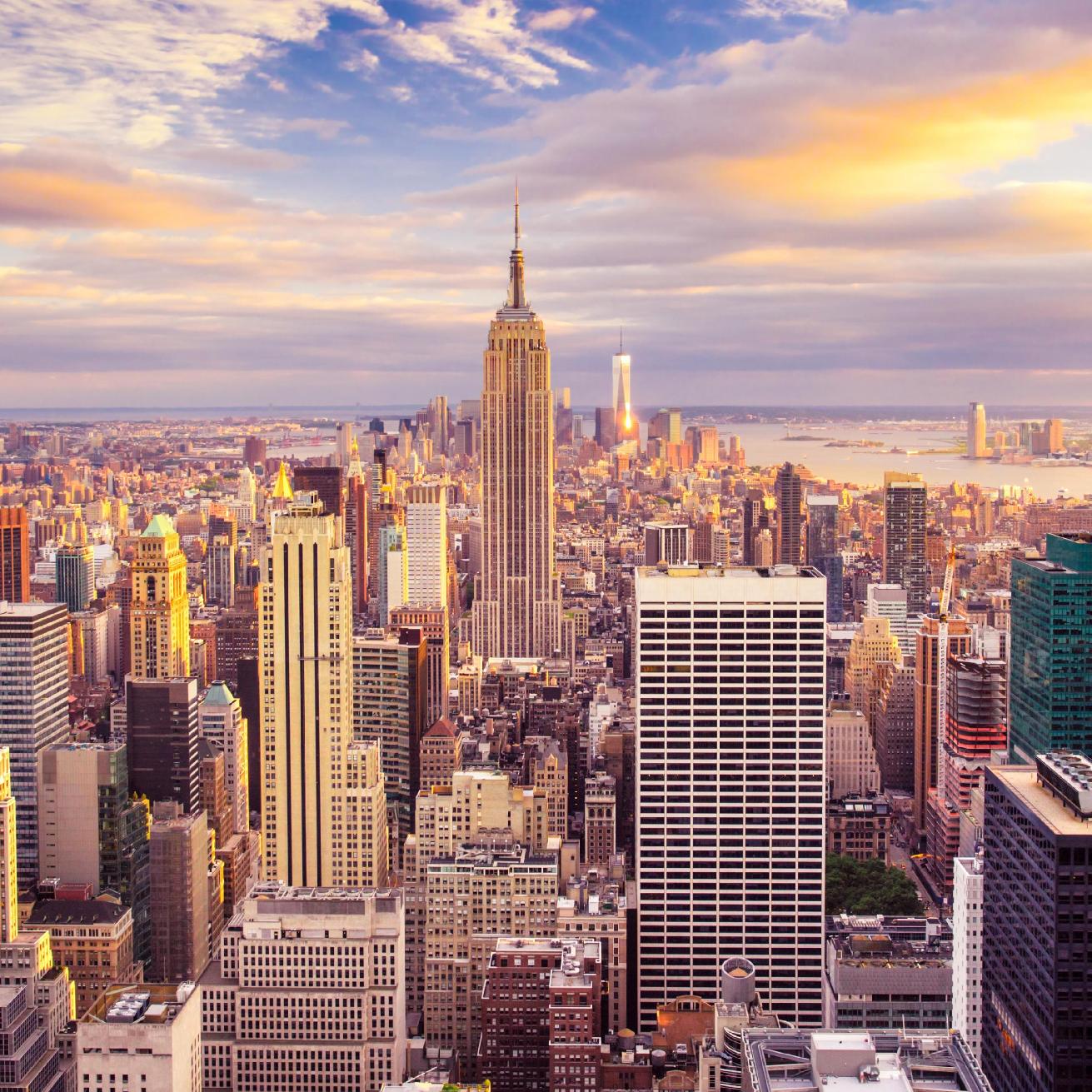} &
		\includegraphics[width = 0.18\linewidth, height=0.17\linewidth]{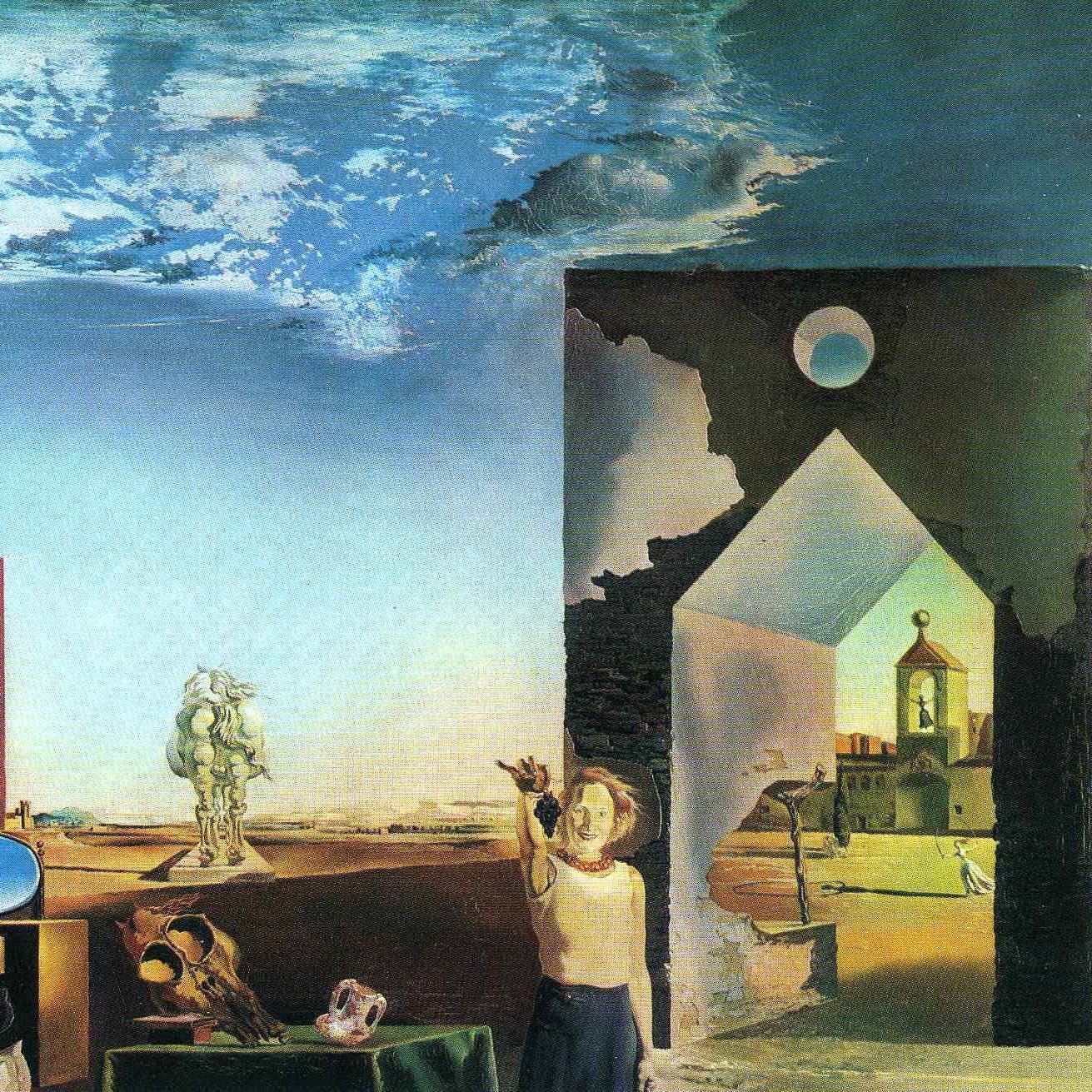} &
		\includegraphics[width = 0.18\linewidth, height=0.17\linewidth]{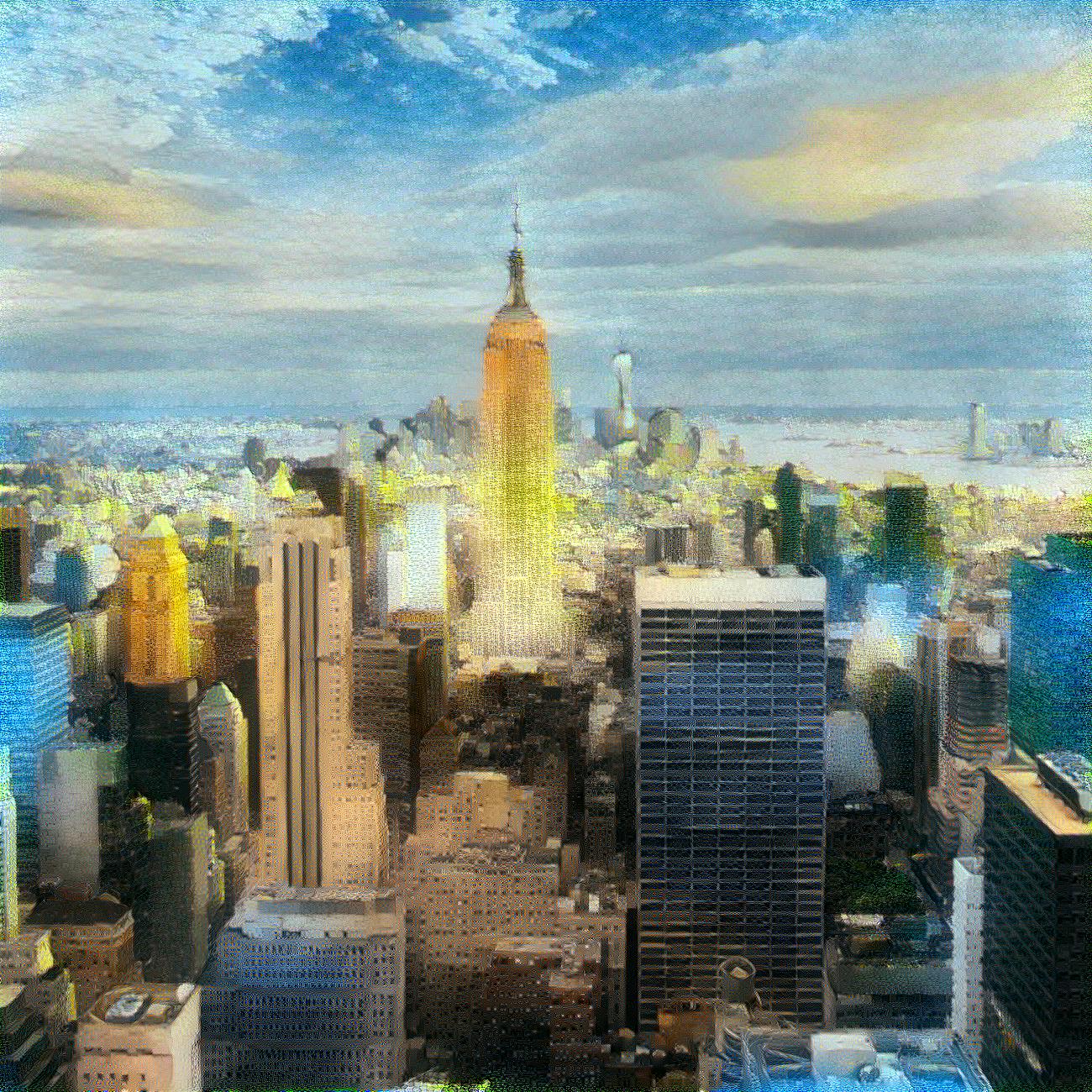} &
		\includegraphics[width = 0.18\linewidth, height=0.17\linewidth]{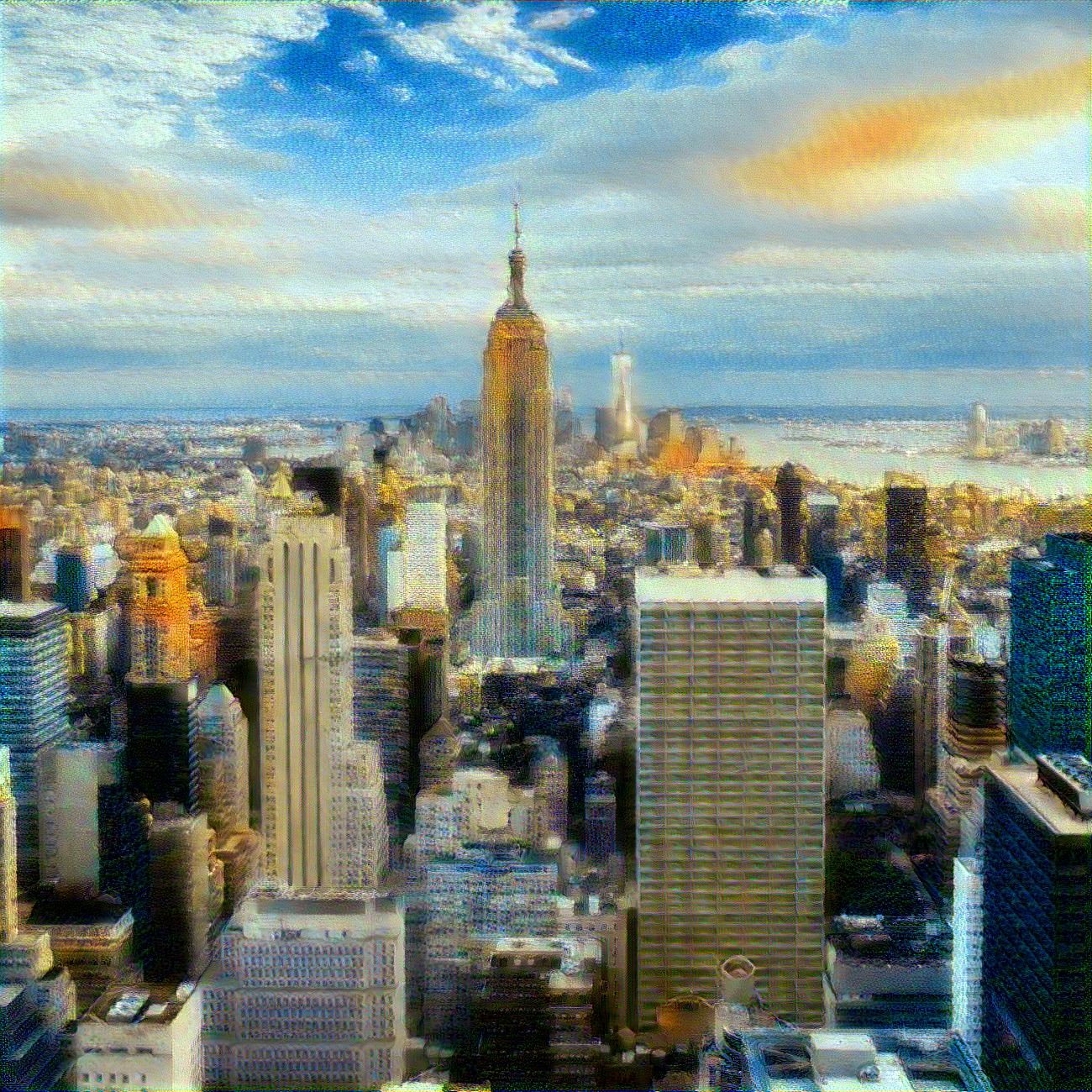} &
		\includegraphics[width = 0.18\linewidth, height=0.17\linewidth]{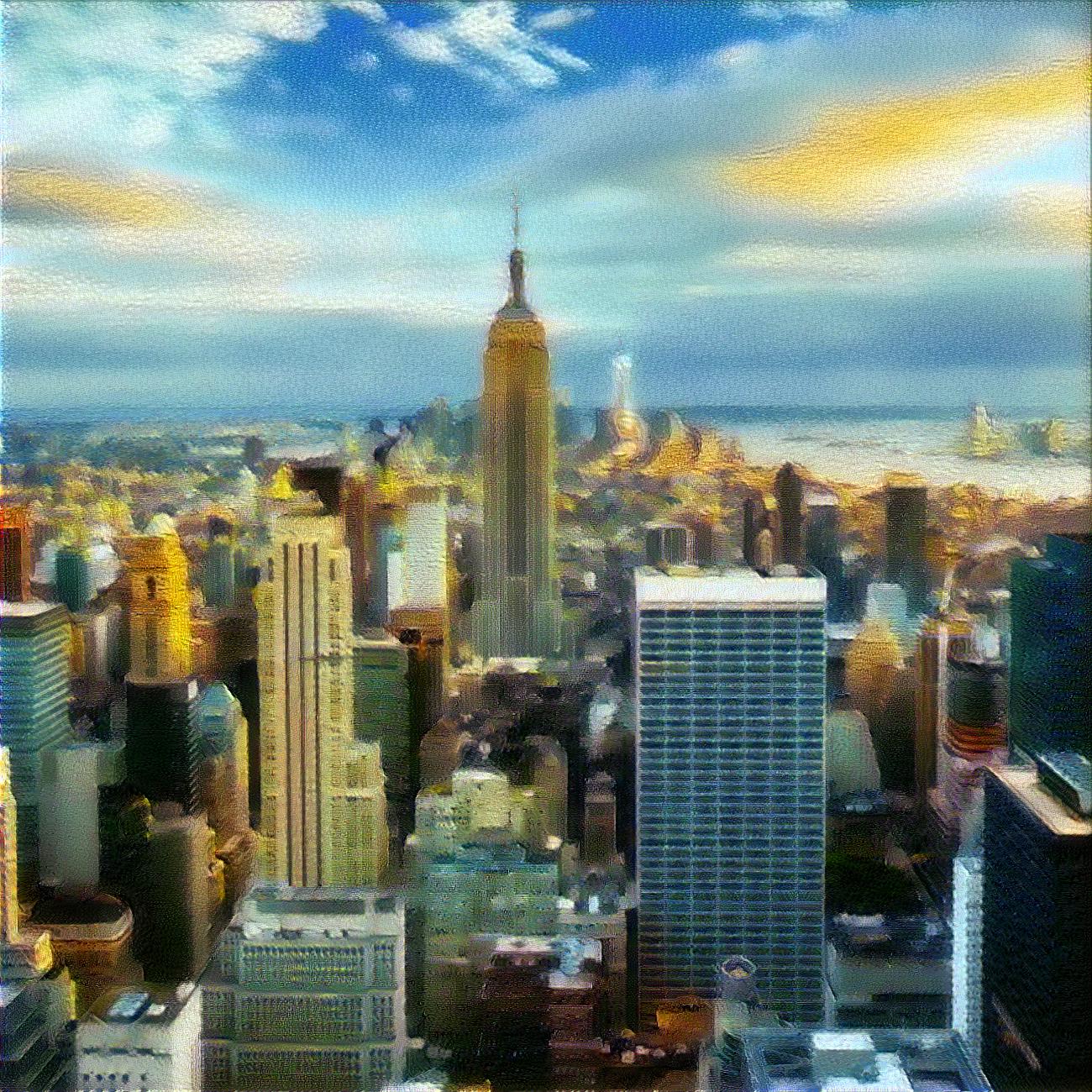} \\
		\includegraphics[width = 0.18\linewidth, height=0.17\linewidth]{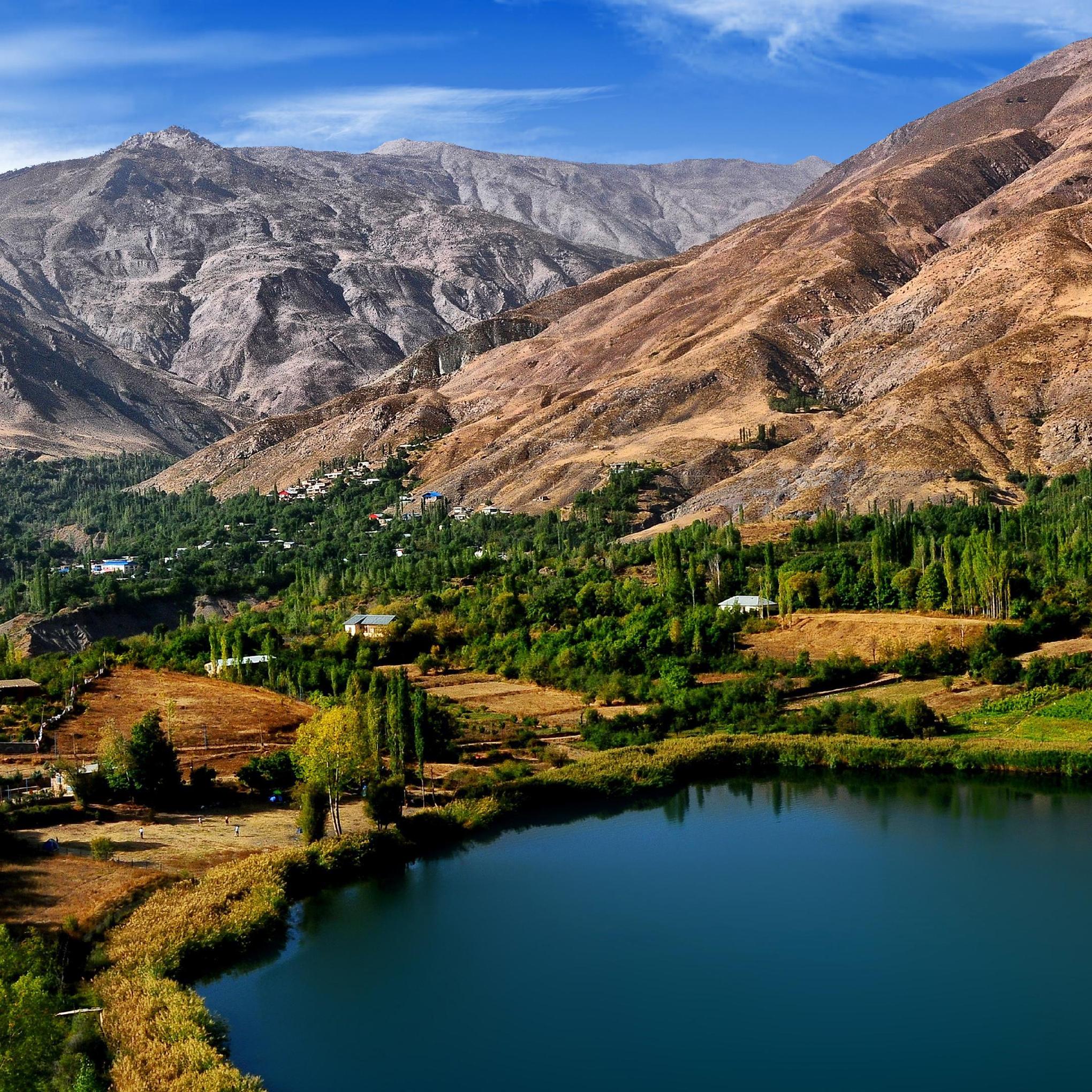} &
		\includegraphics[width = 0.18\linewidth, height=0.17\linewidth]{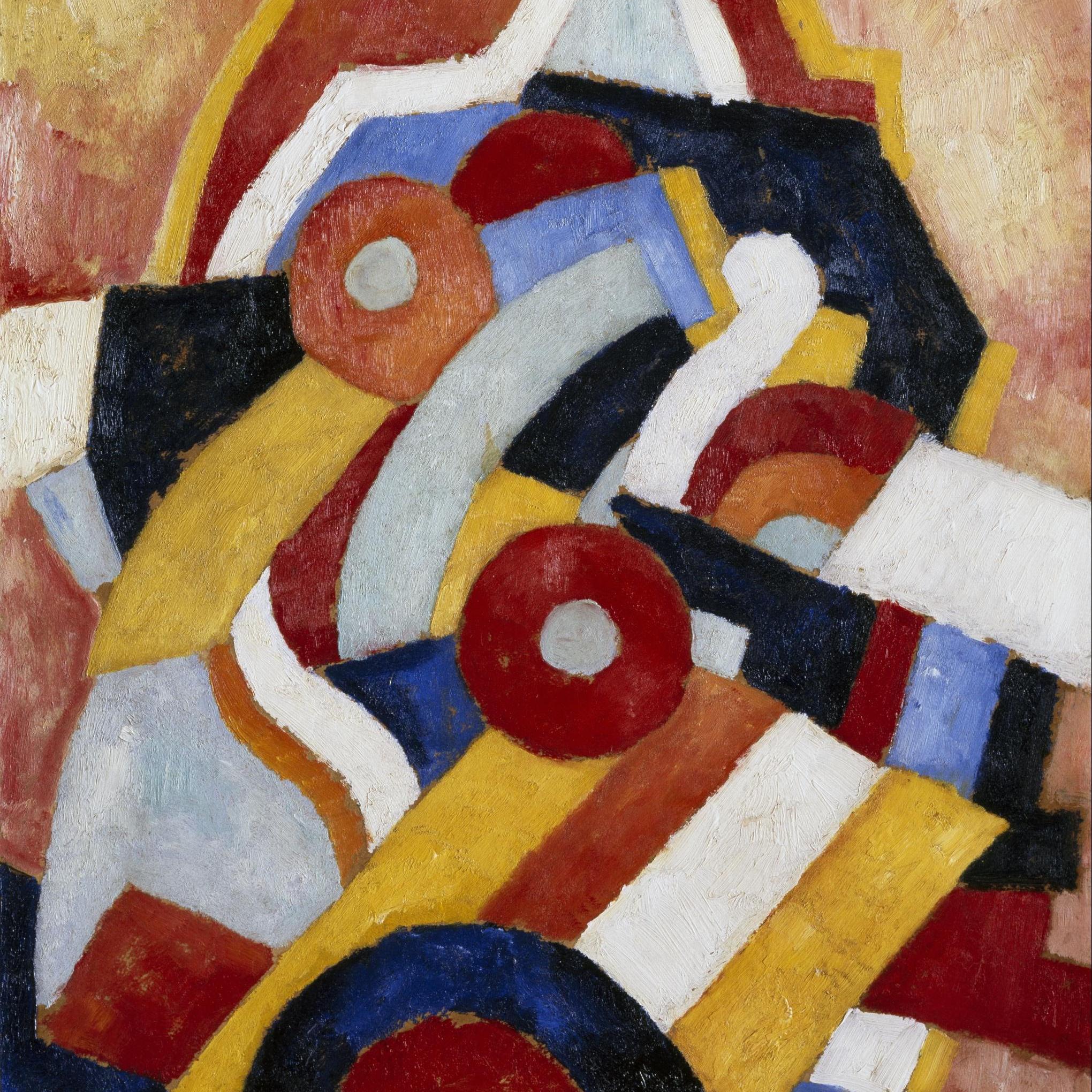} &
		\includegraphics[width = 0.18\linewidth, height=0.17\linewidth]{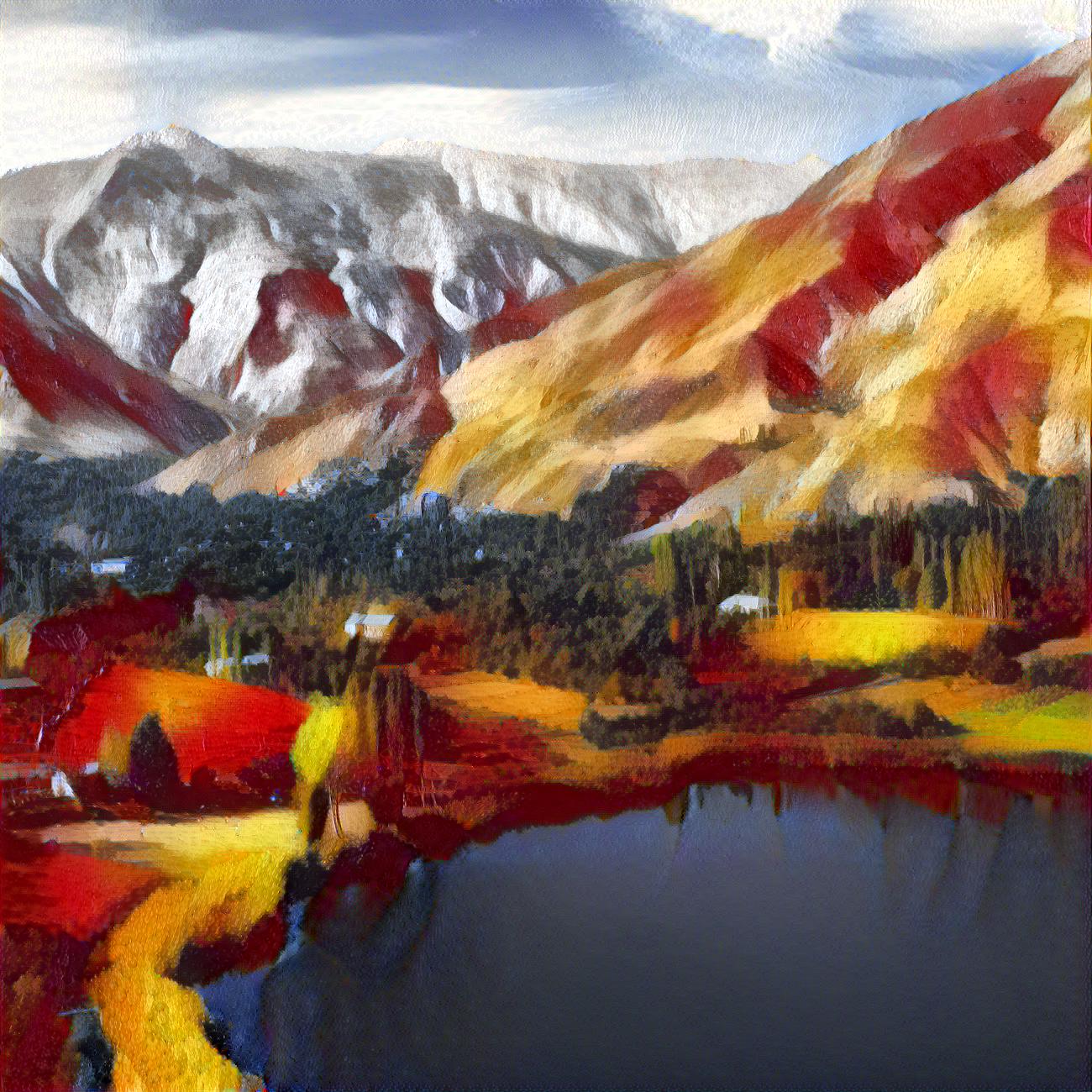} &
		\includegraphics[width = 0.18\linewidth, height=0.17\linewidth]{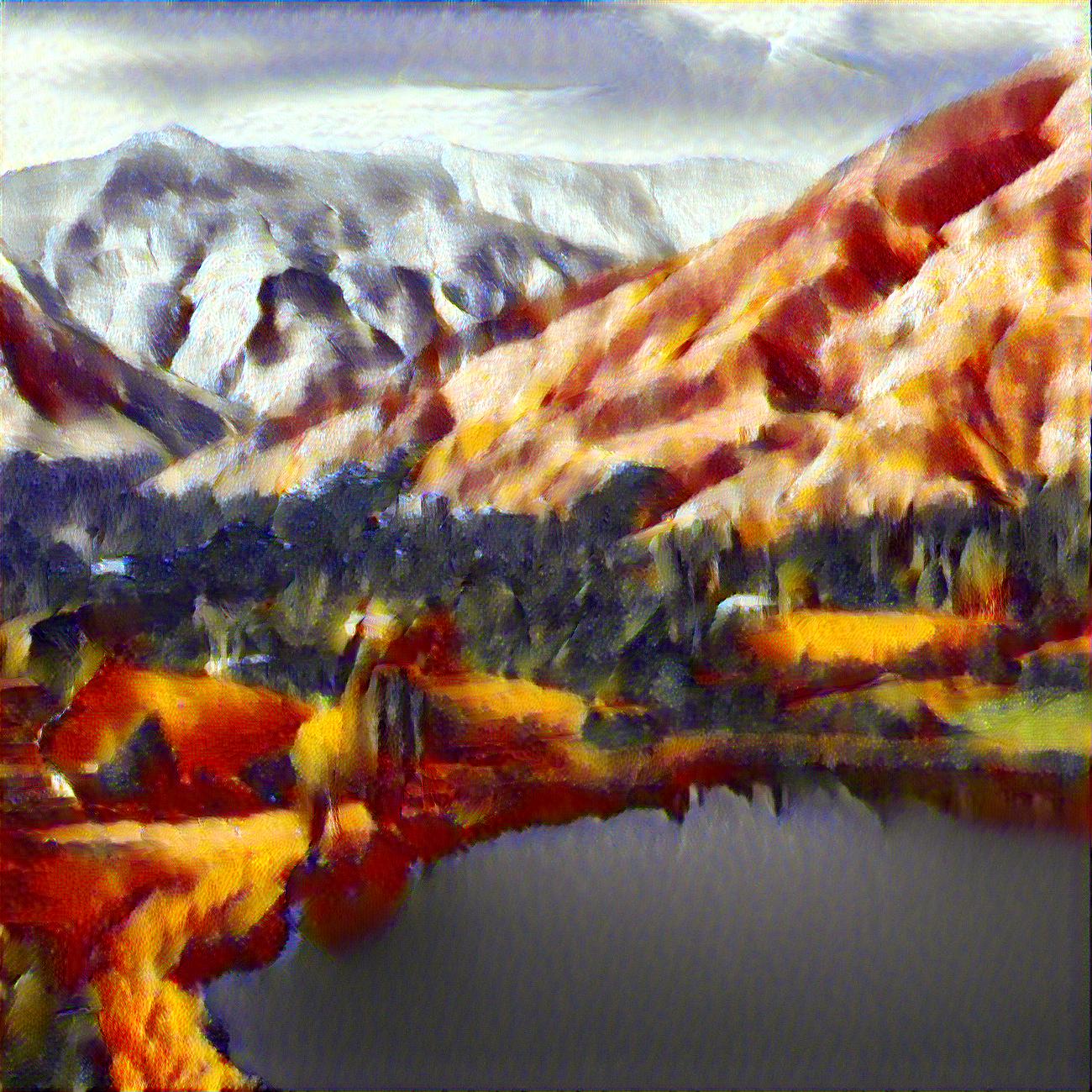} &
		\includegraphics[width = 0.18\linewidth, height=0.17\linewidth]{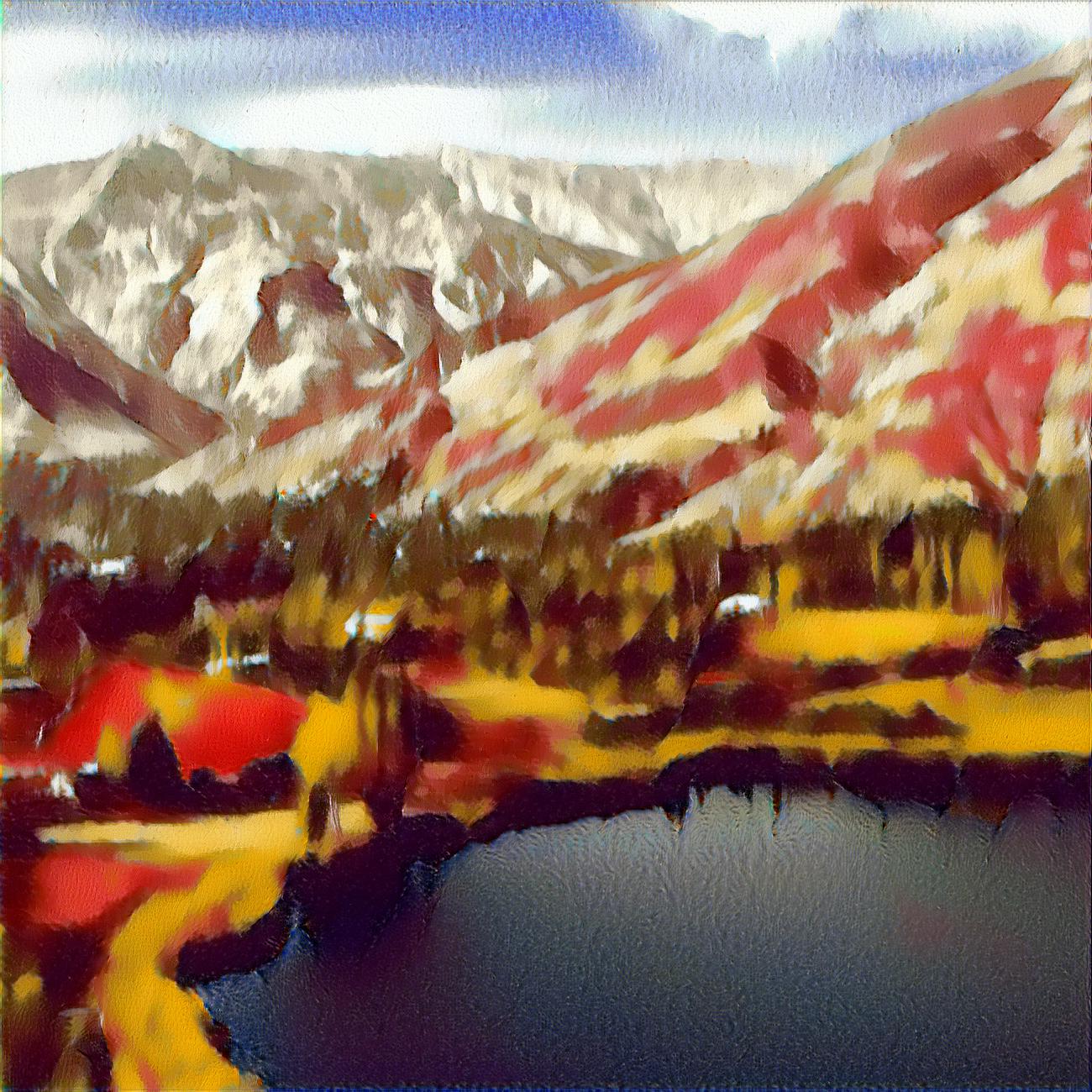} \\
		{Content} & {Style} & {Original} & {FP} & {Ours}
	\end{tabular}
	\vspace{-0.5em}
	\caption{Comparison of stylized images using the method by Gatys~\etal~\cite{GatysTransfer-CVPR2016}, with the original model and our compressed model (best viewed in color and zoomed in).}
	\label{fig:nst_gatys}
	\vspace{-1.5em}
\end{figure}

Although the work of~\cite{GatysTransfer-CVPR2016} is not encoder-decoder based, it is worthwhile to check whether the small encoder compressed by our method can still perform well in this case.
%
%
Layers of \verb+Conv_k_1+ ($k \in \{1,2,3,4,5\}$) are selected for style matching and layer \verb+Conv4_2+ selected for content matching. 
The L-BFGS~\cite{liu1989limited} is used as the optimization solver.
Fig.~\ref{fig:nst_gatys} shows that the compressed model can achieve fairly comparable results with those generated by the original VGG-19.  
Similar to the experiments of artistic style transfer in Sec.~\ref{sec:exp_wct}, we also conduct a user study for a more objective comparison. 
Results in Tab.~\ref{tab:user_study} show that our compressed model is slightly less popular than the original model (which is reasonable considering that the compressed model has $15.5\times$ fewer parameters), while still more preferred than the model compressed by the FP method.
%

\section{Discussions}
\label{sec:discussion}
\begin{figure}[t]
   \renewcommand{\arraystretch}{0.1} 
   \centering
   \begin{tabular}{c@{\hspace{0.008\linewidth}}c@{\hspace{0.008\linewidth}}c@{\hspace{0.008\linewidth}}c @{\hspace{0.008\linewidth}}c @{\hspace{0.008\linewidth}} c}
     \includegraphics[width=0.24\linewidth]{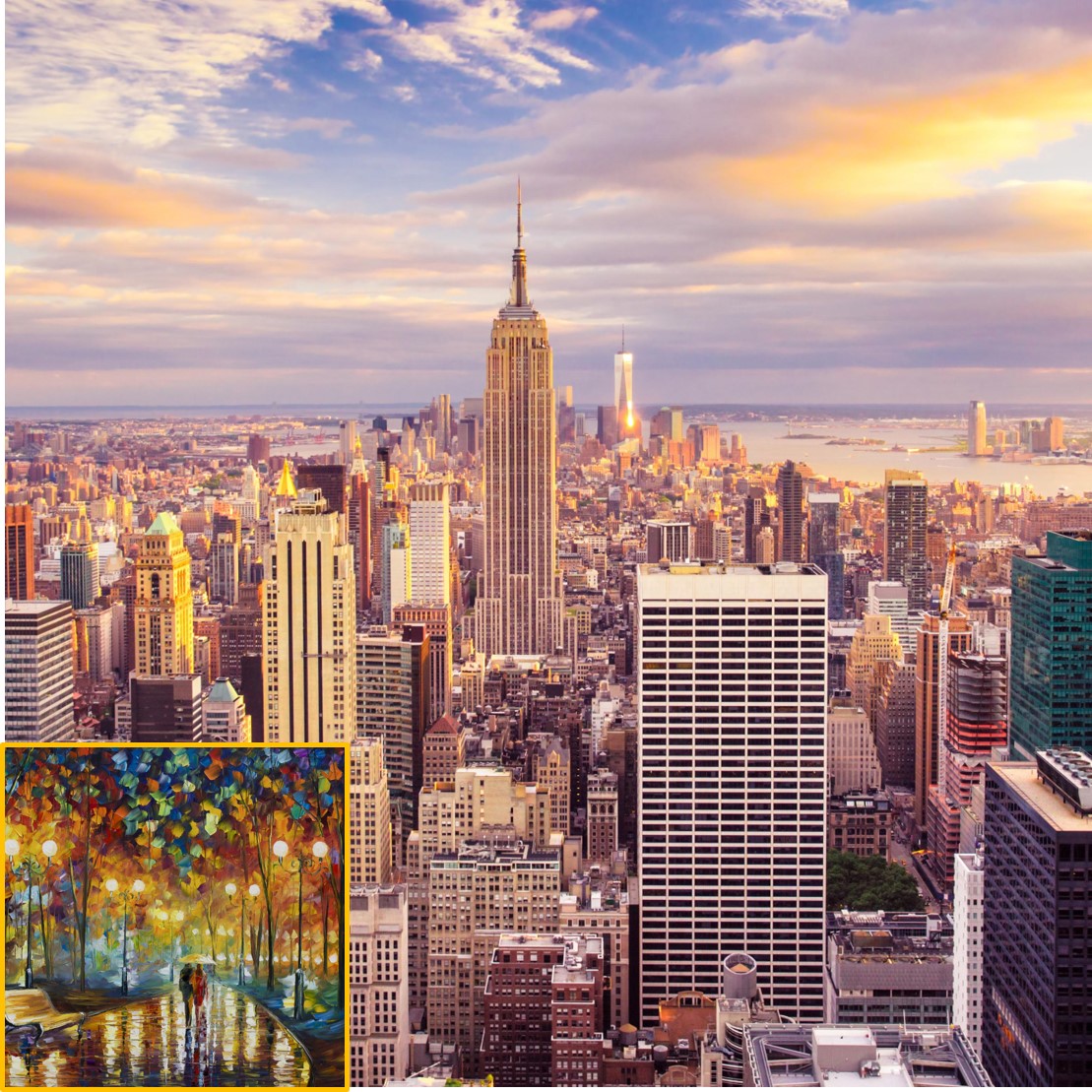} &
     \includegraphics[width=0.24\linewidth]{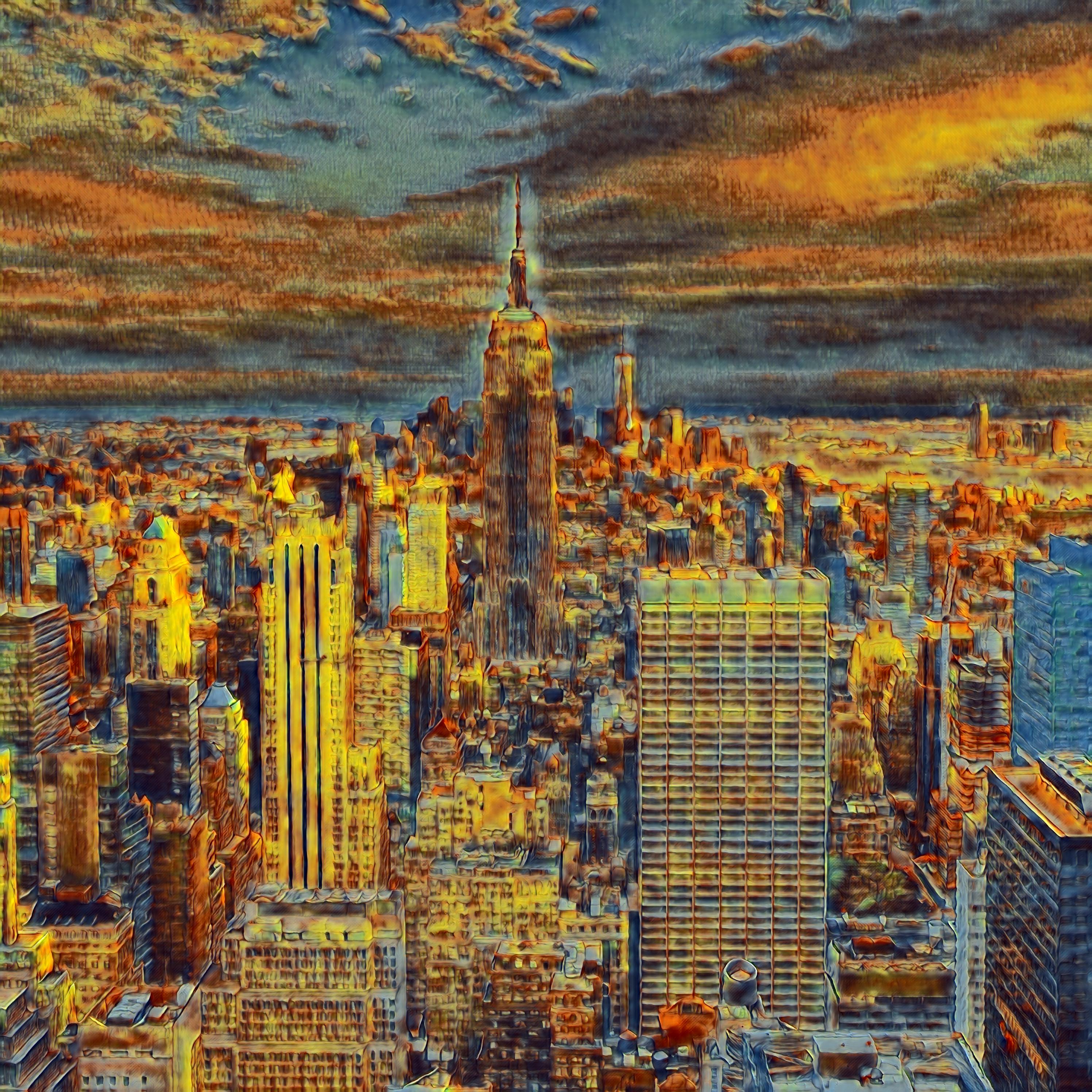} &
     \includegraphics[width=0.24\linewidth]{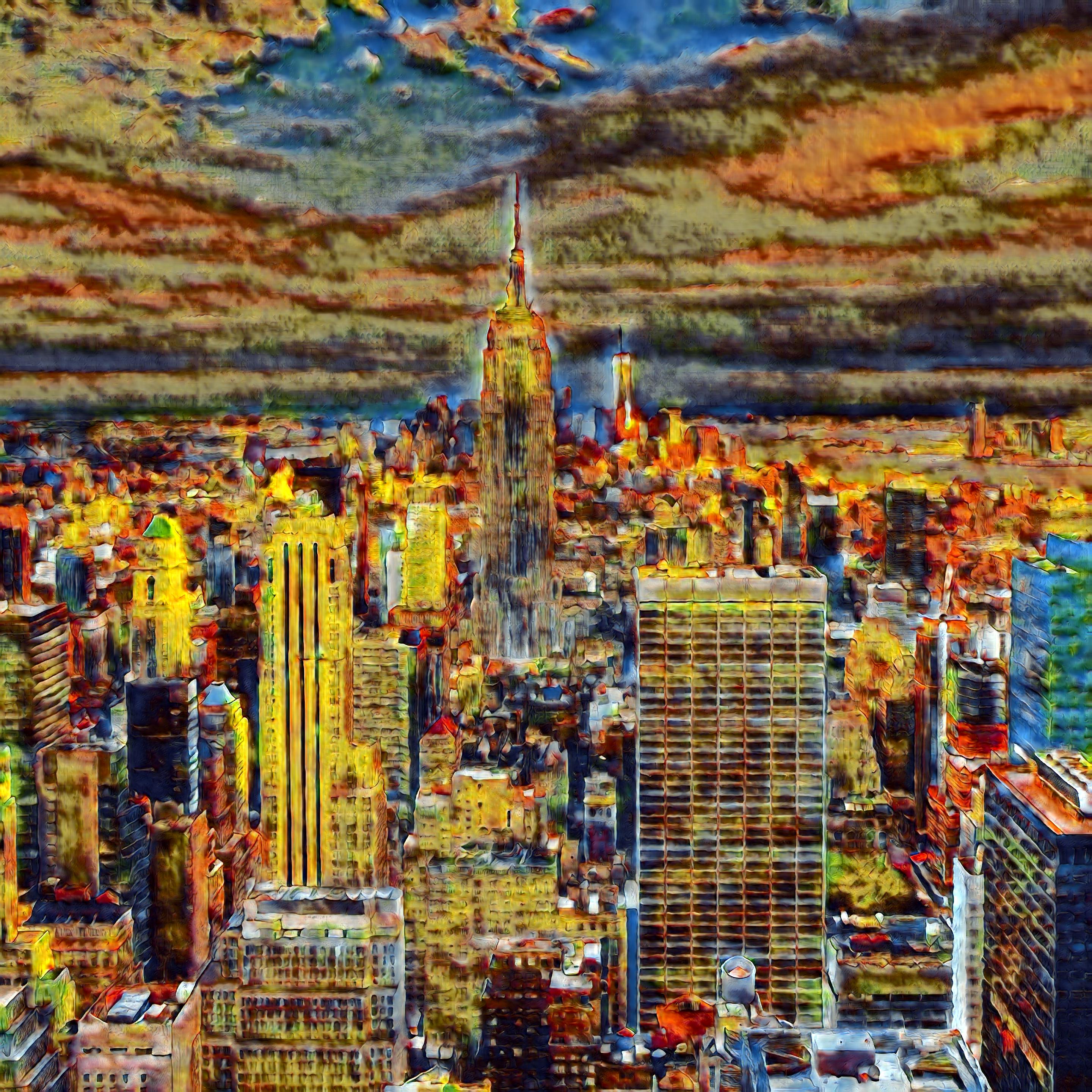} &
      \includegraphics[width=0.24\linewidth]{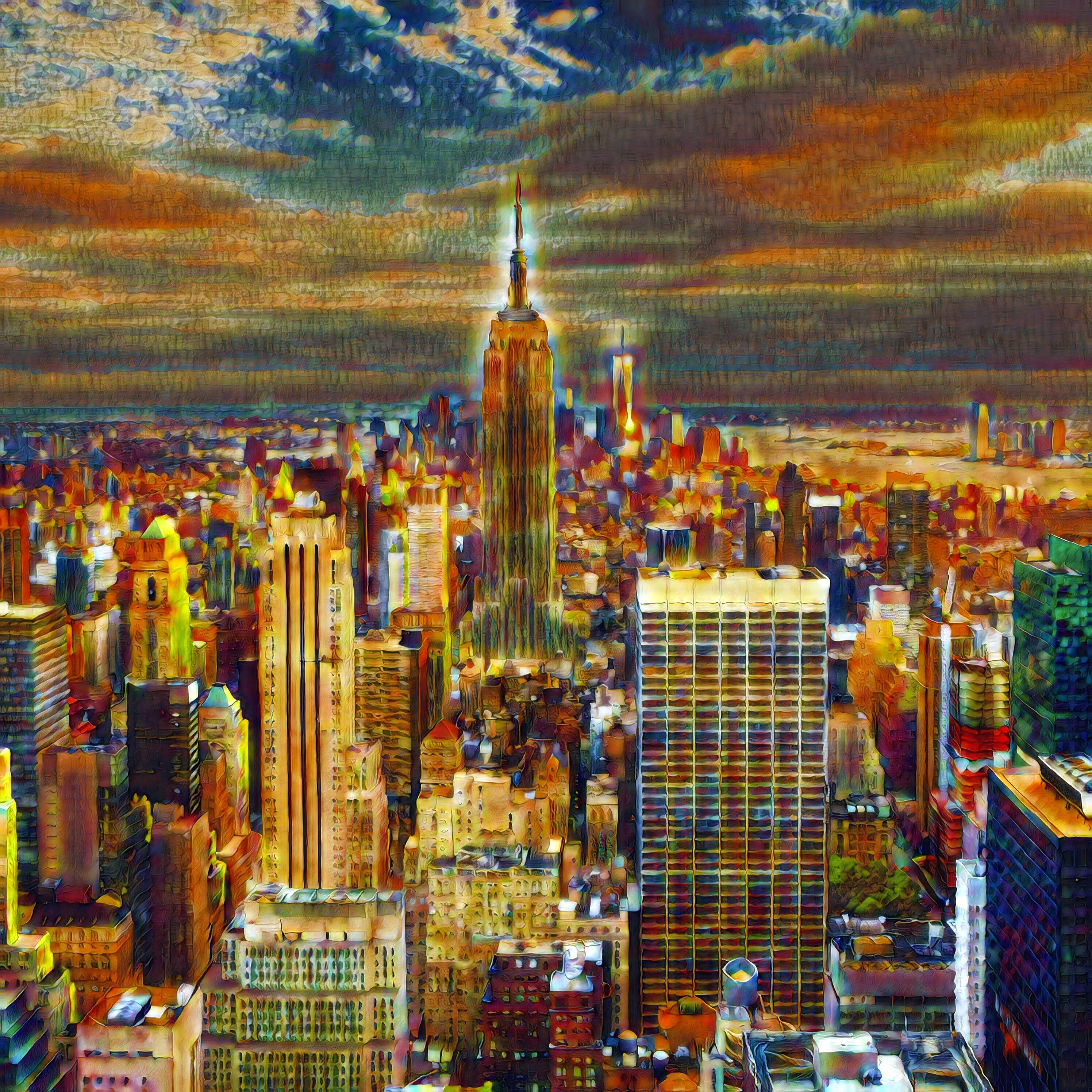} \\
     (a) C/S & (b) FP (\texttt{A}) & (c) FP (\texttt{B}) & (d) Ours \\
     \includegraphics[width=0.24\linewidth]{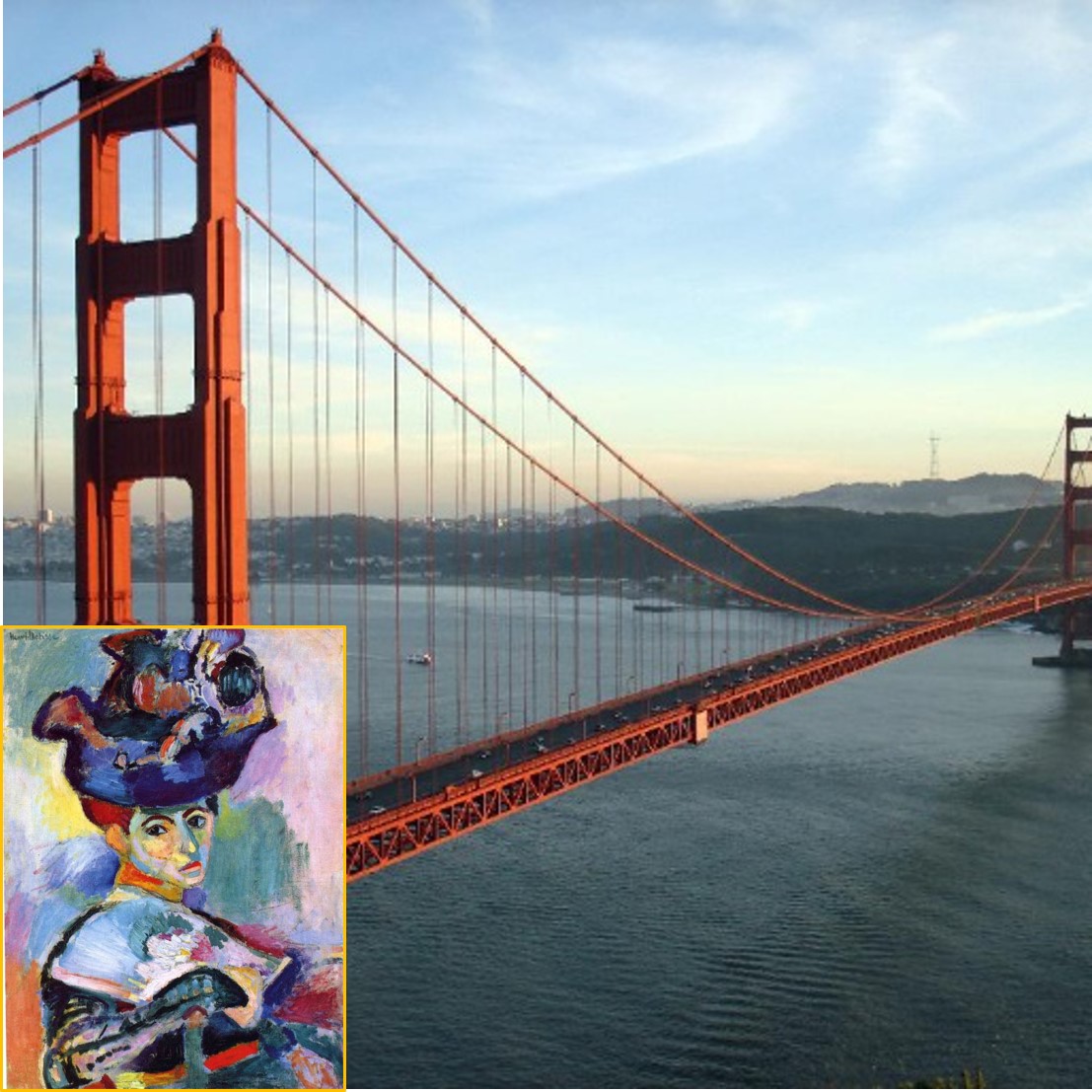} &
     \includegraphics[width=0.24\linewidth]{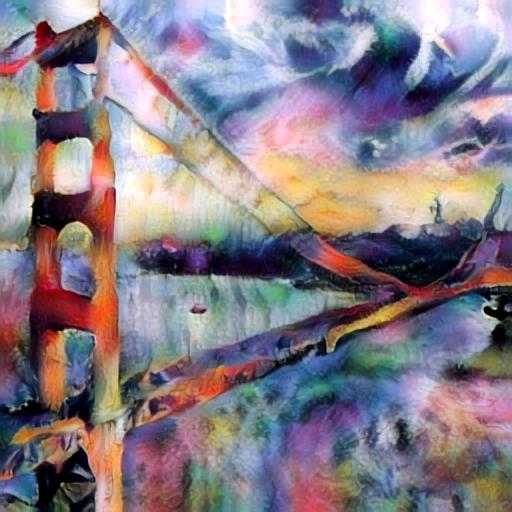} &
     \includegraphics[width=0.24\linewidth]{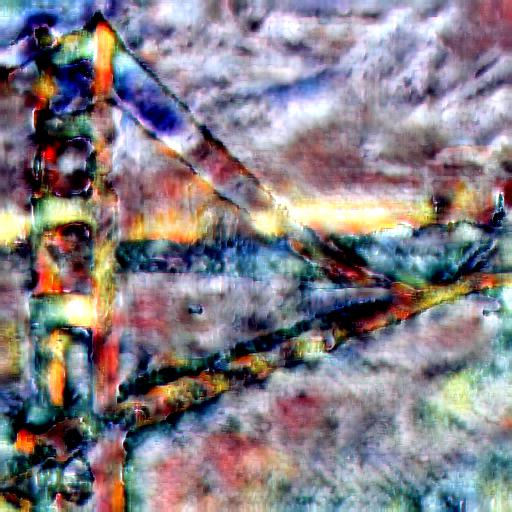} &
     \includegraphics[width=0.24\linewidth]{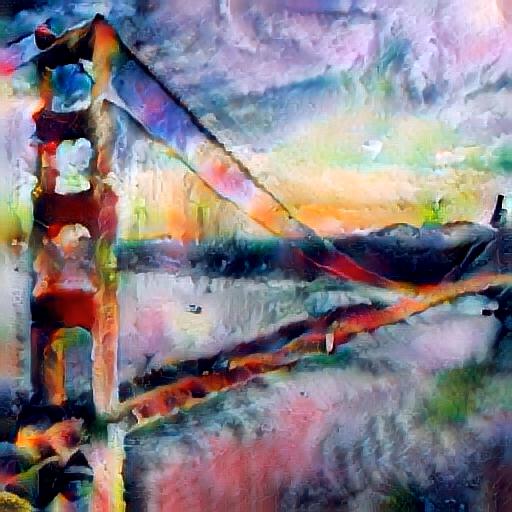} \\
     (a) C/S & (b) Original & (c) Ours$_{\text{KD}}$ & (d) Ours$_{\text{No KD}}$ \\
   \end{tabular}
   \vspace{-0.5em}
   \caption{Row 1: Stylization comparison of FP-pruned model \texttt{A} and \texttt{B}. Row 2: Stylization comparison between decoders using and not using KD with the WCT framework.}
   \label{fig:fp_plus_kd}
   \vspace{-1.5em}
\end{figure}

We briefly explain why existing knowledge distillation methods~\cite{liu2019knowledge,park2019relational,yu2019learning,peng2019correlation} generally work less effectively for style transfer. The FP-pruned VGG-19 (called \verb+A+) in our paper achieves a top-5 accuracy of $83.47\%$ on ImageNet. We had an intermediate model (called \verb+B+) in the middle of the fine-tuning process of \verb+A+, with a lower top-5 accuracy $80.55\%$. We compare the stylization quality of \verb+A+ and \verb+B+ with WCT and find that the results of \verb+A+ does \emph{not} show any advantage over those of \verb+B+ as the unpleasant messy textures still remain (Fig.~\ref{fig:fp_plus_kd}, Row 1). This implies that a small accuracy gain in classification (typically less than $3\%$, which is what the state-of-the-art distillation methods~\cite{liu2019knowledge,park2019relational,yu2019learning,peng2019correlation} can achieve at most) cannot really translate to the \emph{perceptual} improvement in neural style transfer. 
%
In addition, we only apply the proposed method to distilling the encoder, \emph{not the decoder}, because it will simply degrade the visual quality otherwise, as shown in Fig.~\ref{fig:fp_plus_kd} (Row 2). The reason is that, the decoder is responsible for image reconstruction, which is already trained with proper supervisions, \ie, the pixel and perceptual losses in Eq.~(\ref{eqn:pixel_perceptual_loss}). When applying distillation to the small decoder, the extra supervision from the original decoder does not help but undermining the effect of loss~(\ref{eqn:pixel_perceptual_loss}), thus deteriorating the visual quality of stylized results.

\section{Conclusion}
Input resolution is an important limitation for universal neural style transfer due to the large model size of CNNs.  
In this work, we present a new knowledge distillation method (\ie, Collaborative Distillation) to reduce the model size of VGG-19, which exploits the phenomenon that encoder-decoder pairs in universal style transfer construct an exclusive collaborative relationship. To resolve the feature size mismatch problem, a linear embedding scheme is further proposed.
Extensive experiments show the merits of our method on two universal stylization approaches (WCT and AdaIN). Further experiments within the Gatys stylization framework demonstrate the generality of our approach on the optimization-based style transfer paradigm.
Although we mainly focus on neural style transfer in this work, the encoder-decoder scheme is also generally utilized in other low-level vision tasks like super-resolution and image inpainting. The performance of our method on these tasks is worth exploring, which we leave as the future work.

\section*{Acknowledgment}
We thank Wei Gao and Lixin Liu for helpful discussions. This work is supported in part by the Natural Key R\&D Program of China under Grant
No.2017YFB1002400 and US National Science Foundation CAREER Grant No.1149783.

{\small
\bibliographystyle{ieee_fullname}
\bibliography{references}
}
\end{document}